\documentclass{article}

\usepackage{iclr2027_conference,times}
\PassOptionsToPackage{table}{xcolor}

\usepackage[utf8]{inputenc} 
\usepackage[T1]{fontenc}    
\usepackage{hyperref}       
\usepackage{url}            
\usepackage{booktabs}       
\usepackage{amsfonts}       
\usepackage{nicefrac}       
\usepackage{microtype}      
\usepackage{xcolor}         
\usepackage{amssymb}
\usepackage{amsmath}
\usepackage{amsfonts}
\usepackage{optidef}
\usepackage{physics}
\usepackage{bbm}
\usepackage{makecell}
\usepackage{algorithm}
\usepackage{algorithmic}
\usepackage{subcaption}
\usepackage{graphicx}
\usepackage{array}
\usepackage{multirow}
\usepackage{multicol}
\usepackage{caption}
\usepackage{natbib}
\usepackage{bm}
\usepackage{amsthm}
\usepackage{mathtools}
\usepackage{caption}
\usepackage{comment}
\usepackage{booktabs}
\usepackage[table]{xcolor}
\usepackage{tabularx}
\usepackage{longtable}
\usepackage{wrapfig}\usepackage{placeins}
\usepackage{subcaption}
\allowdisplaybreaks

\theoremstyle{plain}
\newtheorem{theorem}{Theorem}

\theoremstyle{definition}
\newtheorem{definition}[theorem]{Definition}

\graphicspath{{Figures/}}
\hypersetup{hidelinks,pdfauthor={Chuiyang Meng, Ming Tang, Vincent W.S. Wong},pdftitle={CORTEX: Learning to Share and Specialize in Dense Language Models}}

\usepackage{xcolor}
\definecolor{cortexgray}{RGB}{242,243,245}
\definecolor{grey}{rgb}{0.5,0.5,0.5}

\makeatletter
\@ifundefined{algorithmicrequire}{}{}
\@ifundefined{algorithmicensure}{}{}
\makeatother

\providecommand{\CORTEXStage}[1]{%
  \STATE {\setlength{\fboxsep}{1.5pt}%
  \colorbox{cortexgray}{%
  \parbox{0.82\linewidth}{\textsc{\textbf{\scriptsize #1}}}}}%
}

\title{CORTEX: Learning to Share and Specialize in Dense Language Models}

\author{\begin{tabular}{c}
Chuiyang Meng$^{1}$ \quad Ming Tang$^{2}$ \quad Vincent W.S. Wong$^{3}$ \\[3pt]
{\normalfont $^{1}$Simon Fraser University} \\
{\normalfont $^{2}$Southern University of Science and Technology} \\
{\normalfont $^{3}$The University of British Columbia} \\[3pt]
{\normalfont\href{mailto:chuiyang_meng@sfu.ca}{{\fontencoding{T1}\selectfont\texttt{chuiyang\_meng@sfu.ca}}} \quad \texttt{tangm3@sustech.edu.cn}} \\
{\normalfont\texttt{vincentw@ece.ubc.ca}}
\end{tabular}}

\iclrfinalcopy

\begin{document}
\maketitle
\lhead{} 

\begin{abstract}
Large language models are trained on heterogeneous data mixtures, where different knowledge domains require both shared knowledge and specialization. 
Existing modular approaches typically impose explicit components or discover modules through interpretability analysis after training.
In this work, we propose CORTEX, a learning dynamics-inspired framework that learns internal modularization within dense language models.
CORTEX partitions trainable matrices into parameter groups and learns module assignments from domain-conditioned gradient and cross-domain gradient similarity. 
We introduce the selective lesion score and module-domain mutual information to characterize the target-domain lesion effects and alignment, and analyze how module assignment affects the trade-off between assignment bias and update magnitude.
Experiments with 160M, Qwen3-8B, and Qwen3-32B backbone models show that CORTEX achieves the highest synthetic-domain exact match and largest average perplexity reduction, while remaining competitive on real-domain evaluations and forming identifiable modules.
\end{abstract}

\section{Introduction}
Large language models (LLMs) have demonstrated outstanding capabilities across various domains \citep{brown2020language,yang2025qwen3}.
They are trained on a wide range of data mixtures. 
Natural language, code, scientific text, legal documents, mathematical reasoning, and instruction-following data all require the model to capture different patterns of prediction and reasoning \citep{gao2020pile, xie2023doremi, ye2025datamixinglaws, liu2025regmix, chen2025aioli}.
While these knowledge domains share vocabulary, syntax, world knowledge,
and reusable computational structure, they require different
abstractions, factual associations, and modes of generalization.

\textcolor{black}{This creates a trade-off between sharing and specialization.
On the one hand, a fully shared dense model allows different knowledge
domains to use the same parameters.
However, gradients from different domains may favor
conflicting updates to the same parameters, which may create interference during training.
On the other hand, a fully partitioned model assigns separate model parameters
to different knowledge domains.
This allows domain-specific updates but limits parameter
sharing for patterns that are common across domains.
}

Existing modular approaches usually take modularity as an explicit architectural component to support domain-specific updates when gradients from different domains favor conflicting updates.
Mixture-of-experts (MoE) introduces separate expert blocks and uses routing to activate them for different inputs, showing that specialization can improve the capacity and efficiency under heterogeneous training data \citep{shazeer2017outrageously,dai2024deepseekmoe,muennighoff2025olmoe,shi2025flexolmo}.
Parameter-efficient adaptation approaches, such as low-rank adaptation (LoRA), provide another form of specialization by introducing task-specific parameters \citep{hu2022lora,liu2024dora,huang2025hira}.
Subnetwork-based approaches identify sparse subnetworks \citep{frankle2019lottery} or allocate parameters to tasks \citep{mallya2018packnet}.
\textcolor{black}{Although the aforementioned approaches support specialization, they do not specify how to jointly learn shared and domain-specialized assignments over the existing model parameters on a heterogeneous data mixture.}

\begin{figure}[t]
    \centering
    \begin{minipage}[t]{0.24\textwidth}
        \centering
        \includegraphics[width=\linewidth]{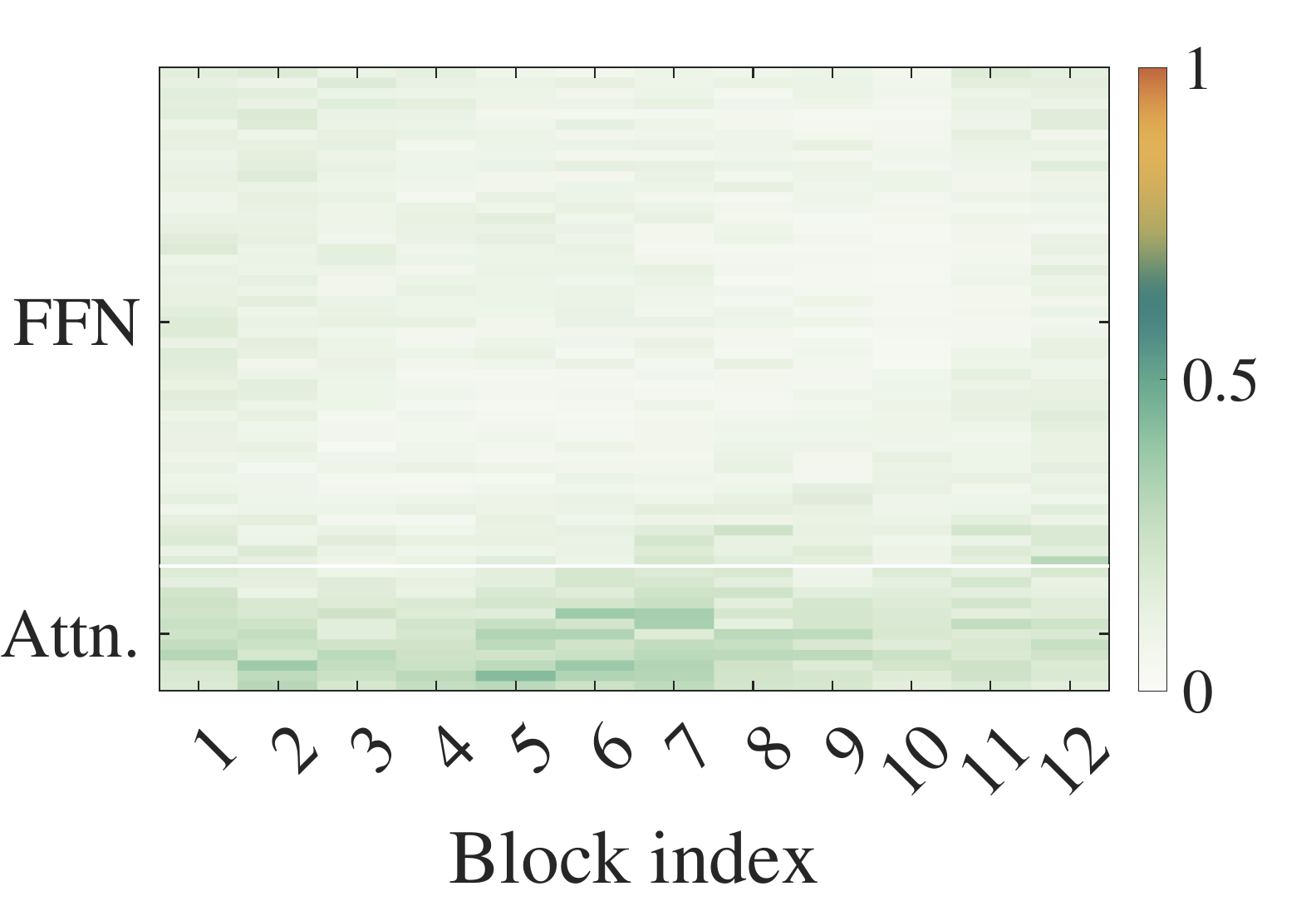}
        {\tiny (a) }
    \end{minipage}
    \begin{minipage}[t]{0.24\textwidth}
        \centering
        \includegraphics[width=\linewidth]{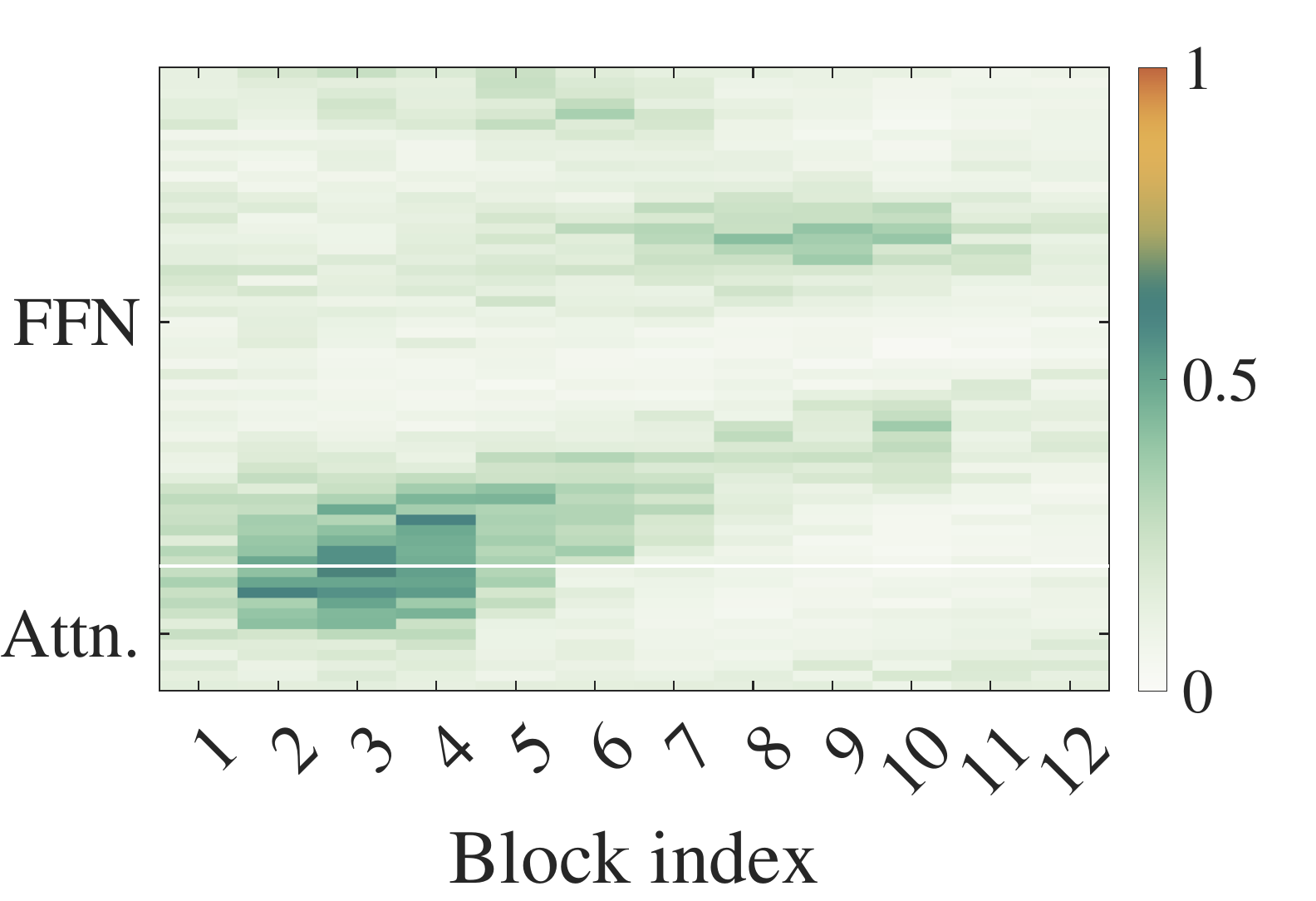}
        {\tiny (b) }
    \end{minipage}
    \begin{minipage}[t]{0.24\textwidth}
        \centering
        \includegraphics[width=\linewidth]{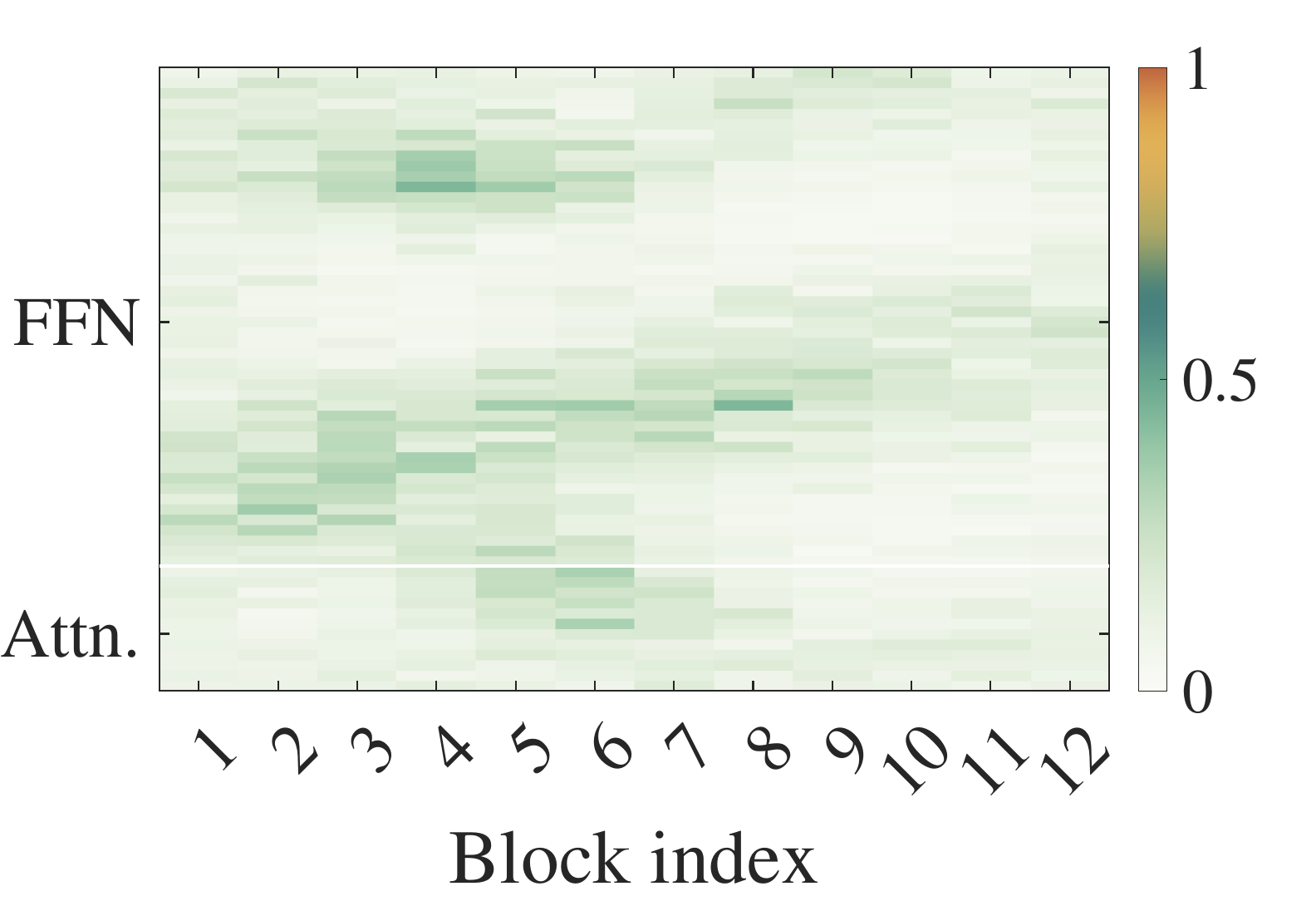}
        {\tiny (c) }
    \end{minipage}
    \begin{minipage}[t]{0.24\textwidth}
        \centering
        \includegraphics[width=\linewidth]{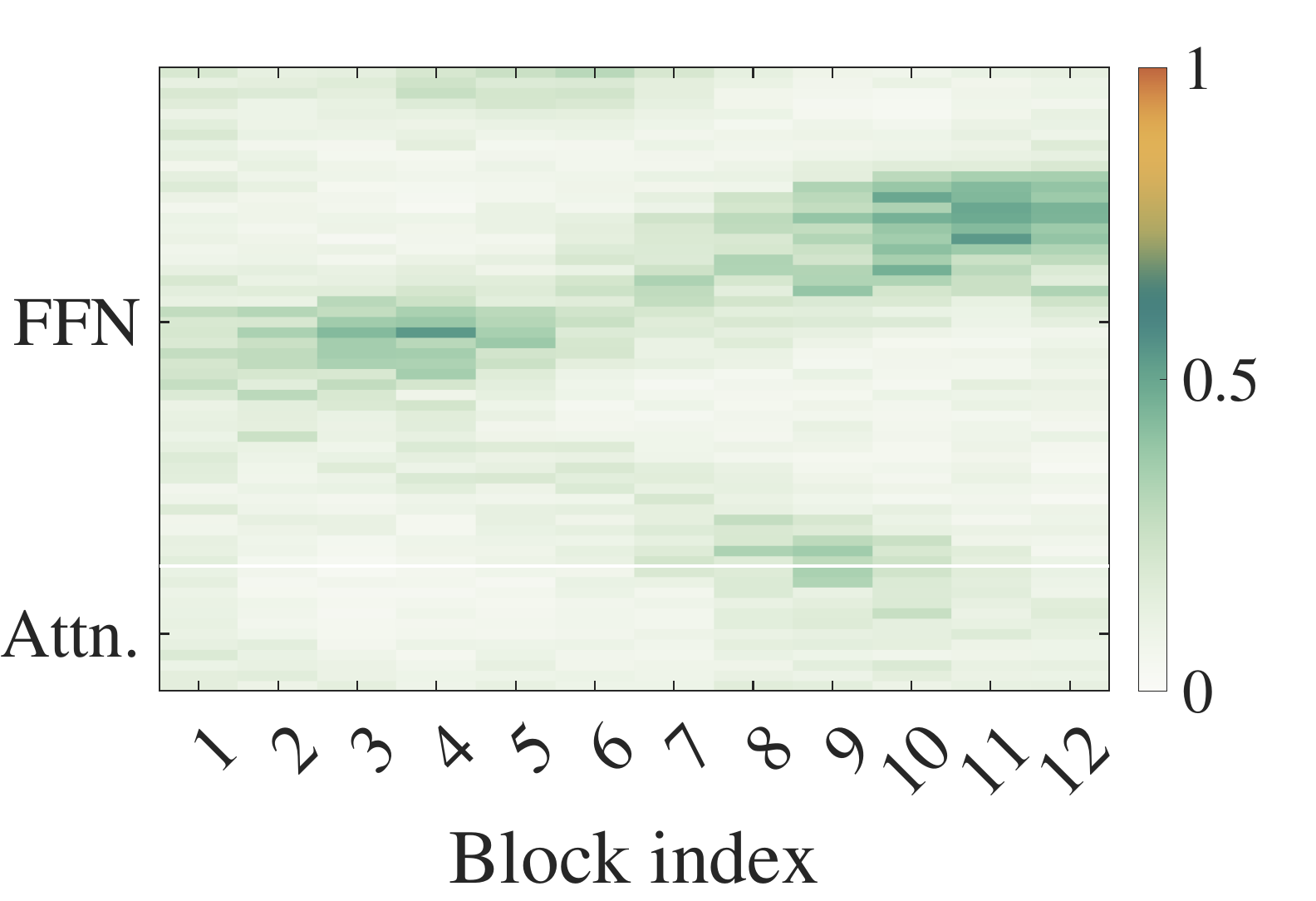}
        {\tiny (d) }
    \end{minipage}


    \begin{minipage}[t]{0.24\textwidth}
        \centering
        \includegraphics[width=\linewidth]{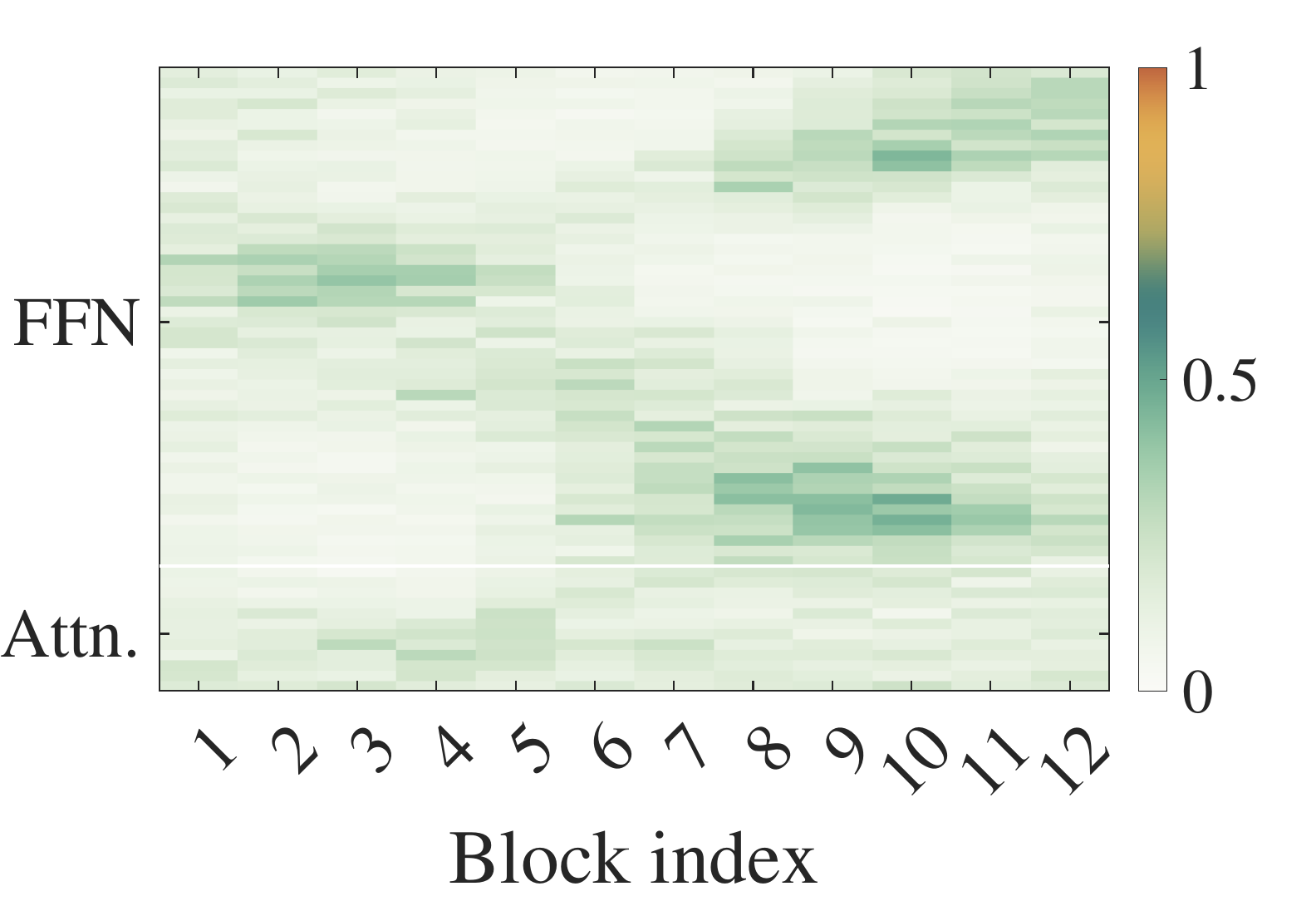}
        {\tiny (e) }
    \end{minipage}
    \begin{minipage}[t]{0.24\textwidth}
        \centering
        \includegraphics[width=\linewidth]{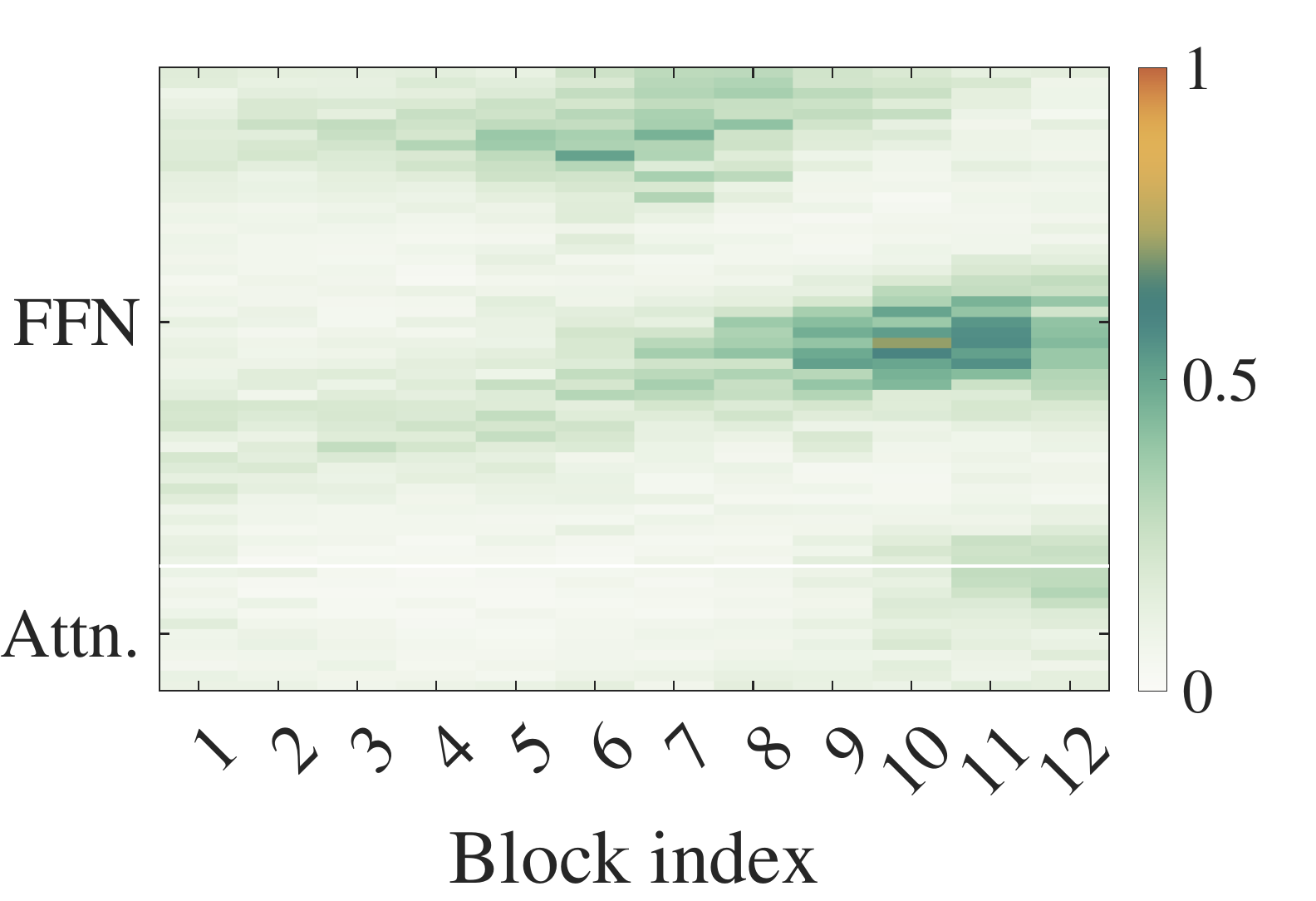}
        {\tiny (f) }
    \end{minipage}
    \begin{minipage}[t]{0.24\textwidth}
        \centering
        \includegraphics[width=\linewidth]{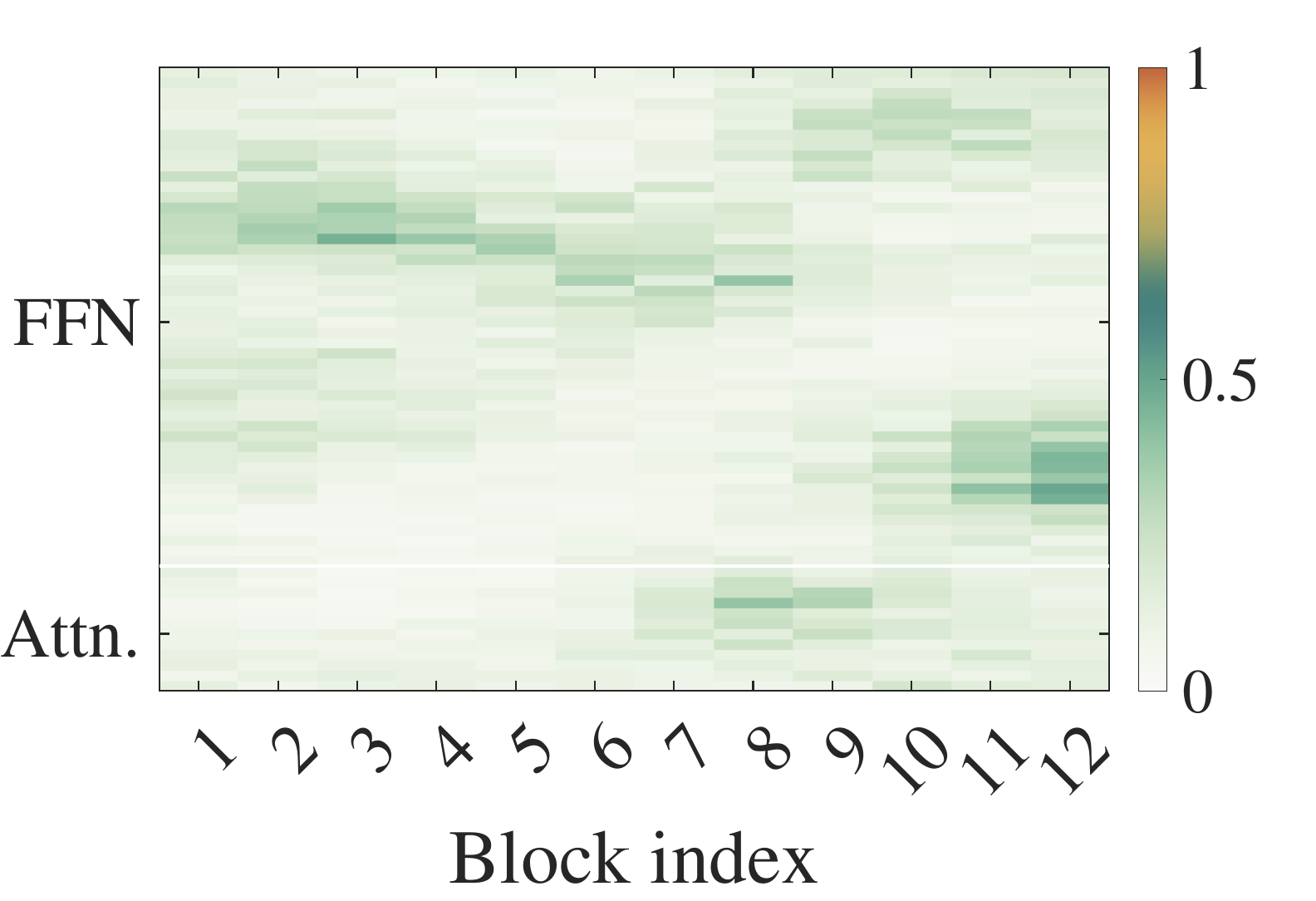}
        {\tiny (g) }
    \end{minipage}

    \caption{
    Visualization of CORTEX modularization on the 160M backbone model trained on the synthetic benchmark. 
    The heatmaps correspond to the shared and six knowledge domains, including (a) Shared, (b) COPY, (c) REVERSE, (d) ADD, (e) LOOKUP, (f) LOOKUP$\rightarrow$ADD, and (g) REVERSE$\rightarrow$LOOKUP, respectively.
    The columns denote Transformer blocks from 1 to 12, and the rows denote parameter groups within attention heads (Attn.) and the feed-forward network (FFN).}
    \label{fig:module_assignment_heatmap}
\end{figure}

Another related line of work studies the multi-task gradient from the optimization perspective. 
These works use learning dynamics to balance task losses and to reduce the cross-task interference caused by conflicting updates to shared parameters \citep{chen2018gradnorm,sener2018multi,yu2020gradient,chen2020graddrop,liu2021conflict}. 
However, these works modify the global optimization trajectory or task weights and do not learn a persistent modularization. 
This motivates us not only to adjust the model updates, but also to form modularization inside the dense parameter space.

Furthermore, studies from dense Transformer-based models suggest that internal functional structure can be observed without fully partitioning the model.
Prior works have identified specialized attention heads, sparse features, and knowledge circuits inside dense Transformer models \citep{voita2019heads,elhage2021transformercircuits,cunningham2024sparse,yao2024knowledgecircuits,gao2025sae}.
These works suggest that dense LLMs contain localized modular structures.
However, most of these works discover such structures after training. 
This leaves a gap between observing modularities in dense LLMs and training an LLM that can be organized as shared and specialized modules.

The aforementioned discussion raises the following question: 

\begin{quote}
\textit{Can modularity be formed inside dense LLMs by shaping learning dynamics?}
\end{quote}

Answering this question imposes the following requirement:
\textit{The model must identify shared and specialized structures inside the dense parameter space, learn stable module assignments during training, and align the learned modules with heterogeneous knowledge domains.}
To achieve this goal, we develop a framework called CORTEX, which supports modularization within dense LLMs. 
Fig. \ref{fig:module_assignment_heatmap} shows the modularization of CORTEX.
Our contributions are summarized as follows:

\begin{itemize}
    \item We propose CORTEX, a learning dynamics-inspired framework for learning modularization in dense LLMs. 
    CORTEX learns soft assignments of parameter groups, forming a shared module and multiple specialized modules during training.

    \item We develop a domain-conditioned module assignment mechanism. 
    CORTEX uses domain-conditioned group gradients to determine the module assignment: 
    parameter groups with cross-domain aligned gradients are assigned to the shared module, while domain-concentrated, less-aligned groups are assigned to the corresponding specialized module.

    \item We introduce the selective lesion score and module-domain mutual information to characterize the target-domain lesion effect and domain alignment. 
    We further show that they characterize module contribution and gradient concentration, while the module assignment weights control the convergence trade-off.

    \item We evaluate CORTEX on synthetic and real mixtures with 160M, Qwen3-8B, and Qwen3-32B, against Dense, Random, MoE \citep{shazeer2017outrageously}, MoM \citep{gong2024mixture}, UpIT \citep{hui2025upcycling}, and Self-MoE \citep{kang2025selfmoe}. 
    CORTEX achieves the highest exact match across all six synthetic domains, obtains the largest average perplexity reduction, and achieves best or competitive results on real domains while forming identifiable modules.
    
\end{itemize}

\section{Related Work}
\paragraph{Explicit experts and modularity.} 
MoE introduces expert blocks and activates only a subset of them for each input.
Building on the classical structure \citep{jacobs1991adaptive}, modern sparse MoE systems demonstrate that specialization can improve capacity and efficiency at scale \citep{shazeer2017outrageously, fedus2022switch, du2022glam, dai2024deepseekmoe, muennighoff2025olmoe, li2026moe_surpass_dense}.
Recent analyses further study the behavior and modularization degree of MoE-based language models \citep{lo2025closer}. 
Subnetworks \citep{frankle2019lottery}, packed architectures \citep{rusu2016progressive, mallya2018packnet}, and parameter-efficient adaptation approaches such as LoRA \citep{houlsby2019adapter, hu2022lora, dettmers2023qlora, liu2024dora, wang2025lorapro, huang2025hira} show that specialization can be achieved without updating all model parameters. 
\textcolor{black}{Modularizing-while-training (MwT) \citep{qi2024modularizing} is trained with modularization objectives to obtain reusable class-specific modules, while MODA \citep{ngo2026dnn} regulates activation patterns and extracts modules for reuse and replacement.}
\textcolor{black}{However, their modularization objectives penalize mask or activation similarity across classes. This can discourage beneficial parameter sharing. }
CORTEX learns shared and specialized modules within dense LLMs by coupling domain-conditioned module assignment with 
model updates, without introducing explicit experts. 

\paragraph{Multi-task gradient optimization.}
Another related line of work balances task losses or adjusts update directions using task-conditioned gradients \citep{chen2018gradnorm,sener2018multi,yu2020gradient,chen2020graddrop,liu2021conflict}. 
These approaches do not learn persistent module assignments within the dense parameter space. \textcolor{black}{Recon \citep{shi2023recon} uses gradient conflicts to select shared layers and copies these layers for different tasks. This changes the model structure and introduces additional parameters. In contrast, CORTEX learns shared and specialized assignments over the existing model parameters during training.}


\paragraph{Implicit modularity and mechanistic interpretability.} 
Prior work has identified specialized attention heads \citep{voita2019heads}, feed-forward layers with key-value memories \citep{geva2021ffn}, knowledge neurons \citep{dai2022knowledge}, and knowledge circuits \citep{yao2024knowledgecircuits}. 
Feed-forward layers can also be converted into expert-like structures \citep{zhang2022moefication}, and emergent modularity has been observed in pretrained LLMs \citep{zhang2023emergent, qiu2024unlocking}.
Recent work further studies sparse features and compositional circuits in language models \citep{elhage2021transformercircuits, cunningham2024sparse, gao2025sae, mondorf2025circuit}.
They do not make module assignment a trainable objective, and the discovered structures are not necessarily aligned with knowledge domains.


\section{CORTEX}
In this section, we introduce the modular parameterization of CORTEX and its domain-conditioned module assignment and training procedure.
A list of key notations is presented in Appendix \ref{sec:notation_table}.

Let $\mathcal{X}$, $\mathcal{Y}$, and $\mathcal{K} = \{1, 2, \ldots, K\}$ denote set of inputs, outputs, and $K$ knowledge domains, respectively.
We represent a training sample by a triplet $(X, Y, C)$, where $X \in \mathcal{X}$ is the input, $Y \in \mathcal{Y}$ is the output, and $C \in \mathcal{K}$ is the label of the knowledge domain. 
A knowledge domain may represent a domain (e.g., legal text), a skill (e.g., math reasoning), or a fact, as long as it is captured by a measurable conditional distribution over $(X,Y)$.
In particular, $C = k$ means that the training sample is from the $k$-th knowledge domain.
Let $\pi_{k}$ denote the probability that a randomly drawn training sample belongs to the $k$-th knowledge domain.
We have $\pi_{k} = \mathbb{P}(C=k) > 0$ and $\sum_{k=1}^{K}\pi_{k} = 1$.
For each knowledge domain $k\in\mathcal{K}$, let $\mathcal{P}_{k}$ denote its conditional joint distribution of $(X,Y)$ given $C = k$.
Then, the overall training data distribution is a mixture of $\mathcal{P}_{k}$, which is given by $\mathcal{P} = \sum_{k=1}^{K}\pi_{k} \mathcal{P}_{k}$.

\subsection{Modular Parameterization}
\begin{figure}[t]
    \centering
    \begin{minipage}[t]{0.48\textwidth}
        \centering
        \includegraphics[width=\linewidth]{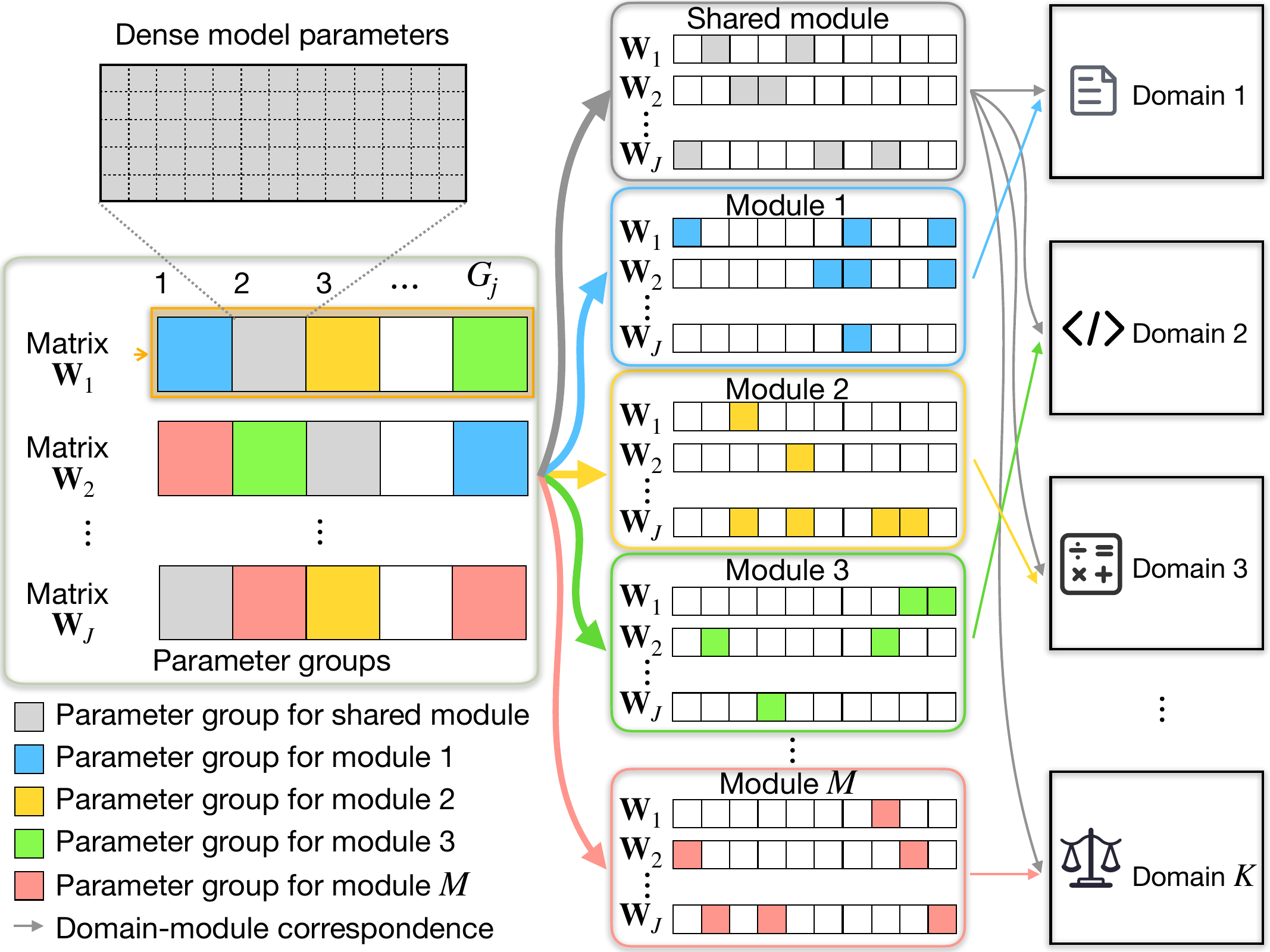}
        {\tiny (a) }
    \end{minipage}
    \hspace{10pt}
    \begin{minipage}[t]{0.48\textwidth}
        \centering
        \includegraphics[width=\linewidth]{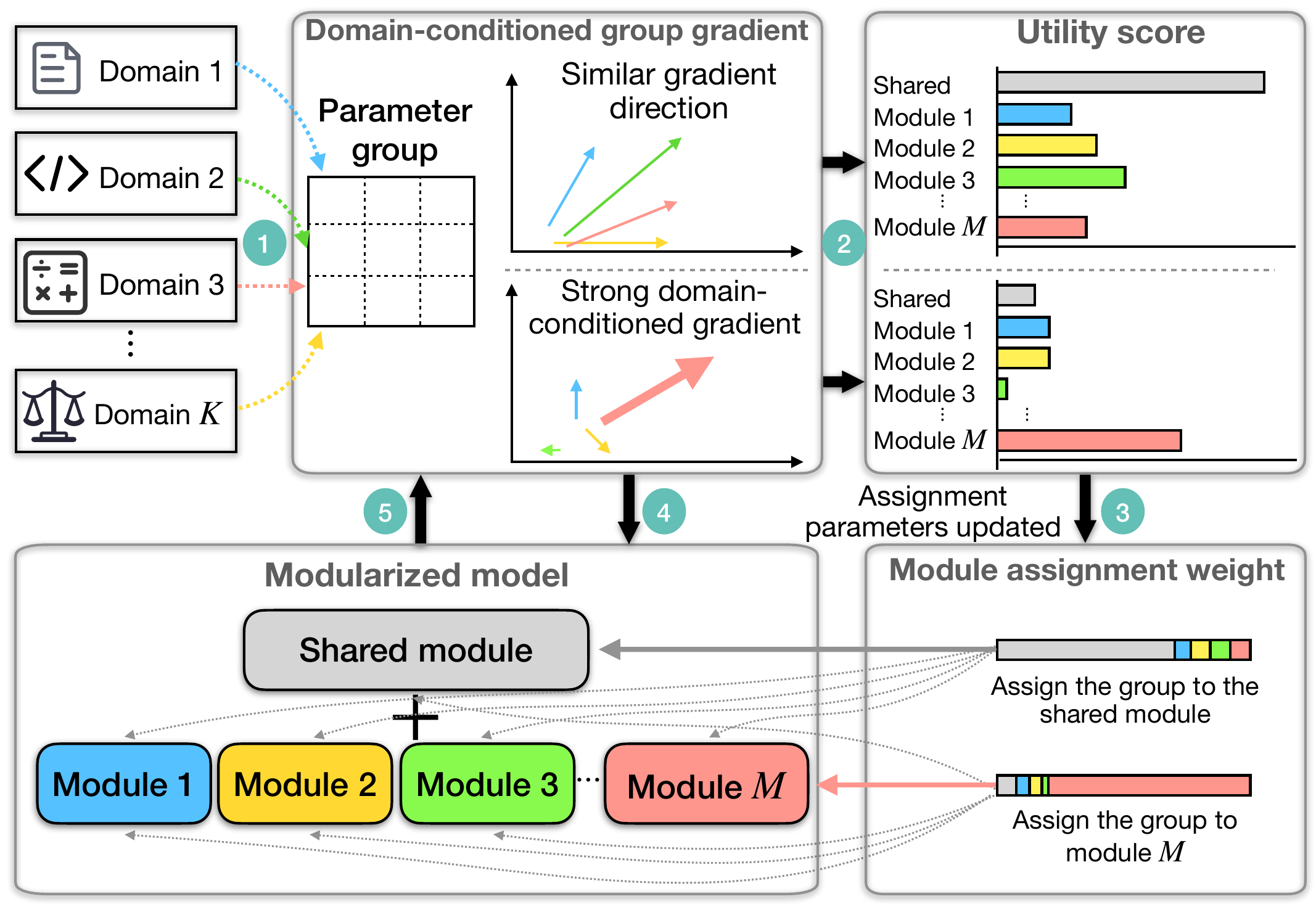}
        {\tiny (b) }
    \end{minipage}

    \caption{
    A schematic illustration of CORTEX, including (a) the relationship among knowledge domains,  parameter groups, and specialized modules, and (b) the modularization process of CORTEX.}
    \label{fig:overall_CORTEX}
\end{figure}

To learn module assignments inside a dense model, CORTEX first specifies the granularity at which trainable parameters are allocated to the shared and specialized modules.
This granularity should distinguish shared and domain-specific knowledge while keeping the assignment stable during training.
Therefore, CORTEX divides each trainable weight matrix into \textit{parameter groups} and learns the assignment weights of each group for the shared and specialized modules.
The relationship among knowledge domains, parameter groups, and specialized modules is presented in Fig. \ref{fig:overall_CORTEX} (a).

\paragraph{Parameter group}
Let $\mathcal{J} = \{1, 2, \ldots, J\}$ denote the set of $J$ indices of trainable weight matrices in the model. 
For each index $j\in\mathcal{J}$, let $\mathbf{W}_{j} \in \mathbb{R}^{d_{j}^{\mathrm{out}}\times d_{j}^{\mathrm{in}}}$ denote the $j$-th trainable weight matrix, where $d_{j}^{\mathrm{out}}$ and $d_{j}^{\mathrm{in}}$ are its output and input dimensions, respectively.
The parameter groups define the allocation granularity within each trainable matrix, and the projection operator selects disjoint groups that can be recombined into the original dense matrix.

\begin{definition}[Parameter group]
\textit{For trainable matrix $\mathbf{W}_{j}$, $j\in\mathcal{J}$, let $\mathcal{G}_{j} = \{1, 2, \ldots, G_{j}\}$ denote the index set of $G_{j}$ corresponding parameter groups.
In particular, we represent the projection of the $j$-th trainable parameter matrix onto the $g$-th parameter group by using a linear projection operator as} $\Pi_{j,g}: \mathbb{R}^{d_j^{\mathrm{out}}\times d_j^{\mathrm{in}}} \to \mathbb{R}^{d_j^{\mathrm{out}}\times d_j^{\mathrm{in}}}, \quad g\in\mathcal{G}_{j}$.
\textit{\(\Pi_{j,g}(\mathbf{W}_j)\) keeps the entries of the \(g\)-th parameter group and sets all other entries to zero.
It satisfies for arbitrary $g \neq g'$, the supports of two projection operators are disjoint, i.e., $\operatorname{supp}_{\mathrm{str}}(\Pi_{j,g})\cap\operatorname{supp}_{\mathrm{str}}(\Pi_{j,g'}) =\emptyset$. 
All parameter groups construct the original matrix, i.e.
} $\sum_{g\in\mathcal{G}_{j}}\Pi_{j,g}(\mathbf{W}_{j})=\mathbf{W}_{j}$.
\end{definition}

This projection-based reparameterization allows us to construct each module by combining selected parameter groups.
For Transformer backbones, the projections $\Pi_{j,g}$ are chosen along the functional axes. For feed-forward network (FFN) matrices, a parameter group is a block along the intermediate-channel axis. 
For attention matrices, a parameter group is a block of channels corresponding to one attention head.

\paragraph{Shared and specialized modules}
While there are heterogeneous knowledge domains, they still share a substantial amount of common knowledge.
This is because in LLMs, a large fraction of the computation is expected to be common instead of domain-specific. 
For this reason, we represent the model as a combination of (i) a shared module that captures the common knowledge and (ii) multiple specialized modules that capture heterogeneous knowledge domains.
We consider that each specialized module corresponds to a unique knowledge domain.
Let $\mathcal{M}=\{1,2,\ldots,M\}$ denote the index set of specialized modules, and let $m=0$ denote the shared module. 
Each specialized module $m\in\mathcal{M}$ is paired with one knowledge domain, denoted by $\kappa(m)\in\mathcal{K}$.
This pairing is fixed and one-to-one, with $M=K$.

{\color{black}CORTEX learns soft assignment weights to distribute each parameter group across the shared module and specialized modules.}
For the $g$-th parameter group in the $j$-th trainable parameter matrix, we define the trainable assignment parameter vector as $\boldsymbol{\psi}_{j,g} = ({\psi}_{j,g,0}, {\psi}_{j,g,1}, \ldots, {\psi}_{j,g,M}) \in \mathbb{R}^{M+1}$.
Let $\Psi = \{\boldsymbol{\psi}_{j,g}: j\in\mathcal{J}, g\in\mathcal{G}_{j}\}$ denote the set of trainable assignment parameter vectors.

\begin{definition}[Module assignment weight]
\textit{Let $s_{j,g,m}\in[0,1], m\in\{0\}\cup\mathcal{M}$ denote the assignment weight of the $g$-th  parameter group in the $j$-th trainable parameter matrix assigned to the $m$-th module.
$s_{j,g,m}$ is defined as} $s_{j,g,m} = \frac{\exp(\psi_{j,g,m})}{\sum_{r=0}^{M}\exp(\psi_{j,g,r})}, j\in\mathcal{J}, g\in\mathcal{G}_{j}, m\in\{0\}\cup\mathcal{M}$.
\end{definition}

We have $\sum_{m=0}^{M} s_{j,g,m}=1$.
We use this softmax reparameterization since it provides a normalized and differentiable relaxation of discrete module assignment such that the module assignment weight can be optimized through $\Psi$.
We denote $\mathbf{s}_{j,g} = (s_{j,g,0}, s_{j,g,1}, \ldots, s_{j,g,M}) \in\mathbb{R}^{M+1}$ as the assignment weight vector for the $g$-th  parameter group of the $j$-th trainable parameter matrix.
For the $j$-th trainable parameter matrix $\mathbf{W}_{j}$, the module-specific component associated with the $m$-th module is defined as $\mathbf{W}_{j,m} = \sum_{g\in\mathcal{G}_{j}} s_{j,g,m}\,\Pi_{j,g}(\mathbf{W}_{j}), j\in\mathcal{J}, m\in\{0\}\cup\mathcal{M}$.
{\color{black}Each module is a weighted combination of parameter groups, and the same parameter group can contribute to multiple modules.} 
Let $\mathcal{W}= \{\mathbf{W}_{1}, \mathbf{W}_{2}, \ldots, \mathbf{W}_{J}\}$ denote the full model.

\subsection{Domain-conditioned Module Assignment and Training}

CORTEX forms modularization during training by coupling module assignment with model updates. 
First, we introduce domain-conditioned group gradients to measure how each knowledge domain updates each parameter group. 
Second, we define the utility scores to determine whether a parameter group is more suitable for the shared module or a specialized module. 
{\color{black}Finally, the module assignment parameters and backbone model parameters are updated simultaneously using the utility scores and the main task loss, respectively.}
An overview of CORTEX is presented in Fig. \ref{fig:overall_CORTEX} (b).


\paragraph{Domain-conditioned group gradients}
A key question of CORTEX is how to update the assignment weights to form modules. 
The assignment should reflect how each parameter group represents different knowledge domains.
To this end, CORTEX uses domain-conditioned group gradients for module assignment. 
The intuition is as follows.
If several knowledge domains have similar update directions on a certain parameter group, then this parameter group should be assigned to the shared module. 
On the other hand, if a specific knowledge domain induces a large gradient magnitude on a certain parameter group, then this group is more suitable for a specialized module.

For a mini-batch $\boldsymbol{\xi}$, let $\boldsymbol{\xi}_{k} = \{(X,Y,C)\in\boldsymbol{\xi}:C=k\}$ denote the subset of samples from the $k$-th knowledge domain.
The loss of $\boldsymbol{\xi}_{k}$ is denoted as $F(\mathbf{\mathcal{W}; \boldsymbol{\xi}}_{k})$.
For the $g$-th parameter group in the $j$-th parameter matrix, its domain-conditioned gradient with respect to (w.r.t.) the $k$-th knowledge domain is defined as $\mathbf{G}_{j,g,k} = \Pi_{j,g}\left(\nabla_{\mathbf{W}_{j}}F(\mathbf{\mathcal{W}; \boldsymbol{\xi}}_{k})\right)$,
where $\nabla_{\mathbf{W}_{j}}F(\mathbf{\mathcal{W}; \boldsymbol{\xi}}_{k}) \in \mathbb{R}^{d_j^{\mathrm{out}}\times d_j^{\mathrm{in}}}$ denotes the gradient of $F(\mathbf{\mathcal{W}; \boldsymbol{\xi}}_{k})$ w.r.t. $\mathbf{W}_{j}$. 
The mixture gradient for this parameter group is
$\sum_{k\in\mathcal{K}}\pi_k\mathbf{G}_{j,g,k}$.

\paragraph{Utility score}
Based on the domain-conditioned group gradients, we define the utility score as a metric which measures how suitable a parameter group is for the shared module or a specialized module.
A higher utility score means that assigning this parameter group to the corresponding module is more consistent with the domain-conditioned gradients.

For the shared module, a parameter group should be useful to multiple knowledge domains, while a specialized module requires that the parameter group can represent a specific knowledge domain.
Hence, we define the utility score of the shared module and the $m$-th module as
$u_{j,g,0} = \frac{2}{K(K-1)}\sum_{1\leq p<q\leq K}\frac{\langle\mathbf{G}_{j,g,p}, \mathbf{G}_{j,g,q}\rangle_{F}}{\Vert\mathbf{G}_{j,g,p}\Vert_{F}\Vert\mathbf{G}_{j,g,q}\Vert_{F}}$ and $u_{j,g,m} = \frac{\Vert\mathbf{G}_{j,g,\kappa(m)}\Vert_{F}}{\sum_{k'\in\mathcal{K}}\Vert\mathbf{G}_{j,g,k'}\Vert_{F}} - \frac{1}{K-1}\sum_{q\in\mathcal{K}\backslash\{\kappa(m)\}}\frac{\langle\mathbf{G}_{j,g,\kappa(m)}, \mathbf{G}_{j,g,q}\rangle_{F}}{\Vert\mathbf{G}_{j,g,\kappa(m)}\Vert_{F}\Vert\mathbf{G}_{j,g,q}\Vert_{F}}, m\in\mathcal{M}$, respectively.
$\langle\cdot,\cdot\rangle_F$ and $\|\cdot\|_F$
denote the Frobenius inner product and norm, respectively.
The inner product is the sum of products of corresponding
matrix entries.
In particular, $u_{j,g,0}$ measures the average similarity between the domain-conditioned gradient across all pairs of knowledge domains.
A larger value of $u_{j,g,0}$ indicates that different domains update this group in similar directions.
Hence, the group is better suited to the shared module.
{\color{black}
For $u_{j,g,m}$, the first term measures the gradient magnitude
of the $\kappa(m)$-th domain normalized by the sum across
all domains on the same parameter group.
The second term measures the average cosine similarity between
this domain's gradient and those of the other domains.
}

\paragraph{Training procedure}
\textcolor{black}{The utility scores defined above provide the training signal for the module assignment.
CORTEX updates the module assignments and backbone parameters simultaneously after gradient collection. It calculates the utility scores from the domain-conditioned group gradients accumulated for the current update.
We use these scores to define the assignment loss as}
$\mathcal{L}_{\mathrm{assign}} = -\frac{1}{\sum_{j\in\mathcal{J}}G_{j}}\sum_{j\in\mathcal{J}}\sum_{g\in\mathcal{G}_{j}}\sum_{m\in\mathcal{M}\cup\{0\}}s_{j,g,m}u_{j,g,m}$.
The assignment parameters $\Psi$ are updated by minimizing $\mathcal{L}_{\mathrm{assign}}$.
{\color{black}The backbone parameters are updated according to the main task loss, which is given by} $\mathcal{L}_{\mathrm{main}} = \mathbb{E}_{\boldsymbol{\xi}\sim\mathcal{P}}\left[F({\mathcal{W}};\boldsymbol{\xi})\right]$.

The shared module is updated using gradients from all knowledge domains, while each specialized module mainly updates its parameters by using the domain-conditioned gradients.
Therefore, the $j$-th trainable parameter matrix is updated by
$\mathbf{W}_{j} \leftarrow \mathbf{W}_{j} - \eta\sum_{g\in\mathcal{G}_{j}}\left(s_{j,g,0}\sum_{k\in\mathcal{K}}\pi_{k}\mathbf{G}_{j,g,k} + \sum_{m\in\mathcal{M}}\pi_{\kappa(m)}s_{j,g,m}\mathbf{G}_{j,g,\kappa(m)}\right), j\in\mathcal{J}$,
where $\eta$ is the learning rate.
The algorithm of CORTEX is presented in Algorithm \ref{alg:cortex} in Appendix \ref{sec:algorithm}.

\section{Theoretical Analysis}
In this section, we provide theoretical analysis of CORTEX. 
\textcolor{black}{We first present Proposition 1 to motivate the utility scores.}
Then, we define two module-level interpretability metrics and introduce Propositions 2 and 3 to characterize specialization and module-domain alignment. 
In addition, Theorem 1 bounds the
average squared gradient norm for the modularized update.
We present the detailed assumptions and proof in Appendix \ref{sec:proof}.

\noindent\textbf{Proposition 1.}
\textit{We consider that $\nabla_{\mathbf{W}_j}F(\mathcal{W};\boldsymbol{\xi}_k)$
is locally Lipschitz continuous in $\mathbf{W}_j$,
with all other matrices fixed.
For the $g$-th parameter group in the $j$-th trainable matrix,
an update using the mixture gradient changes the loss
$F(\mathcal{W};\boldsymbol{\xi}_{k})$ by
$-\eta\left(\pi_k\|\mathbf{G}_{j,g,k}\|_F^2
+\sum_{q\in\mathcal{K}\setminus\{k\}}
\pi_q\langle\mathbf{G}_{j,g,k},\mathbf{G}_{j,g,q}\rangle_F\right)
+O(\eta^2)$}.

\textcolor{black}{
Proposition 1 characterizes the first-order loss change when using the mixture gradient.
The term \(\pi_k\|\mathbf{G}_{j,g,k}\|_F^2\) captures the contribution from the \(k\)-th domain's own gradient, while positive and negative cross-domain inner products contribute to decreasing and increasing its loss, respectively.
These terms motivate the use of gradient magnitude and alignment in the utility scores.
}

We consider that the number of modules matches knowledge domains, i.e., $M = K$.
Let $\mathcal{W}^{\mathrm{out}} = \{\mathbf{W}_{1}^{\mathrm{out}}, \mathbf{W}_{2}^{\mathrm{out}}, \ldots, \mathbf{W}_{J}^{\mathrm{out}}\}$, $\Psi^{\mathrm{out}}$, and $\mathbf{s}_{j,g,m}^{\mathrm{out}}$ denote the model parameters, assignment parameters, and assignment weights after training, respectively.
The $m$-th module in the $j$-th trained parameter matrix can be expressed as $\mathbf{W}_{j,m}^{\mathrm{out}} = \sum_{g=1}^{G_j} s_{j,g,m}^{\mathrm{out}}\,\Pi_{j,g}(\mathbf{W}_{j}^{\mathrm{out}}), j\in\mathcal{J}, m\in\{0\}\cup\mathcal{M}$.
To isolate the contribution of a  specialized module, we denote the model without the $m$-th module as $
\mathcal{W}^{\mathrm{drop},m} = \{\mathbf{W}_{1}^{\mathrm{out}} - \mathbf{W}_{1,m}^{\mathrm{out}}, \mathbf{W}_{2}^{\mathrm{out}} - \mathbf{W}_{2,m}^{\mathrm{out}}, \ldots, \mathbf{W}_{J}^{\mathrm{out}} - \mathbf{W}_{J,m}^{\mathrm{out}}\}, m\in\mathcal{M}$.
We define the empirical risk made with $\mathcal{W}$ on the $k$-th knowledge domain as $R_k(\mathcal{W}) = \mathbb{E}_{\boldsymbol{\xi}_{k}\sim \mathcal{P}_{k}}\left[F(\mathcal{W};\boldsymbol{\xi}_{k})\right]$.

Then, we adopt the idea of lesion study in neuroscience \citep{bates2003voxel, rorden2004using} to quantify the specialization between a module and a knowledge domain as follows.

\begin{definition}[Selective lesion score]
\textit{The lesion effect of removing the $m$-th module on the $k$-th knowledge domain is defined as $\Delta^{\mathrm{les}}_{m,k} = R_{k}(\mathcal{W}^{\mathrm{drop},m}) - R_{k}(\mathcal{W}^{\mathrm{out}})$.
We define the corresponding selective lesion score as} $\mathrm{SLS}(m,k) = \Delta^{\mathrm{les}}_{m,k} - \frac{1}{K-1}\sum_{q\in\mathcal{K}\backslash\{k\}}\Delta^{\mathrm{les}}_{m,q}, k\in\mathcal{K}, m\in\mathcal{M}$.
\end{definition}
Note that a larger selective lesion score indicates that disabling the $m$-th module leads to a larger performance degradation on the $k$-th knowledge domain than on the remaining domains, implying a stronger specialization between the $m$-th module and the $k$-th knowledge domain.

\noindent\textbf{Proposition 2.}
\textit{Removing the $m$-th module increases the empirical risk compared with the trained model if} $\mathrm{SLS}(m,k) > -\frac{1}{\pi_{k}}\sum_{q\in\mathcal{K}\backslash\{k\}}\left(\pi_{q} + \frac{\pi_{k}}{K-1}\right)\Delta^{\mathrm{les}}_{m,q}$.

{\color{black}
Proposition 2 shows that the selective lesion score, the lesion
effects on the other domains, and the domain mixture weights
jointly determine the change in the main task loss after module removal.
A larger value of $\mathrm{SLS}(m,k)$ indicates stronger specialization
of the $m$-th module for the $k$-th knowledge domain.
}

In addition, we analyze the degree of alignment between the assigned module and knowledge domains by adopting the idea of mutual information as follows.

\begin{definition}[Module-domain mutual information]
\textit{For the $m$-th module and the $k$-th knowledge domain, let $\mathrm{GS}_{m,k}
= \frac{\pi_k\sum_{j\in\mathcal J}\sum_{g\in\mathcal G_j}s^\mathrm{out}_{j,g,m}\left\|\mathbf{G}_{j,g,k}^{\mathrm{out}}\right\|_{F}}{\sum_{k'\in\mathcal{K}}\pi_{k'}\sum_{j\in\mathcal J}\sum_{g\in\mathcal G_j}\left\|\mathbf{G}_{j,g,k'}^{\mathrm{out}}\right\|_{F}}$ denote the gradient share which is the fraction of the total domain-conditioned gradient magnitude that comes from the $k$-th knowledge domain and is assigned to the $m$-th module.
We have $\sum_{m\in\{0\}\cup\mathcal{M}}\sum_{k\in\mathcal{K}}\mathrm{GS}_{m,k} = 1$.
The module-domain mutual information is defined as} $I^{\mathrm{mod}} = \sum_{m\in\{0\}\cup\mathcal{M}} \sum_{k\in\mathcal{K}}\mathrm{GS}_{m,k}\ln\frac{\mathrm{GS}_{m,k}}{\mathrm{GS}_{m}^{\mathrm{module}}\mathrm{GS}_{k}^{\mathrm{domain}}}$,
\textit{where $\mathrm{GS}_{m}^{\mathrm{module}} = \sum_{k\in\mathcal{K}}\mathrm{GS}_{m,k}, m\in\{0\}\cup\mathcal{M}$, and $\mathrm{GS}_{k}^{\mathrm{domain}} = \sum_{m\in\{0\}\cup\mathcal{M}}\mathrm{GS}_{m,k}, k\in\mathcal{K}$.}
\end{definition}
A larger value of $I^{\mathrm{mod}}$ indicates that the module assignment is more aligned with knowledge domains.

\noindent\textbf{Proposition 3.} 
\textit{The upper-bound of $I^{\mathrm{mod}}$ is $-\sum_{k\in\mathcal{K}}\mathrm{GS}_{k}^{\mathrm{domain}}\ln\mathrm{GS}_{k}^{\mathrm{domain}}$, which is achieved when each module corresponds to at most one knowledge domain.}

Proposition 3 provides a scale for interpreting the module-domain alignment.
The shared module corresponds to multiple knowledge domains, while each module mainly corresponds to a specific knowledge domain.
Thus, a larger value of \(I^{\mathrm{mod}}\) indicates that the domain-conditioned gradient magnitude is more concentrated on the assigned modules, implying higher module-domain alignment.

While both metrics evaluate the relationship between specialized modules and knowledge domains, the selective lesion score measures whether removing a module causes an increase in the risk for its target knowledge domain than for other domains. 
The module-domain mutual information measures whether the domain-conditioned gradient magnitude is concentrated on the assigned modules.
Then, we present a convergence bound for the modularized update.

\noindent\textbf{Theorem 1} (Convergence bound).
\textit{Let $\mathbf{W}^{t}$ and $\mathbf{W}^{\star}$ denote the model at round $t$ and an optimal model. For the modularized update with fixed $\mathbf{W}^{1}$ and constant learning rate $\eta>0$, Assumptions 1 and 2 in Appendix \ref{sec:proof_th1} imply}
$\frac{1}{T}\sum_{t=1}^{T}\mathbb{E}\left[\left\Vert\nabla\mathcal{L}_{\mathrm{main}}(\mathbf{W}^{t})\right\Vert_{F}^{2}\right]
\leq \frac{2}{T\eta}\left(\mathcal{L}_{\mathrm{main}}(\mathbf{W}^{1}) - \mathcal{L}_{\mathrm{main}}(\mathbf{W}^{\star})\right) + \frac{G^{2}}{T}\sum_{t=1}^{T}\sum_{j\in\mathcal{J}}\sum_{g\in\mathcal{G}_{j}}\sum_{m\in\mathcal{M}}\pi_{\kappa(m)}\mathbb{E}[A_{1,j,g,m}^{t}]+ \frac{2L\eta(G^{2}+\sigma^{2})}{T}\sum_{t=1}^{T}\sum_{j\in\mathcal{J}}\sum_{g\in\mathcal{G}_{j}}\mathbb{E}[A_{2,j,g}^{t}]$, \textit{where} $A_{1,j,g,m}^{t} = \left(1 - s_{j,g,0}^{t} - s_{j,g,m}^{t}\right)^{2}$ \textit{and} $A_{2,j,g}^{t} = \left(s_{j,g,0}^{t} + \sum_{m\in\mathcal{M}}\pi_{\kappa(m)}s_{j,g,m}^{t}\right)^{2}$.
\textit{The expectations are taken over the mini-batches sampled during training.}




In Theorem 1, $L$ is the smoothness constant. 
For each domain-conditioned
group gradient, $G^2$ bounds the squared norm of its conditional
expectation, and $\sigma^2$ bounds its conditional variance.
The first term is the standard initial-optimality gap term. 
The second term captures assignment bias. The third term arises from bounding the second moment of the modularized update. Larger shared assignment weights and smaller specialized assignment weights reduce the second term but increase the third term.
Hence, module assignment controls the trade-off between the assignment bias and update magnitude.


\section{Performance Evaluation}
We evaluate CORTEX on synthetic and real datasets, with additional results in Appendix \ref{sec:additional_results}.

\subsection{Experimental Setup}
\label{sec:experimental_setup}

\paragraph{Benchmarks and backbone models}
We evaluate CORTEX on two settings for the mixture of knowledge domains. 
First, we create a synthetic benchmark which contains six knowledge domains: COPY, REVERSE, ADD, LOOKUP, LOOKUP$\rightarrow$ADD, and REVERSE$\rightarrow$LOOKUP. 
Domain labels are used during CORTEX training to estimate domain-conditioned group gradients and are used for evaluation. 
Second, we use a real text knowledge-domain data mixture with general text (C4)~\citep{raffel2020exploring,dodge2021documenting}, code (Magicoder-OSS-Instruct-75K)~\citep{wei2024magicoder}, math (MathInstruct)~\citep{yue2024mammoth}, biomedical text (PubMedAbstractsSubset)~\citep{stuhlmann2025efficient}, legal text (Caselaw Access Project)~\citep{caselawaccessproject}, and reasoning data (WinoGrande)~\citep{sakaguchi2020winogrande}
For the synthetic benchmark, we use a 160M parameter decoder-only model with 12 layers, hidden size 768, and 12 attention heads.
For the real-domain benchmark, we use Qwen3-8B and Qwen3-32B \citep{yang2025qwen3} as dense backbones.

\paragraph{Baseline schemes}
(i) \textbf{Dense} trains the same dense model on the entire data mixture without module assignment.
{\color{black}(ii) \textbf{Random} uses the same domain-conditioned gradient collection and backbone update rule as CORTEX, while keeping the randomly initialized module assignments fixed throughout training.}
(iii) \textbf{MoE} replaces FFN with sparse experts and a router.
(iv) \textbf{MoM} \citep{gong2024mixture} uses token-level routers to select multi-head attention and FFN. It matches \textbf{Dense} in the number of backbone parameters and layers applied to each token.
(v) We adapt \textbf{UpIT} \citep{hui2025upcycling} using experts from intermediate checkpoints, then train only the router.
(vi) We adapt \textbf{Self-MoE} \citep{kang2025selfmoe} using the same training data without self-generated samples to train LoRA experts and a router.
All baseline schemes use the same data splits, training-token budget, and total number of optimizer updates. The settings are described in Appendix~\ref{sec:additional_results}. 
{\color{black}
CORTEX uses the same backbone as \textbf{Dense} and calculates assignment utilities from the domain-conditioned gradients.
CORTEX has a lower training step time than MoE and MoM.
In Appendix~\ref{sec:additional_results}, we compare the training step time and peak memory per GPU across five schemes.
}

\begin{table}[t]
\scriptsize
\centering
\caption{
Comparison between CORTEX and the best baseline.
We show exact match (EM, in \%) for the synthetic benchmark and negative log likelihood (NLL) for the real-domain benchmarks. 
}
\label{tab:compact_all_results}
\resizebox{\textwidth}{!}{
\begin{tabular}{llcccccc}
\toprule
\textbf{Backbone / Metric}
& \textbf{Approach}
& \textbf{Domain 1}
& \textbf{Domain 2}
& \textbf{Domain 3}
& \textbf{Domain 4}
& \textbf{Domain 5}
& \textbf{Domain 6} \\
\midrule

&
& \textbf{COPY}
& \textbf{REVERSE}
& \textbf{ADD}
& \textbf{LOOKUP}
& \textbf{LOOKUP$\rightarrow$ADD}
& \textbf{REVERSE$\rightarrow$LOOKUP} \\

\rowcolor{grey!5}
\multirow{-1}{*}{}
\multirow{2}{*}{\makecell[c]{\textbf{160M}\\EM (\%)}}
& Best Baseline
& $98.67{\pm}0.21$ {\scriptsize (MoM)}
& $97.84{\pm}0.27$ {\scriptsize (MoM)}
& $96.21{\pm}0.33$ {\scriptsize (MoM)}
& $96.48{\pm}0.32$ {\scriptsize (UpIT)}
& $90.82{\pm}0.49$ {\scriptsize (MoM)}
& $95.71{\pm}0.37$ {\scriptsize (Self-MoE)} \\

\rowcolor{green!12}
& \textbf{CORTEX}
& $\mathbf{99.42{\pm}0.18}$
& $\mathbf{98.83{\pm}0.22}$
& $\mathbf{97.62{\pm}0.29}$
& $\mathbf{98.14{\pm}0.26}$
& $\mathbf{94.76{\pm}0.41}$
& $\mathbf{96.88{\pm}0.33}$ \\

\midrule

&
& \textbf{General}
& \textbf{Code}
& \textbf{Math}
& \textbf{Biomedical}
& \textbf{Legal}
& \textbf{Reasoning} \\

\rowcolor{grey!5}
\multirow{2}{*}{\makecell[c]{\textbf{Qwen3-8B}\\NLL}}
& Best Baseline
& $1.6257{\pm}0.0108$ {\scriptsize (UpIT)}
& $1.6578{\pm}0.0124$ {\scriptsize (MoM)}
& $1.9058{\pm}0.0148$ {\scriptsize (Self-MoE)}
& $1.7709{\pm}0.0134$ {\scriptsize (UpIT)}
& $1.8171{\pm}0.0140$ {\scriptsize (UpIT)}
& $\mathbf{1.7822{\pm}0.0128}$ {\scriptsize (Self-MoE)} \\

\rowcolor{green!12}
& \textbf{CORTEX}
& $\mathbf{1.6199{\pm}0.0097}$
& $\mathbf{1.6096{\pm}0.0115}$
& $\mathbf{1.8694{\pm}0.0137}$
& $\mathbf{1.6778{\pm}0.0126}$
& $\mathbf{1.7791{\pm}0.0131}$
& $1.8258{\pm}0.0124$ \\

\midrule

&
& \textbf{General}
& \textbf{Code}
& \textbf{Math}
& \textbf{Biomedical}
& \textbf{Legal}
& \textbf{Reasoning} \\

\rowcolor{grey!5}
\multirow{2}{*}{\makecell[c]{\textbf{Qwen3-32B}\\NLL}}
& Best Baseline
& $\mathbf{1.3578{\pm}0.0093}$ {\scriptsize (UpIT)}
& $1.4383{\pm}0.0107$ {\scriptsize (MoE)}
& $1.5815{\pm}0.0122$ {\scriptsize (MoM)}
& $1.4878{\pm}0.0114$ {\scriptsize (MoE)}
& $1.5514{\pm}0.0119$ {\scriptsize (UpIT)}
& $1.5597{\pm}0.0116$ {\scriptsize (Self-MoE)} \\

\rowcolor{green!12}
& \textbf{CORTEX}
& $1.3820{\pm}0.0091$
& $\mathbf{1.4051{\pm}0.0102}$
& $\mathbf{1.5798{\pm}0.0117}$
& $\mathbf{1.4344{\pm}0.0109}$
& $\mathbf{1.5426{\pm}0.0110}$
& $\mathbf{1.5279{\pm}0.0111}$ \\

\bottomrule
\end{tabular}
}
\end{table}

\begin{figure}[t]
    \centering
    \begin{minipage}[t]{0.35\textwidth}
        \centering
        \includegraphics[width=\linewidth,trim=0 100bp 0 54bp,clip]{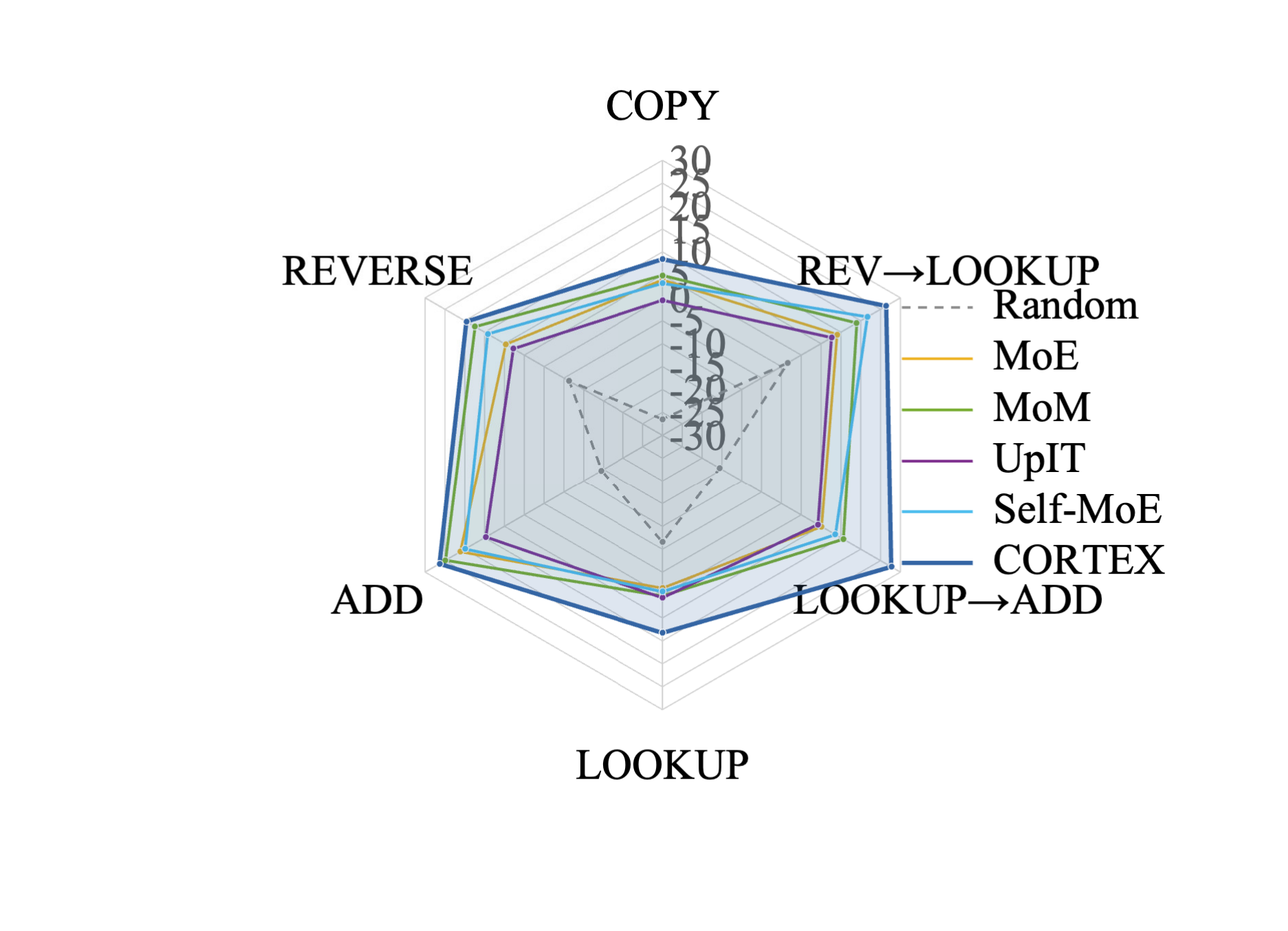}
        {\tiny (a) }
    \end{minipage}
    \hspace{-15pt}
    \begin{minipage}[t]{0.35\textwidth}
        \centering
        \includegraphics[width=\linewidth,trim=0 100bp 0 54bp,clip]{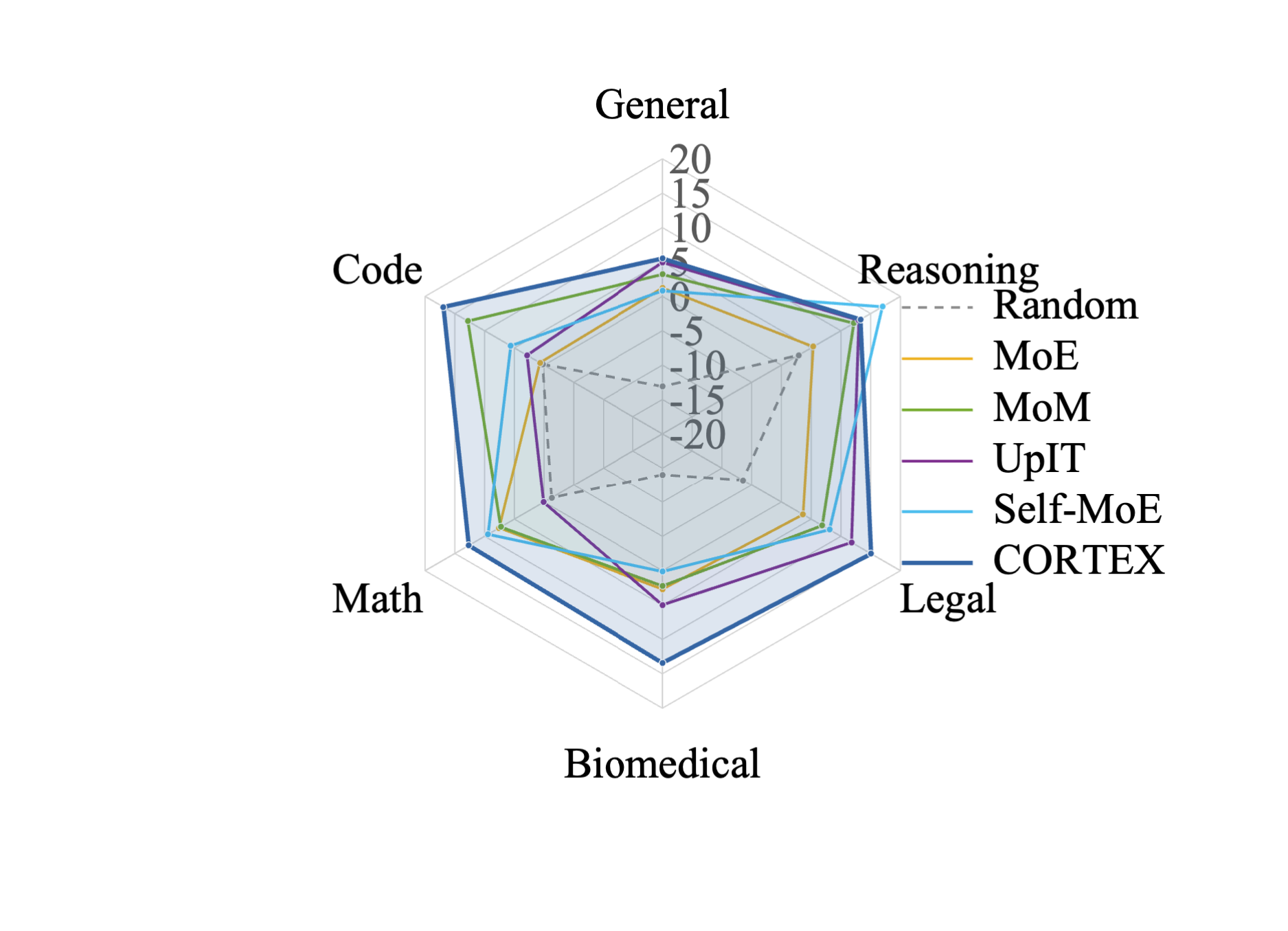}
        {\tiny (b) }
    \end{minipage}
    \hspace{-15pt}
    \begin{minipage}[t]{0.35\textwidth}
        \centering
        \includegraphics[width=\linewidth,trim=0 100bp 0 54bp,clip]{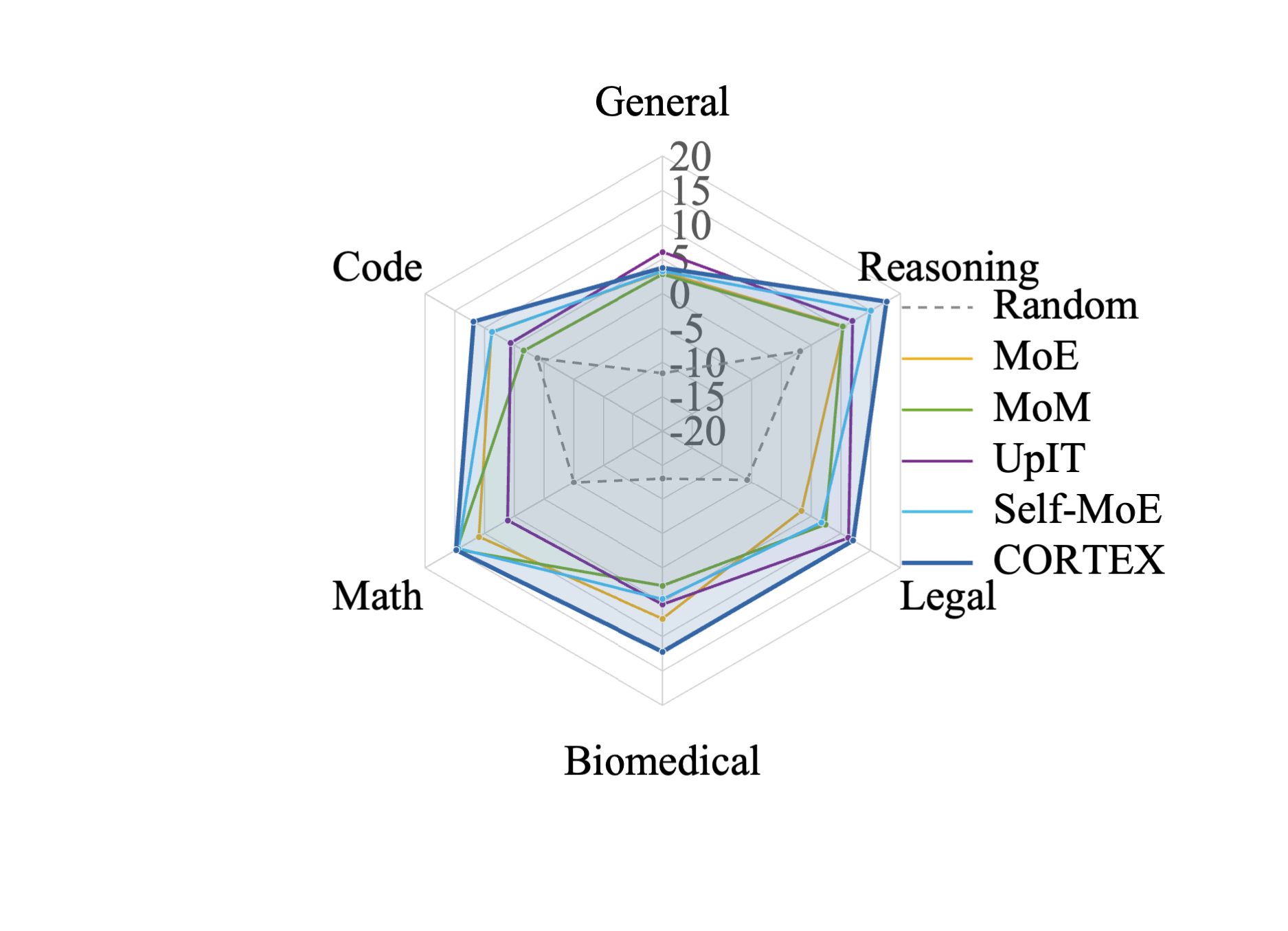}
        {\tiny (c) }
    \end{minipage}

    \caption{Comparison of the PPL reduction over \textbf{Dense} on (a) the 160M model, (b) Qwen3-8B, and (c) Qwen3-32B. The PPL reduction of \textbf{Dense} is zero. }
    \label{fig:core_reduction}
\end{figure}

\begin{table*}[!t]
\scriptsize
\centering
\caption{
Comparison of downstream scores between CORTEX and the best baseline on each domain.
We show the average $\pm$ standard deviation over four seeds.
}
\label{tab:downstream_best}
\setlength{\tabcolsep}{3.2pt}
\renewcommand{\arraystretch}{1}
\resizebox{0.9\textwidth}{!}{
\begin{tabular}{llcccccc}
\toprule
\textbf{Backbone}
& \textbf{Approach}
& \textbf{General}
& \textbf{Code}
& \textbf{Math}
& \textbf{Biomedical}
& \textbf{Legal}
& \textbf{Reasoning} \\
\midrule
\rowcolor{gray!5}
\multirow{2}{*}{\makecell[c]{\textbf{Qwen3-8B}\\Downstream}}
& Best Baseline
& $0.5785{\pm}0.0062$ {\scriptsize (UpIT)}
& $\mathbf{0.3097{\pm}0.0082}$ {\scriptsize (MoM)}
& $0.3414{\pm}0.0098$ {\scriptsize (Self-MoE)}
& $0.7220{\pm}0.0059$ {\scriptsize (UpIT)}
& $0.6219{\pm}0.0069$ {\scriptsize (UpIT)}
& $\mathbf{0.4992{\pm}0.0091}$ {\scriptsize (Self-MoE)} \\
\rowcolor{green!12}
& \textbf{CORTEX}
& $\mathbf{0.5900{\pm}0.0058}$
& $0.3074{\pm}0.0085$
& $\mathbf{0.3754{\pm}0.0102}$
& $\mathbf{0.7295{\pm}0.0055}$
& $\mathbf{0.6472{\pm}0.0064}$
& $0.4929{\pm}0.0088$ \\
\midrule
\rowcolor{gray!5}
\multirow{2}{*}{\makecell[c]{\textbf{Qwen3-32B}\\Downstream}}
& Best Baseline
& $\mathbf{0.6732{\pm}0.0050}$ {\scriptsize (UpIT)}
& $0.3930{\pm}0.0068$ {\scriptsize (MoE)}
& $0.5322{\pm}0.0083$ {\scriptsize (MoM)}
& $0.7950{\pm}0.0053$ {\scriptsize (MoE)}
& $0.6939{\pm}0.0059$ {\scriptsize (UpIT)}
& $0.6211{\pm}0.0076$ {\scriptsize (Self-MoE)} \\
\rowcolor{green!12}
& \textbf{CORTEX}
& $0.6641{\pm}0.0047$
& $\mathbf{0.4253{\pm}0.0071}$
& $\mathbf{0.5471{\pm}0.0080}$
& $\mathbf{0.8280{\pm}0.0049}$
& $\mathbf{0.6974{\pm}0.0056}$
& $\mathbf{0.6494{\pm}0.0072}$ \\
\bottomrule
\end{tabular}
}
\end{table*}

\subsection{Experimental Results}

\label{sec:result}
\paragraph{Learning performance comparison}
Table \ref{tab:compact_all_results} shows that CORTEX consistently outperforms the best baseline on exact match (EM) over all synthetic domains. 
On the real-domain benchmark, CORTEX achieves the best or competitive negative log likelihood (NLL) for both Qwen3-8B and Qwen3-32B when compared with the baseline schemes, which shows the effectiveness of CORTEX.
Fig. \ref{fig:core_reduction} compares the perplexity (PPL) reduction of different schemes over \textbf{Dense} across datasets and backbones.
CORTEX achieves a larger reduction in most cases.

We further perform downstream evaluation for generalization. 
After training, we evaluate frozen checkpoints without fine-tuning, using zero-shot prompts and greedy decoding with Qwen3 thinking disabled (Appendix~\ref{sec:downstream_protocol}). 
We use MMLU-Pro \citep{wang2024mmlu-pro} for general knowledge, LiveCodeBench \citep{jain2025livecodebench} for code, MATH/GSM8K \citep{hendrycks2021math, cobbe2021gsm8k} for mathematics, PubMedQA \citep{jin2019pubmedqa} for biomedical QA, LegalBench \citep{guha2023legalbench} for legal reasoning, and BIG-Bench Hard (BBH) \citep{suzgun2023bbh} for reasoning. 
Table \ref{tab:downstream_best} shows that CORTEX achieves higher average scores than the best baseline
in nine of the twelve backbone--domain settings.
Across four training seeds, the paired gains on mathematics
with Qwen3-8B and biomedical QA with Qwen3-32B are
$0.0340\pm0.0075$ and $0.0330\pm0.0042$, respectively.
These results show that CORTEX improves the learning performance.

\begin{table}[t]
\centering
\caption{
\color{black}Effect of the shared module on the synthetic benchmark. 
}
\label{tab:shared_ablation_synthetic}
\resizebox{\textwidth}{!}{
\begin{tabular}{lcccccccccc}
\toprule
\multirow{2}{*}{\textbf{Variant}}
& \multirow{2}{*}{\textbf{PPL}}
& \multirow{2}{*}{\textbf{NLL}}
& \multirow{2}{*}{\textbf{EM}}
& \multirow{2}{*}{\textbf{$I^{mod}$}}
& \multicolumn{6}{c}{$\mathrm{SLS}$} \\
\cmidrule(lr){6-11}
& & & & 
& \textbf{COPY}
& \textbf{REVERSE}
& \textbf{ADD}
& \textbf{LOOKUP}
& \textbf{LOOKUP$\rightarrow$ADD}
& \textbf{REV$\rightarrow$LOOKUP} \\
\midrule
\rowcolor{gray!5}
CORTEX w/o Shared
& $1.1302$
& $0.1224$
& $95.56$
& $1.1182$
& $0.1482$
& $0.1667$
& $0.1694$
& $0.1318$
& $0.1975$
& $0.1558$ \\

\rowcolor{green!12}
\textbf{CORTEX}
& $\mathbf{1.0762}$
& $\mathbf{0.0734}$
& $\mathbf{97.61}$
& $\mathbf{1.4560}$
& $\mathbf{0.1976}$
& $\mathbf{0.1742}$
& $\mathbf{0.1755}$
& $\mathbf{0.1769}$
& $\mathbf{0.2054}$
& $\mathbf{0.1726}$ \\
\bottomrule
\end{tabular}
}
\end{table}

\begin{figure}[t]
    \centering
    \begin{minipage}[t]{0.3\textwidth}
        \centering
        \includegraphics[width=\linewidth]{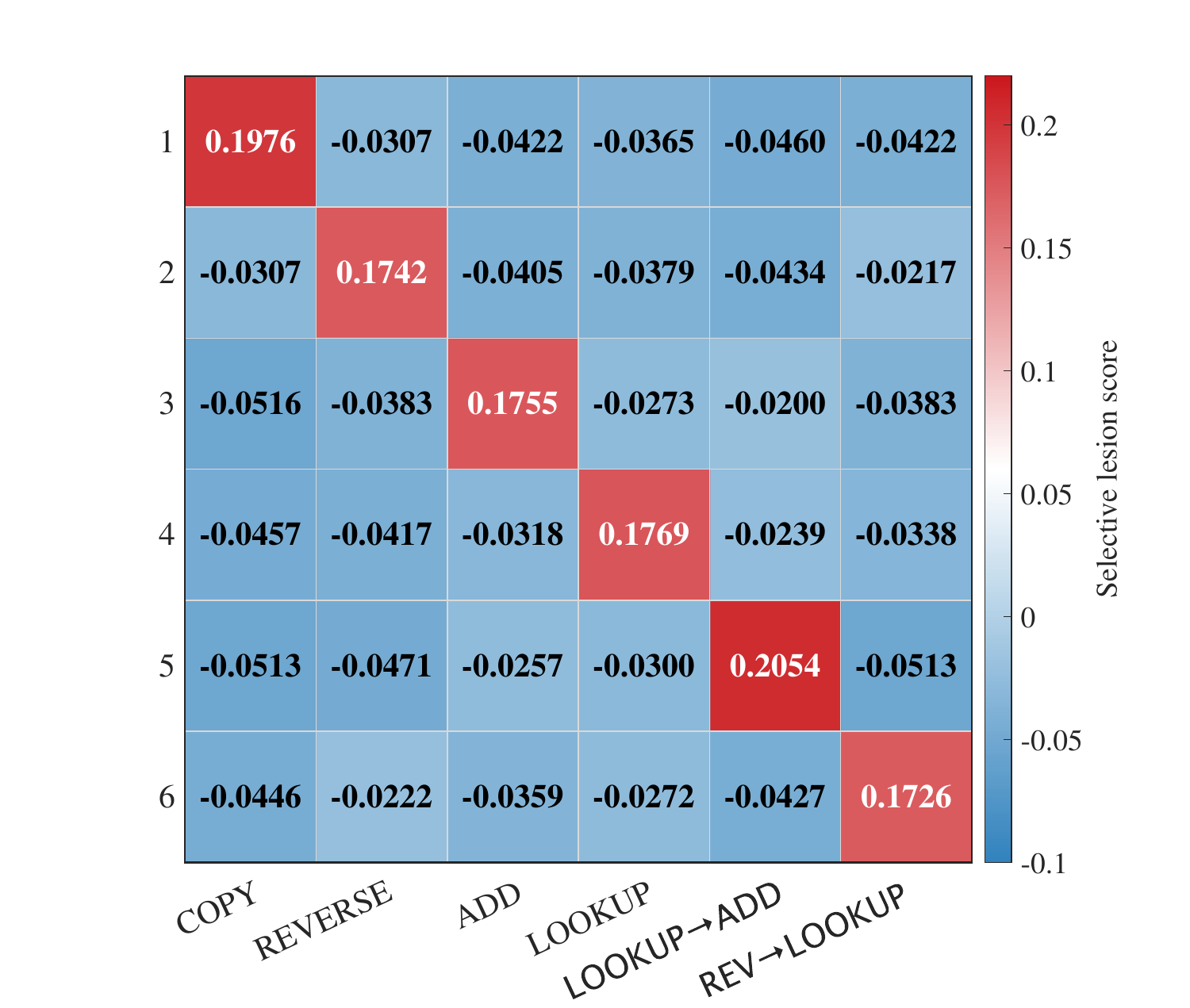}
        {\tiny (a) }
    \end{minipage}
    \begin{minipage}[t]{0.3\textwidth}
        \centering
        \includegraphics[width=\linewidth]{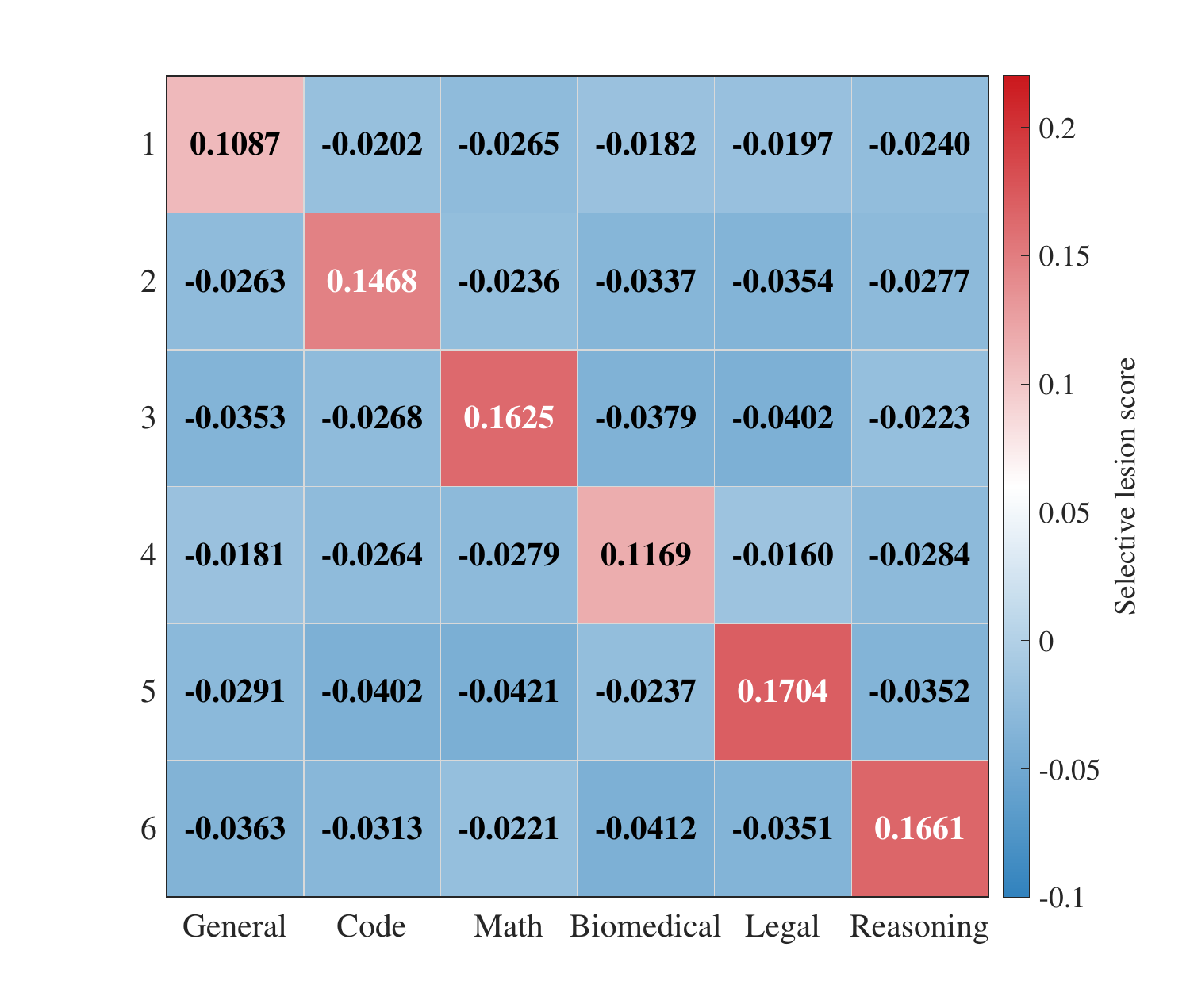}
        {\tiny (b) }
    \end{minipage}
    \begin{minipage}[t]{0.3\textwidth}
        \centering
        \includegraphics[width=\linewidth]{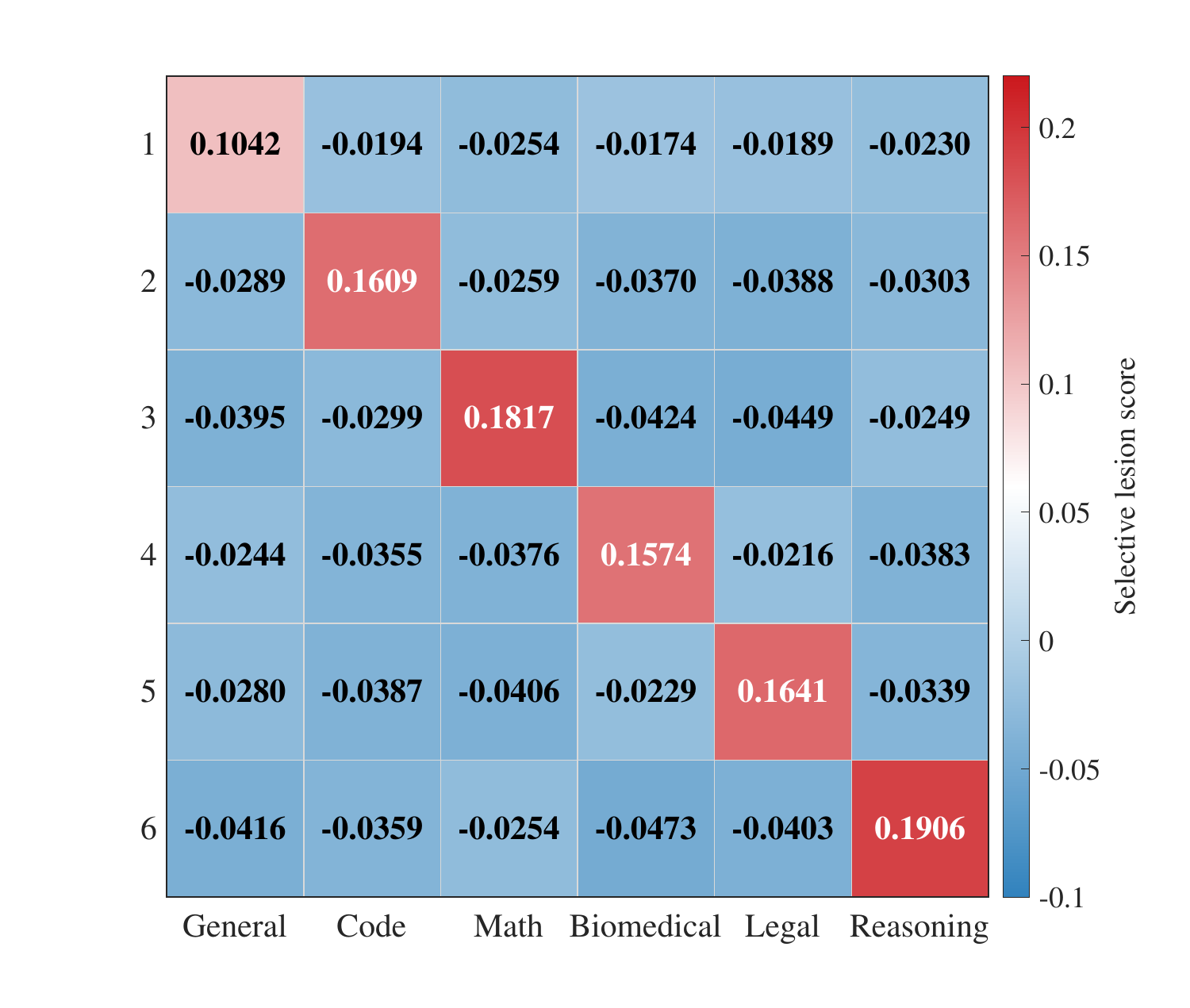}
        {\tiny (c) }
    \end{minipage}

    \caption{
    Selective lesion scores ($\mathrm{SLS}$) of different modules on (a) 160M model, (b) Qwen3-8B, and (c) Qwen3-32B. 
    Each entry measures the difference between the risk change on a domain and the average change on the remaining domains after removing a module.}
    \label{fig:mechanistic_sls}
\end{figure}

\paragraph{Shared module analysis}
We investigate the effect of the shared module in CORTEX using the 160M backbone on the synthetic benchmark.
Table \ref{tab:shared_ablation_synthetic} shows that introducing the shared module reduces PPL and NLL and improves EM when compared with CORTEX without the shared module.
It also increases $I^{\mathrm{mod}}$ and $\mathrm{SLS}$ across all six knowledge domains, indicating stronger domain alignment and selectivity.
These results support the shared module design for improving prediction performance and functional specialization. 

\begin{wraptable}{r}{0.48\textwidth}
\begingroup
\captionsetup{skip=3pt}
\centering
\tiny
\color{black}
\caption{\color{black}NLL, $I^{\mathrm{mod}}$, and lesion effects. \textbf{Target} and \textbf{Other} denote the average NLL changes after module removal on the paired domain and other domains, respectively. Results are averaged across seeds.}
\label{tab:nll_mi}
\setlength{\tabcolsep}{3.5pt}
\renewcommand{\arraystretch}{1.05}
\begin{tabular}{@{}llcccc@{}}
\toprule
\multirow{2}{*}{\textbf{Backbone}}
& \multirow{2}{*}{\textbf{Method}}
& \multirow{2}{*}{\textbf{NLL} $\downarrow$}
& \multirow{2}{*}{$I^{\mathrm{mod}}$}
& \multicolumn{2}{c}{$\Delta^{\mathrm{les}}_{m,k}$} \\
\cmidrule(lr){5-6}
& & & & \textbf{Target} & \textbf{Other} \\
\midrule
\rowcolor{grey!5}
& Random          & 0.4082 & 1.3860 & 0.0097 & 0.0086 \\
\rowcolor{green!12}
\multirow{-2}{*}{160M}
& \textbf{CORTEX}
& 0.0734 & 1.4560 & 0.1901 & 0.0064 \\
\midrule
\rowcolor{grey!5}
& Random          & 1.9179 & 1.0238 & 0.0374 & 0.0108 \\
\rowcolor{green!12}
\multirow{-2}{*}{Qwen3-8B}
& \textbf{CORTEX}
& 1.7303 & 1.1534 & 0.1557 & 0.0106 \\
\midrule
\rowcolor{grey!5}
& Random          & 1.6571 & 1.1996 & 0.0298 & 0.0097 \\
\rowcolor{green!12}
\multirow{-2}{*}{Qwen3-32B}
& \textbf{CORTEX}
& 1.4786 & 1.3331 & 0.1657 & 0.0059 \\
\bottomrule
\end{tabular}
\par\smallskip
\centering
\tiny
\color{black}
\caption{\color{black}Assignment concentration of CORTEX, averaged across seeds.}
\label{tab:assignment_concentration}
\setlength{\tabcolsep}{5pt}
\renewcommand{\arraystretch}{1.05}
\begin{tabular}{@{}lcc@{}}
\toprule
\textbf{Backbone} & \textbf{Average max. weight} & \textbf{Average normalized entropy} \\
\midrule
160M & $0.8755 $ & $0.2783$ \\
Qwen3-8B & $0.8429 $ & $0.3542$ \\
Qwen3-32B & $0.9001$ & $0.2581$ \\
\bottomrule
\end{tabular}
\par\endgroup
\end{wraptable}
\paragraph{Modularization analysis}
\textcolor{black}{Fig.~\ref{fig:mechanistic_sls} shows positive diagonal SLS values across all three backbones.
Appendix~\ref{sec:additional_results} compares learned and shuffled assignments with the same norm of removed weights. The learned assignments achieve higher paired-domain SLS on both 160M and Qwen3-8B.
}
\textcolor{black}{Table \ref{tab:nll_mi} shows that CORTEX achieves lower NLL
and higher module-domain mutual information than Random
across all three backbones.
Removing a CORTEX module on its paired domain causes a larger NLL increase, while the average increase on
the other domains remains comparable or smaller.
Table~\ref{tab:assignment_concentration} reports the average maximum assignment weight and average normalized entropy. They are calculated as $\frac{\sum_{j,g}\max_{0\le m\le M}s_{j,g,m}}{\sum_j|\mathcal{G}_j|}$ and $-\frac{\sum_{j,g}\sum_{m=0}^{M}s_{j,g,m}\ln s_{j,g,m}}{\ln(M+1)\sum_j|\mathcal{G}_j|}$, respectively. 
$\ln(M+1)$ is the maximum entropy over the shared and specialized modules.
Results show that CORTEX achieves high average maximum weights and low average normalized entropy across all three backbones, which indicates concentrated soft module assignments.}
\par

\section{Conclusion}
In this work, we proposed CORTEX to form shared and specialized modules within dense language models.
CORTEX keeps the original dense parameter space and divides trainable matrices into parameter groups.
It learns module assignment weights from domain-conditioned gradients while preserving the dense architecture.
We further introduced two module-level interpretability metrics to evaluate whether the modules are functionally specialized and aligned with knowledge domains. Our theoretical analysis relates module assignment weights to the convergence and connects the proposed metrics to modularization.
Results on the synthetic and real knowledge-domain mixtures demonstrate that CORTEX reduces NLL and PPL and learns domain-specific module assignments while preserving the dense architecture.

\subsection*{AI use statement}
\label{sec:llm_usage}
We used LLMs to assist with language polishing, grammar correction, and improving the clarity of the manuscript. 
For the synthetic benchmark, LLMs were used only to draft candidate task templates. 
The final task definitions, labels, data splits, and all synthetic instances were produced by the deterministic rule-based generator described in Appendix \ref{sec:synthetic}.

\subsection*{Ethics statement}
\label{sec:broader_impact}
This work provides a framework for forming and evaluating internal modularization in dense language models. 
On the positive side, CORTEX can improve transparency by making shared and domain-specific computation more distinctive, and may support more targeted adaptation and diagnosis across different knowledge domains. 
At the same time, stronger specialization can also improve knowledge-specific capabilities in areas such as code, biomedical text, legal text, or security-related content. 
If knowledge-domain labels or training data contain biases, the learned modules may also reflect or reinforce these biases. 
Therefore, applications of CORTEX to high-risk data or models should include careful data governance, privacy protection, bias evaluation, and appropriate safeguards.

\subsection*{Reproducibility statement}
The algorithm of CORTEX is presented in Appendix~\ref{sec:algorithm}.
The assumptions and proofs are provided in Appendix~\ref{sec:proof}.
Hyperparameter settings, dataset preparation, domain-conditioned gradient calculation, and downstream evaluation protocols are described in Appendix~\ref{sec:additional_results}.

\bibliographystyle{iclr2027_conference}

\newpage
\appendix

\section{List of Key Notations}
\label{sec:notation_table}
Table~\ref{tab:notation} summarizes the key notations used throughout the paper. 

\begin{table}[h]
\centering
\scriptsize
\renewcommand{\arraystretch}{1}
\caption{List of key notations.}
\label{tab:notation}
\begin{tabularx}{\textwidth}{p{0.24\textwidth}X}
\toprule
\textbf{Notation} & \textbf{Meaning} \\
\midrule

\(C\) 
& Knowledge-domain label. \\

\(d_j^{\mathrm{in}}, d_j^{\mathrm{out}}\) 
& Input and output dimensions of the \(j\)-th trainable weight matrix. \\

\(\mathcal{G}_j, G_j\) 
& Index set and the number of  parameter groups in \(\mathbf{W}_j\). \\

\(\mathbf{G}_{j,g,k}\) 
& Domain-conditioned gradient of the \(g\)-th parameter group in the \(j\)-th trainable parameter matrix w.r.t. the \(k\)-th knowledge domain. \\

\(\mathcal{J}, J\) 
& Set and the number of trainable weight matrices in the backbone model. \\

\(\mathcal{K}, K\) 
& Set and the number of knowledge domains.\\

\(\mathcal{L}_{\mathrm{main}}, \mathcal{L}_{\mathrm{assign}}\) 
& Main task loss and module assignment loss.\\

\(\mathcal{P}, \mathcal{P}_{k}\) 
& Overall training and conditional distribution of \((X,Y)\) given \(C=k\).\\

\(\mathbf{s}_{j,g}, s_{j,g,m}\) 
& Assignment weight vector and assignment weight of the \(g\)-th parameter group in \(\mathbf{W}_j\) assigned to the $m$-th module. \\

\(u_{j,g,0}, u_{j,g,m}\) 
& Utility scores of the \(g\)-th parameter group in \(\mathbf{W}_j\) for the shared module and the \(m\)-th specialized module. \\

\(\mathbf{W}, \mathbf{W}_j, \mathbf{W}_{j,m}\) 
& Trainable model parameters, the \(j\)-th trainable weight matrix, and its module-specific component associated with module \(m\). \\

\(X,Y,\mathcal{X},\mathcal{Y}\) 
& Input and output variables, together with their corresponding spaces. \\

\(\eta\) 
& Learning rate. \\

\(\boldsymbol{\xi}, \boldsymbol{\xi}_k\) 
& Mini-batch of training samples and its subset from the \(k\)-th knowledge domain. \\

\(\pi_k\) 
& Probability that a randomly drawn sample belongs to the \(k\)-th domain. \\

\(\Pi_{j,g}\) 
& Projection operator that extracts the \(g\)-th  parameter group from \(\mathbf{W}_j\). \\

\(\boldsymbol{\psi}_{j,g}, \psi_{j,g,m}, \Psi\) 
& Assignment parameter vector, its \(m\)-th entry, and the set of all trainable assignment parameter vectors. \\

\bottomrule
\end{tabularx}
\end{table}

\section{Algorithm of CORTEX}
\label{sec:algorithm}

\begin{algorithm}[H]
\caption{\textsc{CORTEX}}
\label{alg:cortex}
\small
\begin{algorithmic}[1]
\REQUIRE Training mixture \(\mathcal{P}=\sum_{k\in\mathcal{K}}\pi_k\mathcal{P}_k\); trainable matrices \(\mathcal{W}=\{\mathbf{W}_j\}_{j\in\mathcal{J}}\); parameter-group projections \(\{\Pi_{j,g}\}_{j,g}\); assignment parameters \(\Psi=\{\boldsymbol{\psi}_{j,g}\}_{j,g}\); learning rate.

\STATE Initialize \(\mathcal{W}^{1}\) from the dense backbone and initialize \(\Psi^{1}\).

\STATE \textcolor{black}{Calculate module assignment weights: $s^{1}_{j,g,m} = \frac{\exp(\psi^{1}_{j,g,m})}{\sum_{r\in\{0\}\cup\mathcal{M}}\exp(\psi^{1}_{j,g,r})}, \quad j\in\mathcal{J},\ g\in\mathcal{G}_{j}, m\in\{0\}\cup\mathcal{M}$.}

\FOR{\(t=1,\ldots,T\)}

    \STATE Sample mini-batches covering all knowledge domains.

    \CORTEXStage{domain-conditioned group gradients}
    \FOR{each \(k\in\mathcal{K}\)}
        \STATE Calculate the domain-conditioned group gradients: $\mathbf{G}^{t}_{j,g,k}
        =
        \Pi_{j,g}\!\left(
        \nabla_{\mathbf{W}_{j}}F(\mathcal{W}^{t};\boldsymbol{\xi}_{k})
        \right),
        \quad j\in\mathcal{J},\ g\in\mathcal{G}_{j}$.
    \ENDFOR

    \CORTEXStage{utility scores}
    \STATE Calculate the utility scores:
    \begin{align}
       u^{t}_{j,g,0} =&\: \frac{2}{K(K-1)}
    \sum_{1\leq p<q\leq K}
    \frac{
    \langle \mathbf{G}^{t}_{j,g,p},\mathbf{G}^{t}_{j,g,q}\rangle_F
    }{
    \|\mathbf{G}^{t}_{j,g,p}\|_F
    \|\mathbf{G}^{t}_{j,g,q}\|_F
    }, \quad j\in\mathcal{J},\ g\in\mathcal{G}_{j},\nonumber\\
    u^{t}_{j,g,m}
    =&\:
    \frac{
    \|\mathbf{G}^{t}_{j,g,\kappa(m)}\|_F
    }{
    \sum_{k'\in\mathcal{K}}\|\mathbf{G}^{t}_{j,g,k'}\|_F
    }
    -
    \frac{1}{K-1}
    \sum_{q\in\mathcal{K}\setminus\{\kappa(m)\}}
    \frac{
    \langle \mathbf{G}^{t}_{j,g,\kappa(m)},\mathbf{G}^{t}_{j,g,q}\rangle_F
    }{
    \|\mathbf{G}^{t}_{j,g,\kappa(m)}\|_F
    \|\mathbf{G}^{t}_{j,g,q}\|_F}, \nonumber\\
    &\quad  j\in\mathcal{J},\ g\in\mathcal{G}_{j}, m\in\mathcal{M}. \nonumber
    \end{align}

    \CORTEXStage{module assignment update}

    \STATE Form the assignment loss: $\mathcal{L}^{t}_{\mathrm{assign}}
    \leftarrow
    -\frac{1}{\sum_{j\in\mathcal{J}}G_j}
    \sum_{j\in\mathcal{J}}
    \sum_{g\in\mathcal{G}_{j}}
    \sum_{m\in\{0\}\cup\mathcal{M}}
    s^{t}_{j,g,m}u^{t}_{j,g,m}$.

    \STATE Update $\Psi^{t+1}$
    and update \(s^{t+1}_{j,g,m}\) from \(\Psi^{t+1}\).

    \CORTEXStage{modularized dense-model update}
    \FOR{each \(j\in\mathcal{J}\)}
        \STATE Update $\mathbf{W}^{t+1}_{j} = 
        \mathbf{W}^{t}_{j}
        -
        \eta
        \sum_{g\in\mathcal{G}_{j}}
        \left(
        \textcolor{black}{s^{t}_{j,g,0}}
        \sum_{k\in\mathcal{K}}\pi_k\mathbf{G}^{t}_{j,g,k}
        +
        \sum_{m\in\mathcal{M}}
        \pi_{\kappa(m)}
        \textcolor{black}{s^{t}_{j,g,m}}
        \mathbf{G}^{t}_{j,g,\kappa(m)}
        \right)$.
    \ENDFOR
\ENDFOR



\ENSURE Trained model \(\mathcal{W}^{\mathrm{out}}\).
\end{algorithmic}
\end{algorithm}

\section{Proofs}
\label{sec:proof}

\subsection{Proof of Proposition 1}

For the $k$-th knowledge domain, we consider the change of the loss
\(F(\mathcal{W};\boldsymbol{\xi}_k)\) caused by updating the $g$-th parameter group in the $j$-th
trainable matrix. 
The corresponding update is
\begin{equation}
\label{eq:prop1_update}
\mathbf{W}'_j-\mathbf{W}_j
=
-\eta\sum_{q\in\mathcal K}\pi_q\mathbf{G}_{j,g,q},
\end{equation}
where $\mathcal{W}'$ denotes the model after applying this update,
with all other matrices unchanged.
By applying the first-order Taylor expansion at $\mathcal{W}$
and substituting this update, we have
\begin{align}
\label{eq:prop1_taylor}
F(\mathcal{W}';\boldsymbol{\xi}_k)
-F(\mathcal{W};\boldsymbol{\xi}_k)
=&\:
\left\langle
\nabla_{\mathbf{W}_j}F(\mathcal{W};\boldsymbol{\xi}_k),
\mathbf{W}'_j-\mathbf{W}_j
\right\rangle_F
+\mathcal{O}\!\left(
\|\mathbf{W}'_j-\mathbf{W}_j\|_F^2
\right)
\nonumber\\
=&\:
-\eta\left\langle
\nabla_{\mathbf{W}_j}F(\mathcal{W};\boldsymbol{\xi}_k),
\sum_{q\in\mathcal K}\pi_q\mathbf{G}_{j,g,q}
\right\rangle_F
+\mathcal{O}(\eta^2),
\end{align}
where $\mathcal{O}(\eta^2)$ denotes a remainder whose absolute
value is bounded by a constant times $\eta^2$ as $\eta\to0$,
with the current model and mini-batch fixed.
Since $\mathbf{G}_{j,g,k}=\Pi_{j,g}\big(\nabla_{\mathbf{W}_{j}}F(\mathcal{W};\boldsymbol{\xi}_k)\big)$, we have
\begin{align}
    \left\langle
\nabla_{\mathbf{W}_j}F(\mathcal{W};\boldsymbol{\xi}_k),\mathbf{G}_{j,g,q}
\right\rangle_F
=
\left\langle
\mathbf{G}_{j,g,k},\mathbf{G}_{j,g,q}
\right\rangle_F.
\end{align}

Therefore,
\begin{align}
F(\mathcal{W}';\boldsymbol{\xi}_k)-F(\mathcal{W};\boldsymbol{\xi}_k) =&\: -\eta
\sum_{q\in\mathcal K}
\pi_q
\left\langle
\mathbf{G}_{j,g,k},\mathbf{G}_{j,g,q}
\right\rangle_F
+ \mathcal{O}(\eta^2) \nonumber\\
=&\: -\eta
\left(
\pi_k\|\mathbf{G}_{j,g,k}\|_F^2
+
\sum_{q\in\mathcal K\setminus\{k\}}
\pi_q
\left\langle
\mathbf{G}_{j,g,k},\mathbf{G}_{j,g,q}
\right\rangle_F
\right)
+\mathcal{O}(\eta^2).
\end{align}

This completes the proof of Proposition 1.

\subsection{Proof of Proposition 2}
Since the mixture of knowledge domains satisfies $\mathcal{P} = \sum_{k\in\mathcal{K}}\pi_{k}\mathcal{P}_{k}$, the main task loss can be decomposed as
\begin{align}
\label{eq:proof2_eq1}
    \mathcal{L}_{\mathrm{main}}(\mathcal{W}) = \mathbb{E}_{\boldsymbol{\xi}\sim\mathcal{P}}\left[F(\mathcal{W}; \boldsymbol{\xi})\right] = \sum_{k\in\mathcal{K}}\pi_{k}R_{k}(\mathcal{W}).
\end{align}

By applying eqn. (\ref{eq:proof2_eq1}) to both $\mathcal{W}^{\mathrm{out}}$ and $\mathcal{W}^{\mathrm{drop},m}$, we have

\begin{align}
    \label{eq:proof2_eq2}
    \mathcal{L}_{\mathrm{main}}(\mathcal{W}^{\mathrm{drop},m}) - \mathcal{L}_{\mathrm{main}}(\mathcal{W}^{\mathrm{out}}) =&\: \sum_{k\in\mathcal{K}}\pi_{k}\left(R_{k}(\mathcal{W}^{\mathrm{drop},m}) - R_{k}(\mathcal{W}^{\mathrm{out}})\right) \nonumber\\
    \overset{(\text{a})}{=}&\: \sum_{k\in\mathcal{K}}\pi_{k}\Delta^{\mathrm{les}}_{m,k},
\end{align}
where equality (a) results from Definition 3.
Based on Definition 3, we have
\begin{align}
    \label{eq:proof2_eq3}
    \Delta^{\mathrm{les}}_{m,k} = \mathrm{SLS}(m,k) + \frac{1}{K-1}\sum_{q\in\mathcal{K}\backslash\{k\}}\Delta^{\mathrm{les}}_{m,q}.
\end{align}

By combining eqns. (\ref{eq:proof2_eq2}) and (\ref{eq:proof2_eq3}), we have
\begin{align}
    &\mathcal{L}_{\mathrm{main}}(\mathcal{W}^{\mathrm{drop},m}) - \mathcal{L}_{\mathrm{main}}(\mathcal{W}^{\mathrm{out}}) \nonumber\\
    =&\: \pi_{k}\Delta^{\mathrm{les}}_{m,k} + \sum_{q\in\mathcal{K}\backslash\{k\}}\pi_{q}\Delta^{\mathrm{les}}_{m,q} \nonumber\\
    =&\: \pi_{k}\left(\mathrm{SLS}(m,k) + \frac{1}{K-1}\sum_{q\in\mathcal{K}\backslash\{k\}}\Delta^{\mathrm{les}}_{m,q}\right) + \sum_{q\in\mathcal{K}\backslash\{k\}}\pi_{q}\Delta^{\mathrm{les}}_{m,q} \nonumber\\
    =&\: \pi_{k}\mathrm{SLS}(m,k) + \sum_{q\in\mathcal{K}\backslash\{k\}}\left(\pi_{q} + \frac{\pi_{k}}{K-1}\right)\Delta^{\mathrm{les}}_{m,q}.
\end{align}

Therefore, $\mathcal{L}_{\mathrm{main}}(\mathcal{W}^{\mathrm{drop},m}) > \mathcal{L}_{\mathrm{main}}(\mathcal{W}^{\mathrm{out}})$ holds if
\begin{align}
    \label{eq:proof2_eq4}
    \pi_{k}\mathrm{SLS}(m,k) + \sum_{q\in\mathcal{K}\backslash\{k\}}\left(\pi_{q} + \frac{\pi_{k}}{K-1}\right)\Delta^{\mathrm{les}}_{m,q} > 0.
\end{align}
Since $\pi_{k} > 0$, inequality (\ref{eq:proof2_eq4}) is equivalent to
\begin{align}
    \mathrm{SLS}(m,k) > -\frac{1}{\pi_{k}}\sum_{q\in\mathcal{K}\backslash\{k\}}\left(\pi_{q} + \frac{\pi_{k}}{K-1}\right)\Delta^{\mathrm{les}}_{m,q}.
\end{align}

This completes the proof of Proposition 2.

\subsection{Proof of Proposition 3}
Based on Definition 4, $I^{\mathrm{mod}}$ satisfies
\begin{align}
    I^{\mathrm{mod}} =&\: \sum_{m\in\{0\}\cup\mathcal{M}} \sum_{k\in\mathcal{K}}\mathrm{GS}_{m,k}\ln\frac{\frac{\mathrm{GS}_{m,k}}{\mathrm{GS}_{m}^{\mathrm{module}}}}{\mathrm{GS}_{k}^{\mathrm{domain}}} \nonumber\\
    =&\: \sum_{m\in\{0\}\cup\mathcal{M}} \sum_{k\in\mathcal{K}}\mathrm{GS}_{m,k}\ln\frac{\mathrm{GS}_{m,k}}{\mathrm{GS}_{m}^{\mathrm{module}}} - \sum_{m\in\{0\}\cup\mathcal{M}} \sum_{k\in\mathcal{K}}\mathrm{GS}_{m,k}\ln\mathrm{GS}_{k}^{\mathrm{domain}} \nonumber\\
    =&\: \sum_{m\in\{0\}\cup\mathcal{M}} \sum_{k\in\mathcal{K}}\mathrm{GS}_{m,k}\ln\frac{\mathrm{GS}_{m,k}}{\mathrm{GS}_{m}^{\mathrm{module}}} -  \sum_{k\in\mathcal{K}}\left(\sum_{m\in\{0\}\cup\mathcal{M}}\mathrm{GS}_{m,k}\right)\ln\mathrm{GS}_{k}^{\mathrm{domain}} \nonumber\\
    \overset{(\text{a})}{=}&\: \sum_{m\in\{0\}\cup\mathcal{M}} \sum_{k\in\mathcal{K}}\mathrm{GS}_{m,k}\ln\frac{\mathrm{GS}_{m,k}}{\mathrm{GS}_{m}^{\mathrm{module}}} -  \sum_{k\in\mathcal{K}}\mathrm{GS}_{k}^{\mathrm{domain}}\ln\mathrm{GS}_{k}^{\mathrm{domain}} \nonumber\\
    =&\:\sum_{m\in\{0\}\cup\mathcal{M}}\mathrm{GS}_{m}^{\mathrm{module}} \sum_{k\in\mathcal{K}}\frac{\mathrm{GS}_{m,k}}{\mathrm{GS}_{m}^{\mathrm{module}}}\ln\frac{\mathrm{GS}_{m,k}}{\mathrm{GS}_{m}^{\mathrm{module}}} -  \sum_{k\in\mathcal{K}}\mathrm{GS}_{k}^{\mathrm{domain}}\ln\mathrm{GS}_{k}^{\mathrm{domain}} \nonumber\\
    =&\: -\sum_{m\in\{0\}\cup\mathcal{M}}\mathrm{GS}_{m}^{\mathrm{module}} \left(-\sum_{k\in\mathcal{K}}\frac{\mathrm{GS}_{m,k}}{\mathrm{GS}_{m}^{\mathrm{module}}}\ln\frac{\mathrm{GS}_{m,k}}{\mathrm{GS}_{m}^{\mathrm{module}}}\right) -  \sum_{k\in\mathcal{K}}\mathrm{GS}_{k}^{\mathrm{domain}}\ln\mathrm{GS}_{k}^{\mathrm{domain}},
\end{align}
where equality (a) follows from Definition 4.
Based on Definition 4, $\frac{\mathrm{GS}_{m,k}}{\mathrm{GS}_{m}^{\mathrm{module}}}\in[0,1]$.
Hence, we have
\begin{align}
    -\sum_{k\in\mathcal{K}}\frac{\mathrm{GS}_{m,k}}{\mathrm{GS}_{m}^{\mathrm{module}}}\ln\frac{\mathrm{GS}_{m,k}}{\mathrm{GS}_{m}^{\mathrm{module}}} \geq 0.
\end{align}

Therefore, $I^{\mathrm{mod}}$ satisfies
\begin{align}
    I^{\mathrm{mod}}\leq -\sum_{k\in\mathcal{K}}\mathrm{GS}_{k}^{\mathrm{domain}}\ln\mathrm{GS}_{k}^{\mathrm{domain}} \overset{(\text{a})}{\leq} \ln K,
\end{align}
where inequality (a) holds due to the fact that $\mathrm{GS}_{k}^{\mathrm{domain}}$ is a distribution over $K$ knowledge domains.
The equality $I^{\mathrm{mod}} = -\sum_{k\in\mathcal{K}}\mathrm{GS}_{k}^{\mathrm{domain}}\ln\mathrm{GS}_{k}^{\mathrm{domain}}$ holds if $-\sum_{k\in\mathcal{K}}\frac{\mathrm{GS}_{m,k}}{\mathrm{GS}_{m}^{\mathrm{module}}}\ln\frac{\mathrm{GS}_{m,k}}{\mathrm{GS}_{m}^{\mathrm{module}}} = 0$.
It means each module obtains the gradient from at most one knowledge domain.
This completes the proof of Proposition 3.

\subsection{Proof of Theorem 1}
\label{sec:proof_th1}
Expectations without conditioning are taken over all mini-batches
sampled during training.
Conditional expectations are taken given the model parameters
$\mathbf{W}^{t}$ and assignment parameters $\Psi^{t}$
before sampling.
We denote the update direction in the $t$-th training round as 
\begin{align}
    \label{eq:delta}
    \boldsymbol{\Phi}_{j}^{t} = \sum_{g\in\mathcal{G}_{j}}\left(s_{j,g,0}^{t}\sum_{k\in\mathcal{K}}\pi_{k}\mathbf{G}_{j,g,k}^{t} + \sum_{m\in\mathcal{M}}\pi_{\kappa(m)}s_{j,g,m}^{t}\mathbf{G}_{j,g,\kappa(m)}^{t}\right), \quad j\in\mathcal{J}.
\end{align}
First, we introduce two assumptions which are widely used to facilitate our proof.

\noindent\textbf{Assumption 1} ($L$-Smoothness) \textit{$\mathcal{L}_{\mathrm{main}}(\mathbf{W})$ is lower-bounded by $\mathcal{L}_{\mathrm{main}}(\mathbf{W}^{\mathrm{\star}})$ and $L$-smooth.
That is, for two arbitrary matrices $\mathbf{W}$ and $\mathbf{W}'$, we have}
\begin{align}
    \mathcal{L}_{\mathrm{main}}(\mathbf{W}') \leq \mathcal{L}_{\mathrm{main}}(\mathbf{W}) + \langle\nabla\mathcal{L}_{\mathrm{main}}(\mathbf{W}), \mathbf{W}' - \mathbf{W}\rangle_{F} + \frac{L}{2}\Vert\mathbf{W}' - \mathbf{W}\Vert_{F}^{2}.
\end{align}

\noindent\textbf{Assumption 2} (Bounded gradient and variance)
\textit{For each knowledge domain $k$, the collected group gradient is conditionally unbiased:
$\mathbb{E}[\mathbf{G}_{j,g,k}^{t} \mid \mathbf{W}^{t},\Psi^{t}] = \bar{\mathbf{G}}_{j,g,k}^{t} = \Pi_{j,g}\left(\nabla_{\mathbf{W}_{j}}R_k(\mathbf{W}^{t})\right)$.
For all $t,j,g,k$, its conditional mean and variance satisfy}
\begin{align}
&\left\Vert\bar{\mathbf{G}}_{j,g,k}^{t}\right\Vert_{F}^{2} \leq G^{2}, \\
&\mathbb{E}\left[\left\Vert\mathbf{G}_{j,g,k}^{t} - \bar{\mathbf{G}}_{j,g,k}^{t}\right\Vert_{F}^{2}\,\middle|\,\mathbf{W}^{t},\Psi^{t}\right] \leq \sigma^{2}.
\end{align}

Then, we present the proof of Theorem 1 as follows.

\textbf{Proof of Theorem 1}
Based on the update rule and (\ref{eq:delta}), we have
\begin{align}
    \mathbf{W}_{j}^{t+1} = \mathbf{W}_{j}^{t} - \eta\boldsymbol{\Phi}_{j}^{t}, \quad j\in\mathcal{J}.
\end{align}
Based on Assumption 1, we have
\begin{align}
\label{eq:proof1_main}
    \mathcal{L}_{\mathrm{main}}(\mathbf{W}^{t+1})
    \leq&\: \mathcal{L}_{\mathrm{main}}(\mathbf{W}^{t}) + \left\langle\nabla\mathcal{L}_{\mathrm{main}}(\mathbf{W}^{t}), \mathbf{W}^{t+1} - \mathbf{W}^{t}\right\rangle_{F} + \frac{L}{2}\left\Vert\mathbf{W}^{t+1} - \mathbf{W}^{t}\right\Vert_{F}^{2} \nonumber\\
    =&\: \mathcal{L}_{\mathrm{main}}(\mathbf{W}^{t}) - \eta\underbrace{\sum_{j\in\mathcal{J}}\left\langle\nabla_{\mathbf{W}_{j}}\mathcal{L}_{\mathrm{main}}(\mathbf{W}^{t}), \boldsymbol{\Phi}_{j}^{t}\right\rangle_{F}}_{T_{1}} + \frac{L\eta^{2}}{2}\underbrace{\sum_{j\in\mathcal{J}}\left\Vert\boldsymbol{\Phi}_{j}^{t}\right\Vert_{F}^{2}}_{T_{2}}.
\end{align}

In particular, $T_{2}$ satisfies

\begin{align}
    \label{eq:T_2}
    T_{2} =&\: \sum_{j\in\mathcal{J}}\sum_{g\in\mathcal{G}_{j}}\left\Vert s_{j,g,0}^{t}\sum_{k\in\mathcal{K}}\pi_{k}\mathbf{G}_{j,g,k}^{t} + \sum_{m\in\mathcal{M}}\pi_{\kappa(m)}s_{j,g,m}^{t}\mathbf{G}_{j,g,\kappa(m)}^{t}\right\Vert_{F}^{2} \nonumber\\
    \overset{(\text{a})}{=}&\: \sum_{j\in\mathcal{J}}\sum_{g\in\mathcal{G}_{j}}\left\Vert \sum_{m\in\mathcal{M}}\pi_{\kappa(m)}\left(s_{j,g,0}^{t} + s_{j,g,m}^{t}\right)\mathbf{G}_{j,g,\kappa(m)}^{t}\right\Vert_{F}^{2},
\end{align}
where equality (a) follows from the one-to-one pairing between specialized modules and knowledge domains.
The weights $\pi_{\kappa(m)}(s_{j,g,0}^{t}+s_{j,g,m}^{t})$
are nonnegative.
We apply the Cauchy-Schwarz inequality to
$\sqrt{\pi_{\kappa(m)}(s_{j,g,0}^{t}+s_{j,g,m}^{t})}$
and
$\sqrt{\pi_{\kappa(m)}(s_{j,g,0}^{t}+s_{j,g,m}^{t})}
\mathbf{G}_{j,g,\kappa(m)}^{t}$
for each matrix entry, and then sum over all entries to obtain
\begin{align}
    \label{eq:proof1_CS}
    &\left\Vert \sum_{m\in\mathcal{M}}\pi_{\kappa(m)}\left(s_{j,g,0}^{t} + s_{j,g,m}^{t}\right)\mathbf{G}_{j,g,\kappa(m)}^{t}\right\Vert_{F}^{2} \nonumber\\
    \leq&\: \left(\sum_{m\in\mathcal{M}}\pi_{\kappa(m)}\left(s_{j,g,0}^{t} + s_{j,g,m}^{t}\right)\right)\left(\sum_{m\in\mathcal{M}}\pi_{\kappa(m)}\left(s_{j,g,0}^{t} + s_{j,g,m}^{t}\right)\left\Vert\mathbf{G}_{j,g,\kappa(m)}^{t}\right\Vert_{F}^{2}\right).
\end{align}

Taking total expectations on both sides of inequality
(\ref{eq:proof1_CS}), we have
\begin{align}
\label{eq:T_2_extend}
&\mathbb{E}\left[
\left\Vert
\sum_{m\in\mathcal{M}}
\pi_{\kappa(m)}
\left(s_{j,g,0}^{t}+s_{j,g,m}^{t}\right)
\mathbf{G}_{j,g,\kappa(m)}^{t}
\right\Vert_F^2
\right]
\nonumber\\
\overset{(\text{a})}{\leq}&\:
\mathbb{E}\Bigg[
\left(
\sum_{m\in\mathcal{M}}
\pi_{\kappa(m)}
\left(s_{j,g,0}^{t}+s_{j,g,m}^{t}\right)
\right)
\nonumber\\
&\quad\times
\sum_{m\in\mathcal{M}}
\pi_{\kappa(m)}
\left(s_{j,g,0}^{t}+s_{j,g,m}^{t}\right)
\mathbb{E}\left[
\left\Vert\mathbf{G}_{j,g,\kappa(m)}^{t}\right\Vert_F^2
\,\middle|\,\mathbf{W}^{t},\Psi^{t}
\right]
\Bigg]
\nonumber\\
\overset{(\text{b})}{\leq}&\:
2(G^2+\sigma^2)
\mathbb{E}\left[
\left(
\sum_{m\in\mathcal{M}}
\pi_{\kappa(m)}
\left(s_{j,g,0}^{t}+s_{j,g,m}^{t}\right)
\right)^2
\right]
\nonumber\\
=&\:
2(G^2+\sigma^2)
\mathbb{E}\left[
\left(
s_{j,g,0}^{t}
+\sum_{m\in\mathcal{M}}\pi_{\kappa(m)}s_{j,g,m}^{t}
\right)^2
\right].
\end{align}
Inequality (a) follows from (\ref{eq:proof1_CS})
by taking expectations over the conditional expectation,
since the assignment weights are fixed given
$\mathbf{W}^{t}$ and $\Psi^{t}$.
Inequality (b) follows from Assumption 2 and
$\|\mathbf{A}+\mathbf{B}\|_F^2
\leq 2\|\mathbf{A}\|_F^2+2\|\mathbf{B}\|_F^2$.
By combining eqn. (\ref{eq:T_2}) and inequality
(\ref{eq:T_2_extend}), we have
\begin{align}
\label{eq:T_2_final}
\mathbb{E}[T_2]
\leq
2(G^2+\sigma^2)
\sum_{j\in\mathcal{J}}\sum_{g\in\mathcal{G}_j}
\mathbb{E}\left[
\left(
s_{j,g,0}^{t}
+\sum_{m\in\mathcal{M}}\pi_{\kappa(m)}s_{j,g,m}^{t}
\right)^2
\right].
\end{align}

Then, we bound $\mathbb{E}[T_1]$.
We first condition on $\mathbf{W}^{t}$ and $\Psi^{t}$.
Note that $\nabla_{\mathbf{W}_{j}}\mathcal{L}_{\mathrm{main}}(\mathbf{W}^{t}) = \sum_{g\in\mathcal{G}_{j}}\sum_{m\in\mathcal{M}}\pi_{\kappa(m)}\bar{\mathbf{G}}_{j,g,\kappa(m)}^{t}$.
And we have
\begin{align}
    \mathbb{E}\left[\boldsymbol{\Phi}_{j}^{t}\,\middle|\,\mathbf{W}^{t},\Psi^{t}\right] 
    =&\:  \sum_{g\in\mathcal{G}_{j}}\sum_{m\in\mathcal{M}}\pi_{\kappa(m)}\left(s_{j,g,0}^{t}+s_{j,g,m}^{t}\right)\bar{\mathbf{G}}_{j,g,\kappa(m)}^{t} \nonumber\\
    =&\: \nabla_{\mathbf{W}_{j}}\mathcal{L}_{\mathrm{main}}(\mathbf{W}^{t}) - \sum_{g\in\mathcal{G}_{j}}\sum_{m\in\mathcal{M}}\pi_{\kappa(m)}\left(1 - s_{j,g,0}^{t} - s_{j,g,m}^{t}\right)\bar{\mathbf{G}}_{j,g,\kappa(m)}^{t}.
\end{align}

The squared norm of the full gradient satisfies
$\|\nabla\mathcal{L}_{\mathrm{main}}(\mathbf{W}^{t})\|_{F}^{2}
=
\sum_{j\in\mathcal{J}}
\|\nabla_{\mathbf{W}_{j}}
\mathcal{L}_{\mathrm{main}}(\mathbf{W}^{t})\|_{F}^{2}$.
Hence, $\mathbb{E}[T_{1} \mid \mathbf{W}^{t},\Psi^{t}]$ satisfies
\begin{align}
    \label{eq:T_1}
&\mathbb{E}[T_{1} \mid \mathbf{W}^{t},\Psi^{t}] \nonumber\\
=&\:  - \sum_{j\in\mathcal{J}}\left\langle\nabla_{\mathbf{W}_{j}}\mathcal{L}_{\mathrm{main}}(\mathbf{W}^{t}), \sum_{g\in\mathcal{G}_{j}}\sum_{m\in\mathcal{M}}\pi_{\kappa(m)}\left(1 - s_{j,g,0}^{t} - s_{j,g,m}^{t}\right)\bar{\mathbf{G}}_{j,g,\kappa(m)}^{t}\right\rangle_{F} \nonumber\\
    &+ \left\Vert\nabla\mathcal{L}_{\mathrm{main}}(\mathbf{W}^{t})\right\Vert_{F}^{2} \nonumber\\
    \overset{(\text{a})}{\geq}&\: \frac{1}{2}\left\Vert\nabla\mathcal{L}_{\mathrm{main}}(\mathbf{W}^{t})\right\Vert_{F}^{2} - \frac{1}{2}\underbrace{\sum_{j\in\mathcal{J}}\left\Vert\sum_{g\in\mathcal{G}_{j}}\sum_{m\in\mathcal{M}}\pi_{\kappa(m)}\left(1 - s_{j,g,0}^{t} - s_{j,g,m}^{t}\right)\bar{\mathbf{G}}_{j,g,\kappa(m)}^{t}\right\Vert_{F}^{2}}_{T_{3}},
\end{align}
where inequality (a) follows from $\textcolor{black}{\langle\mathbf{A}, \mathbf{B}\rangle_{F}} \leq \frac{1}{2}\Vert\mathbf{A}\Vert_{F}^{2} + \frac{1}{2}\Vert\mathbf{B}\Vert_{F}^{2}$.
Then, we bound $T_{3}$ as follows:
\begin{align}
    \label{eq:T_3}
    T_{3} 
    \overset{(\text{a})}{=}&\: \sum_{j\in\mathcal{J}}\sum_{g\in\mathcal{G}_{j}}\left\Vert\sum_{m\in\mathcal{M}}\pi_{\kappa(m)}\left(1 - s_{j,g,0}^{t} - s_{j,g,m}^{t}\right)\bar{\mathbf{G}}_{j,g,\kappa(m)}^{t}\right\Vert_{F}^{2} \nonumber\\
    \overset{(\text{b})}{\leq}&\: \sum_{j\in\mathcal{J}}\sum_{g\in\mathcal{G}_{j}}\left(\sum_{m\in\mathcal{M}}\pi_{\kappa(m)}\left(1 - s_{j,g,0}^{t} - s_{j,g,m}^{t}\right)^{2}\right)\sum_{m\in\mathcal{M}}\pi_{\kappa(m)}\left\Vert\bar{\mathbf{G}}_{j,g,\kappa(m)}^{t}\right\Vert_{F}^{2} \nonumber\\
    \overset{(c)}{\leq}&\: G^{2}\sum_{j\in\mathcal{J}}\sum_{g\in\mathcal{G}_{j}}\sum_{m\in\mathcal{M}}\pi_{\kappa(m)}\left(1 - s_{j,g,0}^{t} - s_{j,g,m}^{t}\right)^{2},
\end{align}

where equality (a) is obtained due to the fact that different parameter groups are disjoint as described in Definition 1.
Inequality (b) results from Cauchy-Schwarz inequality.
Inequality (c) follows from Assumption 2 and the fact that $\sum_{m\in\mathcal{M}}\pi_{\kappa(m)} = 1$.
By combining inequalities (\ref{eq:T_1}) and (\ref{eq:T_3})
and taking total expectations, we have
\begin{align}
\label{eq:T_1_final}
\mathbb{E}[T_1]
\geq&\:
\frac{1}{2}
\mathbb{E}\left[
\left\Vert
\nabla\mathcal{L}_{\mathrm{main}}(\mathbf{W}^{t})
\right\Vert_F^2
\right] -
\frac{G^2}{2}
\sum_{j\in\mathcal{J}}
\sum_{g\in\mathcal{G}_j}
\sum_{m\in\mathcal{M}}
\pi_{\kappa(m)}
\mathbb{E}\left[
\left(1-s_{j,g,0}^{t}-s_{j,g,m}^{t}\right)^2
\right].
\end{align}

Taking total expectations in (\ref{eq:proof1_main})
and applying (\ref{eq:T_2_final}) and (\ref{eq:T_1_final}),
we obtain
\begin{align}
\mathbb{E}\left[
\mathcal{L}_{\mathrm{main}}(\mathbf{W}^{t+1})
\right]
\leq&\:
\mathbb{E}\left[
\mathcal{L}_{\mathrm{main}}(\mathbf{W}^{t})
\right]
-\frac{\eta}{2}
\mathbb{E}\left[
\left\Vert
\nabla\mathcal{L}_{\mathrm{main}}(\mathbf{W}^{t})
\right\Vert_F^2
\right]
\nonumber\\
&+
\frac{G^2\eta}{2}
\sum_{j\in\mathcal{J}}
\sum_{g\in\mathcal{G}_j}
\sum_{m\in\mathcal{M}}
\pi_{\kappa(m)}
\mathbb{E}\left[
\left(1-s_{j,g,0}^{t}-s_{j,g,m}^{t}\right)^2
\right]
\nonumber\\
&+
L\eta^2(G^2+\sigma^2)
\sum_{j\in\mathcal{J}}
\sum_{g\in\mathcal{G}_j}
\mathbb{E}\left[
\left(
s_{j,g,0}^{t}
+\sum_{m\in\mathcal{M}}\pi_{\kappa(m)}s_{j,g,m}^{t}
\right)^2
\right].
\end{align}

Rearranging and summing over $T$ training rounds yields
\begin{align}
&\frac{\eta}{2}\sum_{t=1}^{T}\mathbb{E}\left[\left\Vert\nabla\mathcal{L}_{\mathrm{main}}(\mathbf{W}^{t})\right\Vert_{F}^{2}\right] \nonumber\\
\leq&\: \mathcal{L}_{\mathrm{main}}(\mathbf{W}^{1}) - \mathbb{E}\left[\mathcal{L}_{\mathrm{main}}(\mathbf{W}^{T+1})\right] \nonumber\\
&+ \frac{G^{2}\eta}{2}\sum_{t=1}^{T}\sum_{j\in\mathcal{J}}\sum_{g\in\mathcal{G}_{j}}\sum_{m\in\mathcal{M}}\pi_{\kappa(m)}\mathbb{E}\left[\left(1 - s_{j,g,0}^{t} - s_{j,g,m}^{t}\right)^{2}\right] \nonumber\\
&+ L\eta^{2}(G^{2}+\sigma^{2})\sum_{t=1}^{T}\sum_{j\in\mathcal{J}}\sum_{g\in\mathcal{G}_{j}}\mathbb{E}\left[\left(s_{j,g,0}^{t} + \sum_{m\in\mathcal{M}}\pi_{\kappa(m)}s_{j,g,m}^{t}\right)^{2}\right].
\end{align}

Note that $\mathcal{L}_{\mathrm{main}}$ is lower-bounded by $\mathcal{L}_{\mathrm{main}}(\mathbf{W}^{\star})$.
We multiply $\frac{2}{T\eta}$ on both sides and obtain
\begin{align}
&\frac{1}{T}\sum_{t=1}^{T}\mathbb{E}\left[\left\Vert\nabla\mathcal{L}_{\mathrm{main}}(\mathbf{W}^{t})\right\Vert_{F}^{2}\right] \nonumber\\
\leq&\: \frac{2}{T\eta}\left(\mathcal{L}_{\mathrm{main}}(\mathbf{W}^{1}) - \mathcal{L}_{\mathrm{main}}(\mathbf{W}^{\star})\right) + \frac{G^{2}}{T}\sum_{t=1}^{T}\sum_{j\in\mathcal{J}}\sum_{g\in\mathcal{G}_{j}}\sum_{m\in\mathcal{M}}\pi_{\kappa(m)}\mathbb{E}\left[A_{1,j,g,m}^{t}\right] \nonumber\\
&+ \frac{2L\eta(G^{2}+\sigma^{2})}{T}\sum_{t=1}^{T}\sum_{j\in\mathcal{J}}\sum_{g\in\mathcal{G}_{j}}\mathbb{E}\left[A_{2,j,g}^{t}\right].
\end{align}

This completes the proof of Theorem 1.

\begin{table}[t]
\centering
\scriptsize
\caption{Training and modularization hyperparameters for the synthetic benchmark.}
\label{tab:hyper_synthetic}
\setlength{\tabcolsep}{5pt}
\renewcommand{\arraystretch}{1.08}
\begin{tabular}{lc}
\toprule
\textbf{Hyperparameter} & \textbf{Synthetic 160M} \\
\midrule
Domains & 6 \\
Backbone & 160M decoder-only \\
Optimizer & AdamW \\
Selected learning rate & $3{\times}10^{-4}$ \\
Learning-rate candidates 
& $\{5{\times}10^{-4}, 4{\times}10^{-4}, 3{\times}10^{-4}, 2{\times}10^{-4}, 1{\times}10^{-4}\}$ \\
Mini-batch size / samples per update & 32 / 32 \\
Batch-size candidates & $\{8,16,32,64\}$ \\
Weight decay & 0.01 \\
Precision & bf16 \\
Seeds & 10 (2026--2035) \\
Total optimizer updates / epochs & 15360 / 20 \\
Warmup updates / validation interval & 768 / 256 \\
Specialized modules & 6 \\
Parameter grouping & \makecell{One attention head;\\128 FFN intermediate channels} \\
\bottomrule
\end{tabular}
\end{table}

\begin{table}[t]
\centering
\scriptsize
\caption{Training and modularization hyperparameters for Qwen3-8B.}
\label{tab:hyper_qwen3_8b}
\setlength{\tabcolsep}{5pt}
\renewcommand{\arraystretch}{1.08}
\begin{tabular}{lc}
\toprule
\textbf{Hyperparameter} & \textbf{Qwen3-8B} \\
\midrule
Domains & 6 \\
Backbone & Qwen3-8B \\
Optimizer & Adafactor \\
Selected learning rate & $5{\times}10^{-5}$ \\
Learning-rate candidates 
& \makecell{
$\{1{\times}10^{-4}, 9{\times}10^{-5}, 8{\times}10^{-5}, 7{\times}10^{-5}, 6{\times}10^{-5},$\\
$5{\times}10^{-5}, 4{\times}10^{-5}, 3{\times}10^{-5}, 2{\times}10^{-5}, 1{\times}10^{-5}\}$
} \\
Mini-batch size / samples per update & 1 / 48 \\
Precision & bf16 \\
Seeds & 4 (2026--2029) \\
Total optimizer updates / epochs & 5000 / 1 \\
Warmup updates / validation interval & 150 / 250 \\
Specialized modules & 6 \\
Parameter grouping & \makecell{One attention head;\\128 FFN intermediate channels} \\
\bottomrule
\end{tabular}
\end{table}

\begin{table}[!t]
\centering
\scriptsize
\caption{Training and modularization hyperparameters for Qwen3-32B.}
\label{tab:hyper_qwen3_32b}
\setlength{\tabcolsep}{5pt}
\renewcommand{\arraystretch}{1.08}
\begin{tabular}{lc}
\toprule
\textbf{Hyperparameter} & \textbf{Qwen3-32B} \\
\midrule
Domains & 6 \\
Backbone & Qwen3-32B \\
Optimizer & Adafactor \\
Selected learning rate & $2{\times}10^{-5}$ \\
Learning-rate candidates 
& \makecell{
$\{1{\times}10^{-4}, 9{\times}10^{-5}, 8{\times}10^{-5}, 7{\times}10^{-5}, 6{\times}10^{-5},$\\
$5{\times}10^{-5}, 4{\times}10^{-5}, 3{\times}10^{-5}, 2{\times}10^{-5}, 1{\times}10^{-5}\}$
} \\
Mini-batch size / samples per update & 1 / 48 \\
Precision & bf16 \\
Seeds & 4 (2026--2029) \\
Total optimizer updates / epochs & 5000 / 1 \\
Warmup updates / validation interval & 150 / 250 \\
Specialized modules & 6 \\
Parameter grouping & \makecell{One attention head;\\128 FFN intermediate channels} \\
\bottomrule
\end{tabular}
\end{table}

\section{Implementation Settings and Additional Results}
\label{sec:additional_results}
\subsection{Hyperparameter Settings}
We present the hyperparameter settings of the 160M model, Qwen3-8B, and Qwen3-32B in Tables \ref{tab:hyper_synthetic}, \ref{tab:hyper_qwen3_8b}, and \ref{tab:hyper_qwen3_32b}, respectively.
For each backbone, we select the learning rate using Dense validation NLL and use it for all methods.
We average validation NLL equally over the six knowledge domains.
The learning rate increases linearly during warmup and then follows cosine decay to zero.
Training uses the full update budget without early stopping.
We save checkpoints at each validation interval. Among checkpoints with the final model structure, we select the one with the lowest validation NLL. If multiple checkpoints have the same lowest NLL, we select the earliest one.
The same selected checkpoint is used for all downstream benchmarks.
For AdamW, we use $(\beta_1,\beta_2)=(0.9,0.999)$, $\epsilon=10^{-8}$, and gradient clipping at an $L_2$ norm of 1.
For Adafactor, we disable relative steps and parameter scaling. We also disable the first moment and external gradient clipping. We use $\epsilon=(10^{-30},10^{-3})$, decay rate $-0.8$, clipping threshold 1, and zero weight decay.

For CORTEX, we learn module assignments for the attention and FFN matrices.
For attention matrices, we divide the query, key, and value matrices along the output axis and the attention output matrix along the input axis.
A parameter group contains 64 channels for the 160M model and 128 channels for Qwen3-8B and Qwen3-32B, corresponding to one attention head.
For FFN matrices, a parameter group contains 128 intermediate channels.
The input, gate, and up matrices are divided along the output axis, while the output and down matrices are divided along the input axis.
We learn the module assignment parameters separately for each of these matrices.
If the remaining channels do not form a complete parameter group, they are kept as a smaller parameter group.
The other model parameters remain trainable and are updated using the mixture gradient.

The assignment parameters $\Psi$ are initialized to zero, with all initial module assignment weights equal to $1/7$.
We use Adam to update $\Psi$ by minimizing $\mathcal{L}_{\mathrm{assign}}$.
The learning rate is fixed at $10^{-3}$, with $(\beta_1,\beta_2)=(0.9,0.999)$, $\epsilon=10^{-8}$, and zero weight decay.
We update $\Psi$ once after gradients from all knowledge domains are available.
We use FP32 to store $\Psi$ and calculate the utility scores.
We calculate the gradient of $\mathcal{L}_{\mathrm{assign}}$ w.r.t. $\Psi$, with the utility scores fixed.

For \textbf{Random}, we independently sample seven assignment logits from $\mathcal{N}(0,1)$ for each parameter group. We apply softmax with temperature 1.
The random seed is the training seed plus 10000.
These assignments remain fixed throughout training.

For MoE, MoM, UpIT, and Self-MoE, we use top-1 activation.
MoE replaces each FFN with six experts and a linear router. We set the load-balancing loss coefficient to 0.01 and keep all tokens.
For Qwen3-8B and Qwen3-32B, each MoE expert is initialized from the corresponding FFN weights in the same pretrained checkpoint used by \textbf{Dense} and CORTEX.
The router is randomly initialized, and all other layers use the original checkpoint weights.

For MoM, we divide the backbone into sets of six blocks.
The 160M model, Qwen3-8B, and Qwen3-32B use 2, 6, and 10 sets, respectively. Qwen3-32B also uses four dense blocks.
We use separate routers for attention and FFN. Within each set, each token uses each attention module and each FFN module once over six steps.
MoM matches \textbf{Dense} in the number of backbone parameters and layers applied to each token. The routers introduce additional parameters.

For UpIT, we construct six FFN experts from equally spaced checkpoints during the first 60\% of training.
This stage uses 9216 updates for the 160M model and 3000 updates for each Qwen3 model.
We initialize the other parameters from the last of these checkpoints.
We then freeze the backbone and experts and train the router for the remaining 6144 and 2000 updates, respectively.
We first train the router for one pass on a subset of the training data. Each domain contributes 256 samples for the 160M model and 512 samples for each Qwen3 model.
This uses 48 and 64 updates, respectively, within the router training budget. The remaining numbers of updates are 6096 and 1936.

For Self-MoE, we use six LoRA experts with rank 8, alpha 16, and dropout 0.05.
LoRA is applied to the query, key, value, attention output, and FFN matrices.
We use the original training data without self-generated samples.
We train the dense backbone during the first 40\% of updates. We then fix the backbone and train the domain experts during the next 40\%. The remaining 20\% train the router.
Each domain uses 256 seed samples for the 160M model and 512 seed samples for each Qwen3 model.

\subsection{Dataset Preparation}
\label{sec:synthetic}
The synthetic benchmark is produced by a deterministic rule-based generator with fixed random seeds. 
We define six synthetic knowledge domains: COPY, REVERSE, ADD, LOOKUP, LOOKUP->ADD, and REVERSE->LOOKUP. 
For each example, the generator first samples a domain label from fixed domain priors, and then instantiates the corresponding symbolic task using a per-example pseudorandom seed. 
COPY samples a digit sequence and outputs the same sequence. 
REVERSE outputs the reversed sequence.
ADD samples two 2-digit or 3-digit integers and outputs their decimal sum.
LOOKUP maps a three-letter key to the weighted sum of its zero-based alphabet positions modulo 10, using weights $(1,2,4)$.
REVERSE->LOOKUP applies this rule to the reversed key; both tasks retain only keys whose outputs differ under reversal.
LOOKUP->ADD outputs the sum of two lookup values.
In our implementation, the default synthetic seed is 2026, with domain priors [0.16, 0.16, 0.16, 0.18, 0.17, 0.17].
For the synthetic benchmark, the train/validation/test sizes are 24576/3072/3072, with each of the six domains uniformly contributing 4096/512/512 samples, respectively.
We stop adding samples to a domain after reaching these sample counts. We remove duplicate input--target pairs across splits.
We calculate the loss on the answer and EOS tokens.
The domain mixture weights used during training follow the stated domain priors. 
All samples are generated by a deterministic rule-based generator and are truncated or padded to a maximum sequence length of 48 for the 160M decoder-only backbone.

For the real-domain benchmark, the train/validation/test sizes are 240000/7200/10200, with each of the six domains uniformly contributing 40000/1200/1700 samples, respectively.
All samples are truncated or padded to a maximum sequence length of 768 for both Qwen3-8B and Qwen3-32B.
We split the real-domain data using seed 2026 after removing duplicate source examples.
For WinoGrande, we split 41665 labeled source examples into 38765 training, 1200 validation, and 1700 test examples.
We exchange the two answer options in 1235 training examples and add the resulting examples to the training set. This produces 40000 training samples.
For general, biomedical, and legal text, we calculate the loss on the full text.
For code and mathematics, we calculate the loss on the answer and EOS tokens. For WinoGrande, we use the answer label.
All methods use the same tokenized samples. Each sequence contains one sample.

\subsection{Implementation of Domain-conditioned Group Gradients}
We clarify how the domain-conditioned group gradients in Section 3.2 are calculated in our implementation. 
{\color{black}The mini-batch size denotes the number of samples used for each gradient calculation. Each update uses 32 samples for the 160M model and 48 samples for each Qwen3 backbone. For the 160M model, each update uses five samples per domain and two additional samples. The two additional samples come from domain pairs $(0,1)$, $(2,3)$, and $(4,5)$ in turn. For both Qwen3 backbones, each update uses eight samples per domain. Qwen3-8B uses 12 mini-batches per GPU across 4 GPUs. Qwen3-32B uses 3 mini-batches per GPU across 16 GPUs. CORTEX collects domain-conditioned group gradients from all samples used for the update. These samples cover all six knowledge domains. During gradient collection, the backbone parameters remain fixed. For each knowledge domain, we weight each mini-batch gradient by the number of tokens used to calculate its loss and divide the sum by the total number of these tokens. Padding tokens and tokens excluded from the loss calculation are not counted. 
For each knowledge domain, Dense and CORTEX calculate the average loss over the same tokens.
For the real-domain benchmark, both methods use equal domain weights, i.e., $\pi_k=1/6$.
CORTEX further weights each domain-conditioned group gradient by the sum of the shared and corresponding specialized module assignment weights.

When the mini-batch size is one, each mini-batch contributes gradients to a single knowledge domain. After collection, CORTEX calculates the utility scores from the average domain-conditioned group gradients and updates $\Psi$ and the backbone parameters simultaneously. The domain-conditioned group gradients are combined before updating the backbone parameters. We use AdamW for the 160M model and Adafactor for Qwen3-8B and Qwen3-32B. The backbone optimizer uses one set of optimizer states for each parameter and performs one update after gradient collection. The accumulated gradients are cleared after each update.} All approaches use the same training-token budget and total number of optimizer updates. The learning-rate schedule is based on the number of optimizer updates. For UpIT and Self-MoE, this budget includes expert construction and router training.
Domain labels are used only for training and to report two module-level interpretability metrics. 
During inference, CORTEX uses the trained dense model directly.
The learned modules are soft components of the dense model rather than routed experts.
\textcolor{black}{For Table~\ref{tab:nll_mi}, we remove each specialized module separately from Random and CORTEX and calculate the NLL changes on its paired domain and the other domains.}

\subsection{Downstream Evaluation Protocol}
\label{sec:downstream_protocol}
We use the corresponding Qwen3 chat template with one user message and no system message, demonstrations, or instructions for intermediate reasoning.
We disable Qwen3 thinking and use greedy decoding in bf16, generating one answer per question.
All methods use the same prompts, generation limits, and scorers. The maximum input length is 32768 tokens. We generate at most 32 tokens for MMLU-Pro, 4096 for LiveCodeBench, and 2048 for MATH/GSM8K. The limits are 16 for PubMedQA, 256 for LegalBench, and 256 for BBH. Generation stops at EOS or the output length limit.
Unparseable answers are counted as incorrect without regeneration.
The prompts request an option letter for MMLU-Pro, Python 3 code for LiveCodeBench, a boxed final answer for MATH/GSM8K, and \texttt{yes}, \texttt{no}, or \texttt{maybe} for PubMedQA.
LegalBench and BBH retain their task instructions and answer formats.
Generated code is executed only for scoring. We use  \texttt{lcb\_runner} scorer. The limits are 6 seconds per test, 60 seconds per problem, and 4 GiB of memory.
MMLU-Pro uses all 12032 test questions. PubMedQA uses the 500 official test IDs from \texttt{pqa\_labeled}. Both use accuracy over all evaluated questions. LiveCodeBench uses all 880 problems in \texttt{code\_generation\_lite}, \texttt{release\_v5}, from May 2023 to January 2025. We report single-generation pass@1.
For Math, we use all 5000 MATH test questions and 1319 GSM8K test questions. We calculate accuracy for each benchmark and average the two scores equally.
LegalBench uses revision \texttt{6b4b377}. We exclude \texttt{rule\_qa} and use all 90844 test examples from the remaining 161 tasks. We average the official scores equally across these tasks.
BBH uses all 6511 examples from 27 task configurations. We calculate exact match for each configuration and average the scores equally. The object-count variants of logical deduction and shuffled-object tracking are scored separately.
Each domain score is averaged over four independently trained seeds, with sample standard deviation. For paired differences, we first subtract the baseline score from the CORTEX score for each seed. We then calculate the average and sample standard deviation.

\FloatBarrier

\subsection{Additional Results}
In this section, we present additional experimental results of our work.

\begin{table}[!t]
\scriptsize
\centering
\caption{Synthetic benchmark EM (\%). We show the mean $\pm$ standard deviation over 10 different seeds.}
\label{tab:synthetic_em}
\resizebox{\textwidth}{!}{
\begin{tabular}{lcccccc}
\toprule
\textbf{Baseline} & \textbf{COPY} & \textbf{REVERSE} & \textbf{ADD} & \textbf{LOOKUP} & \textbf{LOOKUP$\rightarrow$ADD} & \textbf{REVERSE$\rightarrow$LOOKUP} \\
\midrule
\rowcolor{grey!5}Dense
& $96.82{\pm}0.31$
& $91.36{\pm}0.44$
& $88.27{\pm}0.52$
& $92.74{\pm}0.39$
& $84.91{\pm}0.61$
& $86.38{\pm}0.57$\\

\rowcolor{grey!5}Random
& $88.54{\pm}0.66$
& $87.92{\pm}0.58$
& $80.76{\pm}0.73$
& $91.18{\pm}0.47$
& $73.64{\pm}0.85$
& $86.95{\pm}0.60$ \\

\rowcolor{grey!5}MoE
& $98.12{\pm}0.25$
& $95.06{\pm}0.34$
& $94.38{\pm}0.41$
& $95.71{\pm}0.33$
& $87.46{\pm}0.56$
& $91.64{\pm}0.46$ \\

\rowcolor{grey!5}MoM
& $98.67{\pm}0.21$
& $97.84{\pm}0.27$
& $96.21{\pm}0.33$
& $96.14{\pm}0.31$
& $90.82{\pm}0.49$
& $94.76{\pm}0.39$ \\

\rowcolor{grey!5}UpIT
& $96.21{\pm}0.36$
& $93.71{\pm}0.42$
& $92.04{\pm}0.48$
& $96.48{\pm}0.32$
& $86.39{\pm}0.59$
& $90.43{\pm}0.50$ \\

\rowcolor{grey!5}Self-MoE
& $97.84{\pm}0.28$
& $96.52{\pm}0.31$
& $93.91{\pm}0.43$
& $95.18{\pm}0.35$
& $89.34{\pm}0.52$
& $95.71{\pm}0.37$ \\

\rowcolor{green!12}
\textbf{CORTEX}
& $\mathbf{99.42{\pm}0.18}$
& $\mathbf{98.83{\pm}0.22}$
& $\mathbf{97.62{\pm}0.29}$
& $\mathbf{98.14{\pm}0.26}$
& $\mathbf{94.76{\pm}0.41}$
& $\mathbf{96.88{\pm}0.33}$ \\
\bottomrule
\end{tabular}
}
\end{table}

Table \ref{tab:synthetic_em} shows that CORTEX outperforms the baseline schemes on the synthetic benchmark using the 160M model.

\begin{table}[!t]
\centering
\caption{Real-domain benchmark NLL on Qwen3-8B. We show the mean $\pm$ standard deviation over 4 different seeds.}
\label{tab:real8b_nll}
\resizebox{\textwidth}{!}{
\begin{tabular}{lcccccc}
\toprule
\textbf{Baseline} & \textbf{General} & \textbf{Code} & \textbf{Math} & \textbf{Biomedical} & \textbf{Legal} & \textbf{Reasoning} \\
\midrule
\rowcolor{grey!5}Dense
& $1.6766{\pm}0.0124$
& $1.7949{\pm}0.0141$
& $2.0047{\pm}0.0168$
& $1.8220{\pm}0.0149$
& $1.9425{\pm}0.0161$
& $1.9683{\pm}0.0157$ \\

\rowcolor{grey!5}Random
& $1.7997{\pm}0.0175$
& $1.7924{\pm}0.0158$
& $2.0172{\pm}0.0194$
& $1.9534{\pm}0.0182$
& $2.0053{\pm}0.0197$
& $1.9392{\pm}0.0171$ \\

\rowcolor{grey!5}MoE
& $1.6642{\pm}0.0119$
& $1.7878{\pm}0.0136$
& $1.9262{\pm}0.0155$
& $1.7947{\pm}0.0142$
& $1.9054{\pm}0.0153$
& $1.9133{\pm}0.0149$ \\

\rowcolor{grey!5}MoM
& $1.6441{\pm}0.0112$
& $1.6578{\pm}0.0124$
& $1.9290{\pm}0.0151$
& $1.7995{\pm}0.0138$
& $1.8714{\pm}0.0147$
& $1.8385{\pm}0.0136$ \\

\rowcolor{grey!5}UpIT
& $1.6257{\pm}0.0108$
& $1.7658{\pm}0.0131$
& $2.0039{\pm}0.0162$
& $1.7709{\pm}0.0134$
& $1.8171{\pm}0.0140$
& $1.8288{\pm}0.0132$ \\

\rowcolor{grey!5}Self-MoE
& $1.6684{\pm}0.0121$
& $1.7373{\pm}0.0129$
& $1.9058{\pm}0.0148$
& $1.8208{\pm}0.0145$
& $1.8583{\pm}0.0146$
& $\mathbf{1.7822{\pm}0.0128}$ \\

\rowcolor{green!12}
\textbf{CORTEX}
& $\mathbf{1.6199{\pm}0.0097}$
& $\mathbf{1.6096{\pm}0.0115}$
& $\mathbf{1.8694{\pm}0.0137}$
& $\mathbf{1.6778{\pm}0.0126}$
& $\mathbf{1.7791{\pm}0.0131}$
& $1.8258{\pm}0.0124$ \\
\bottomrule
\end{tabular}
}
\end{table}

\begin{table}[!t]
\centering
\caption{Real-domain benchmark NLL on Qwen3-32B. We show the mean $\pm$ standard deviation over 4 different seeds.}
\label{tab:real32b_nll}
\resizebox{\textwidth}{!}{
\begin{tabular}{lcccccc}
\toprule
\textbf{Baseline} & \textbf{General} & \textbf{Code} & \textbf{Math} & \textbf{Biomedical} & \textbf{Legal} & \textbf{Reasoning} \\
\midrule
\rowcolor{grey!5}Dense
& $1.4202{\pm}0.0108$
& $1.5311{\pm}0.0122$
& $1.7400{\pm}0.0145$
& $1.5648{\pm}0.0129$
& $1.6712{\pm}0.0136$
& $1.7221{\pm}0.0140$ \\

\rowcolor{grey!5}Random
& $1.5298{\pm}0.0149$
& $1.5193{\pm}0.0135$
& $1.7889{\pm}0.0161$
& $1.6876{\pm}0.0152$
& $1.7272{\pm}0.0157$
& $1.6897{\pm}0.0148$ \\

\rowcolor{grey!5}MoE
& $1.3884{\pm}0.0099$
& $1.4383{\pm}0.0107$
& $1.6240{\pm}0.0128$
& $1.4878{\pm}0.0114$
& $1.6364{\pm}0.0130$
& $1.6111{\pm}0.0125$ \\

\rowcolor{grey!5}MoM
& $1.3918{\pm}0.0101$
& $1.4967{\pm}0.0119$
& $1.5815{\pm}0.0122$
& $1.5384{\pm}0.0120$
& $1.5942{\pm}0.0127$
& $1.6129{\pm}0.0124$ \\

\rowcolor{grey!5}UpIT
& $\mathbf{1.3578{\pm}0.0093}$
& $1.4733{\pm}0.0112$
& $1.6766{\pm}0.0134$
& $1.5100{\pm}0.0117$
& $1.5514{\pm}0.0119$
& $1.5945{\pm}0.0121$ \\

\rowcolor{grey!5}Self-MoE
& $1.3871{\pm}0.0098$
& $1.4395{\pm}0.0108$
& $1.5845{\pm}0.0120$
& $1.5180{\pm}0.0118$
& $1.6011{\pm}0.0126$
& $1.5597{\pm}0.0116$ \\

\rowcolor{green!12}
\textbf{CORTEX}
& $1.3820{\pm}0.0091$
& $\mathbf{1.4051{\pm}0.0102}$
& $\mathbf{1.5798{\pm}0.0117}$
& $\mathbf{1.4344{\pm}0.0109}$
& $\mathbf{1.5426{\pm}0.0110}$
& $\mathbf{1.5279{\pm}0.0111}$ \\
\bottomrule
\end{tabular}
}
\end{table}

In addition, Tables \ref{tab:real8b_nll} and \ref{tab:real32b_nll} compare the NLL on the real-domain benchmark using Qwen3-8B and Qwen3-32B, respectively. 
CORTEX achieves the best or highly competitive NLL across most knowledge domains under both backbone models. 
Compared with the \textbf{Dense} baseline, CORTEX consistently reduces NLL on general text, code, math, biomedical text, legal text, and reasoning data, showing that the learned modularization improves the learning performance under heterogeneous real-domain mixtures. 
Compared with MoE-based and other modular baselines, CORTEX provides a more balanced performance across domains, which indicates that the shared and specialized modules within the dense model can better preserve common knowledge while capturing domain-specific knowledge.

\begin{table}[!t]
\centering
\scriptsize
\caption{Paired differences in downstream scores (CORTEX minus the best baseline in Table~\ref{tab:downstream_best}). We show the average $\pm$ standard deviation of the differences over four paired training seeds.}
\label{tab:downstream_paired}
\setlength{\tabcolsep}{4pt}
\resizebox{\textwidth}{!}{
\begin{tabular}{lcccccc}
\toprule
\textbf{Backbone} & \textbf{General} & \textbf{Code} & \textbf{Math} & \textbf{Biomedical} & \textbf{Legal} & \textbf{Reasoning} \\
\midrule
Qwen3-8B & $+0.0115{\pm}0.0046$ & $-0.0023{\pm}0.0065$ & $+0.0340{\pm}0.0075$ & $+0.0075{\pm}0.0044$ & $+0.0253{\pm}0.0051$ & $-0.0063{\pm}0.0069$ \\
Qwen3-32B & $-0.0091{\pm}0.0038$ & $+0.0323{\pm}0.0053$ & $+0.0149{\pm}0.0061$ & $+0.0330{\pm}0.0042$ & $+0.0035{\pm}0.0046$ & $+0.0283{\pm}0.0058$ \\
\bottomrule
\end{tabular}
}
\end{table}

Table~\ref{tab:downstream_paired} reports the paired differences in downstream scores between CORTEX and the best baseline in each domain over four training seeds.

\begin{figure}[!t]
    \centering
    \includegraphics[width=\linewidth]{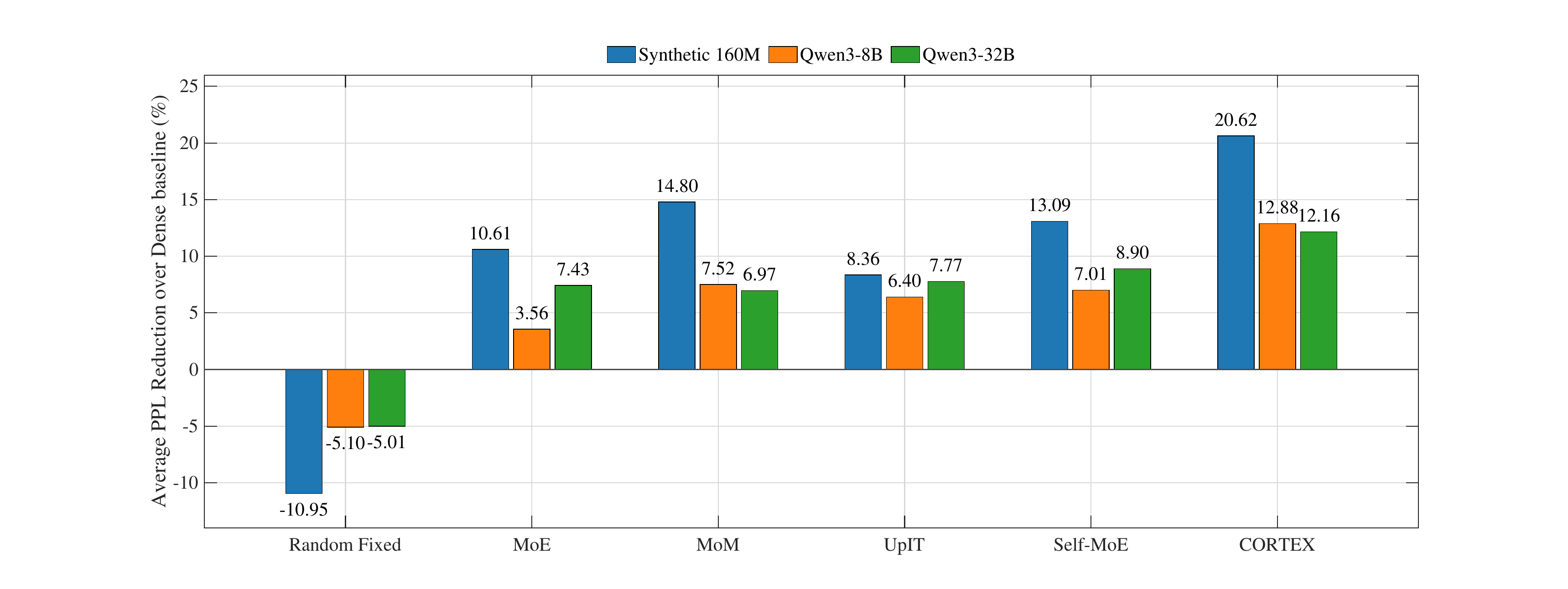}
    \caption{PPL reduction over the Dense baseline, calculated from the average NLL across knowledge domains.}
    \label{fig:average_core}
\end{figure}

Fig.~\ref{fig:average_core} further shows the average PPL reduction over the Dense baseline across the synthetic benchmark, Qwen3-8B, and Qwen3-32B, respectively. 
CORTEX obtains the largest average PPL reduction in all three settings, while random assignment consistently performs worse than Dense. 
Under the same domain-conditioned gradient collection and backbone update rule, CORTEX consistently outperforms fixed random module assignments. 
The consistent improvement from the 160M model to Qwen3-8B and Qwen3-32B further indicates that CORTEX can scale to larger dense backbones while maintaining its advantage over existing baseline schemes.

\begin{table}[!t]
\centering
\scriptsize
\caption{Average paired-domain SLS for original and matched-shuffle assignments. Difference denotes original minus matched-shuffle SLS. Values are averages $\pm$ standard deviations across training seeds.}
\label{tab:matched_shuffle}
\setlength{\tabcolsep}{5pt}
\begin{tabular}{lccc}
\toprule
\textbf{Backbone} & \textbf{Original} & \textbf{Matched shuffle} & \textbf{Difference} \\
\midrule
160M & $0.1837{\pm}0.0120$ & $0.0548{\pm}0.0100$ & $+0.1289{\pm}0.0115$ \\
Qwen3-8B & $0.1452{\pm}0.0150$ & $0.0427{\pm}0.0120$ & $+0.1026{\pm}0.0140$ \\
\bottomrule
\end{tabular}
\end{table}

We further compare the learned assignments with randomly shuffled assignments at the same trained checkpoints. We shuffle complete assignment vectors among same-shaped parameter groups within each matrix and keep the module--domain pairing and evaluation set fixed. For each matrix and module, we rescale the weights removed by the shuffled assignment to match the Frobenius norm of those removed by the original assignment. This matched-shuffle control accounts for differences in removal magnitude when comparing SLS. We use 10 training seeds with 100 shuffles per seed for 160M and 4 training seeds with 20 shuffles per seed for Qwen3-8B. We average SLS across shuffles within each seed and calculate paired differences before reporting the average $\pm$ standard deviation across training seeds. Table~\ref{tab:matched_shuffle} shows higher average SLS for the original assignments on both backbones, supporting domain-selective localization under the matched-shuffle control.

\FloatBarrier
\begingroup
\setlength{\floatsep}{6pt}
\setlength{\textfloatsep}{8pt}

\begin{figure}[!t]
    \captionsetup{skip=3pt}
    \centering
    \begin{minipage}[t]{0.32\textwidth}
        \centering
        \includegraphics[width=\linewidth]{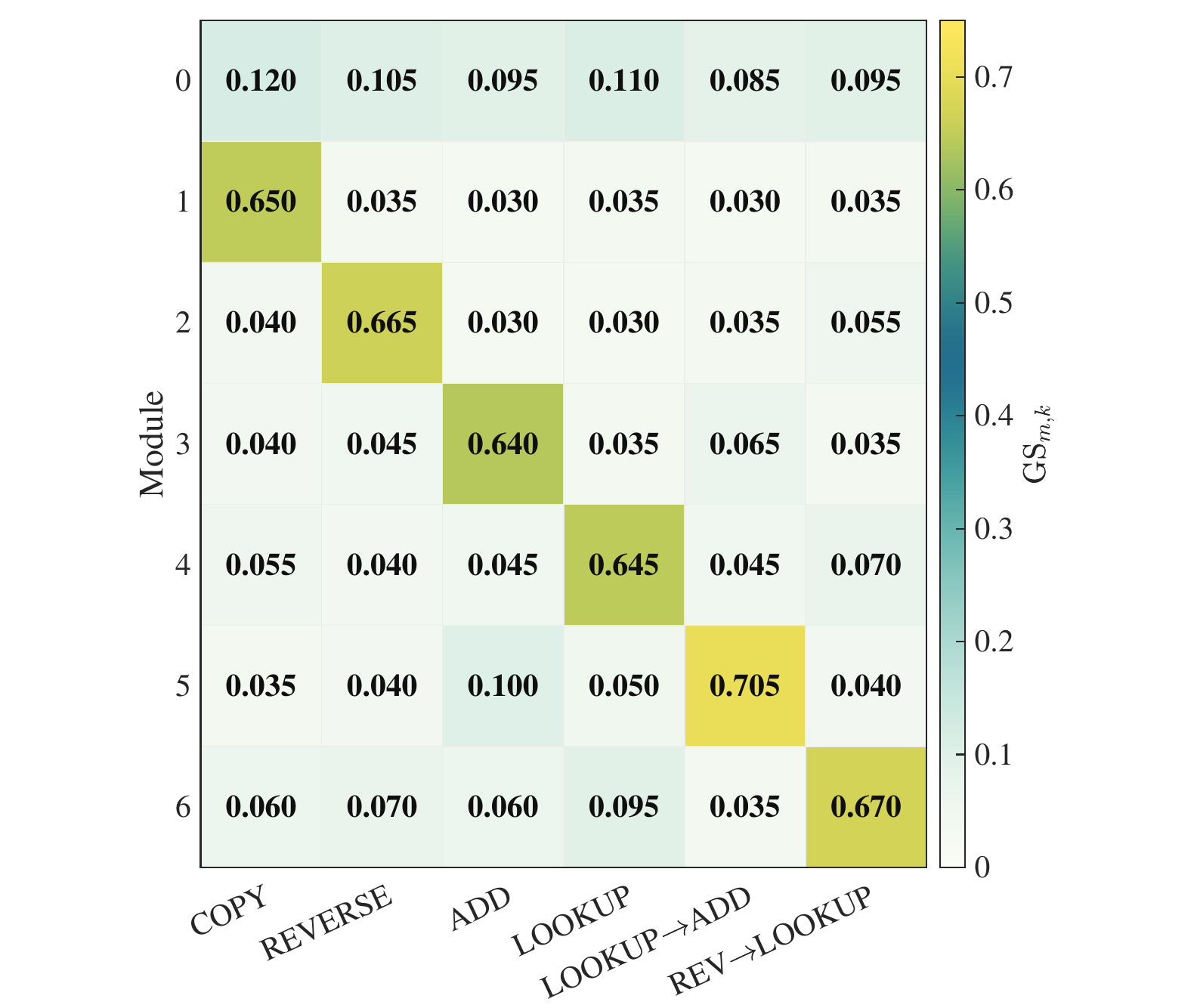}
        {\tiny (a) }
    \end{minipage}
    \begin{minipage}[t]{0.32\textwidth}
        \centering
        \includegraphics[width=\linewidth]{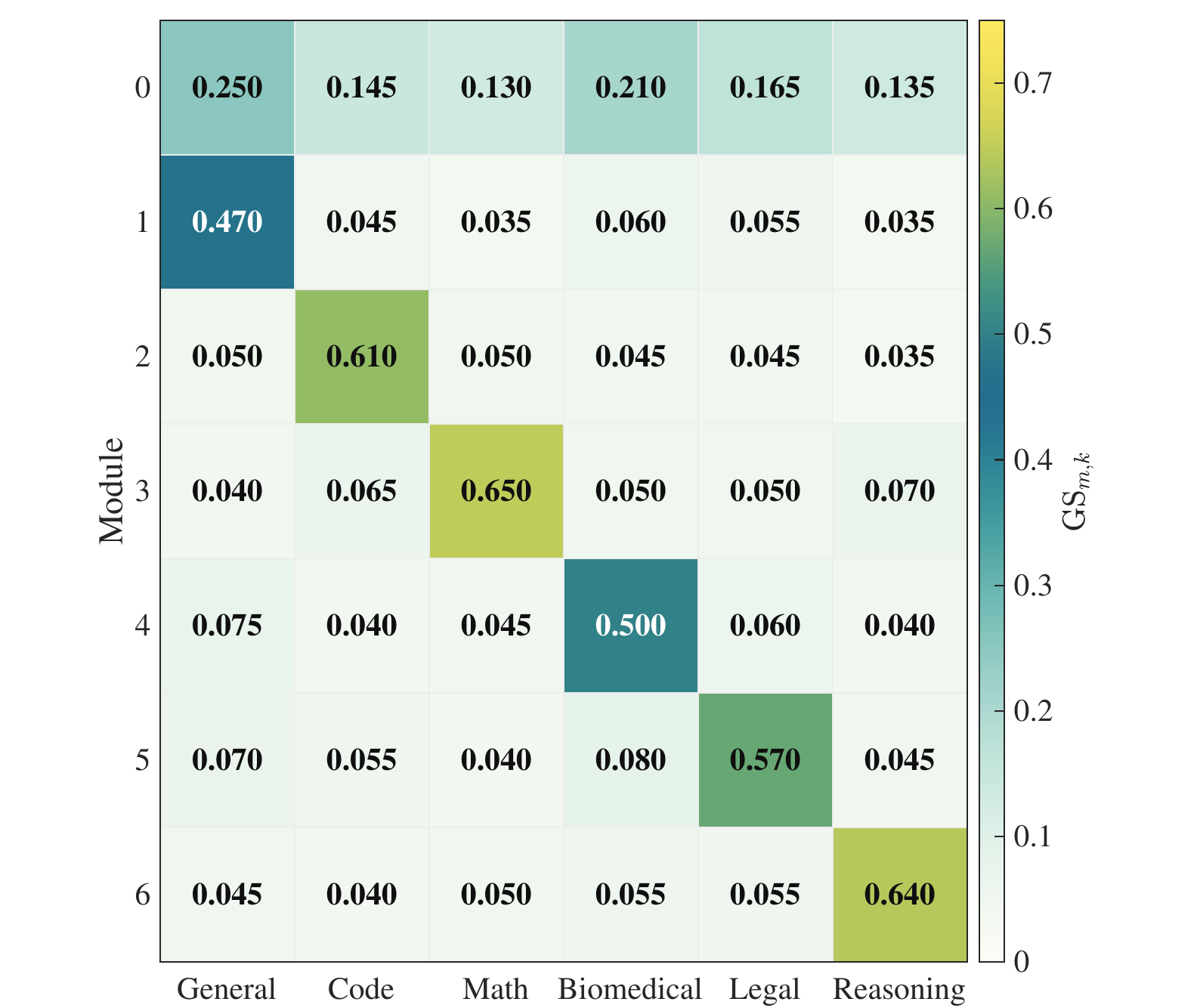}
        {\tiny (b) }
    \end{minipage}
    \begin{minipage}[t]{0.32\textwidth}
        \centering
        \includegraphics[width=\linewidth]{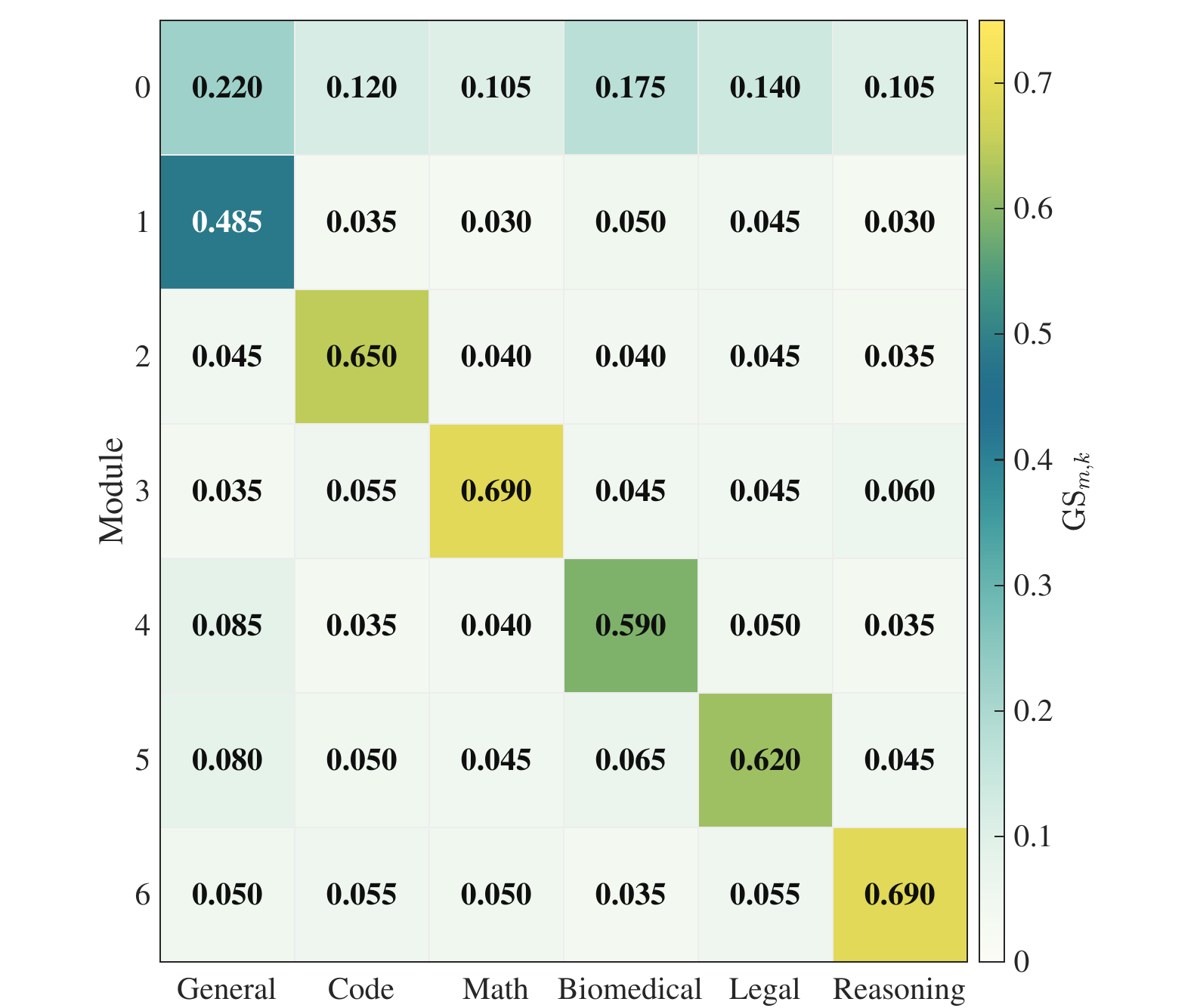}
        {\tiny (c) }
    \end{minipage}

    \caption{
    Domain-normalized gradient share ($\mathrm{GS}_{m,k}/\mathrm{GS}_{k}^{\mathrm{domain}}$) on (a) 160M model, (b) Qwen3-8B, and (c) Qwen3-32B. 
    The columns denote knowledge domains and the rows denote modules. Each column sums to one.}
    \label{fig:mechanistic_GS}
\end{figure}

For Fig.~\ref{fig:mechanistic_GS}, we average the joint gradient shares across seeds and then normalize each column to sum to one. For each knowledge domain, the corresponding specialized module receives the largest gradient share, while the shared module receives gradients from all domains. We calculate $I^{\mathrm{mod}}$ from the joint gradient shares in Definition~4 for each seed and report the average in Table~\ref{tab:nll_mi}.

\begin{figure}[!t]
    \captionsetup{skip=3pt}
    \centering
    \begin{minipage}[t]{0.24\textwidth}
        \centering
        \includegraphics[width=\linewidth]{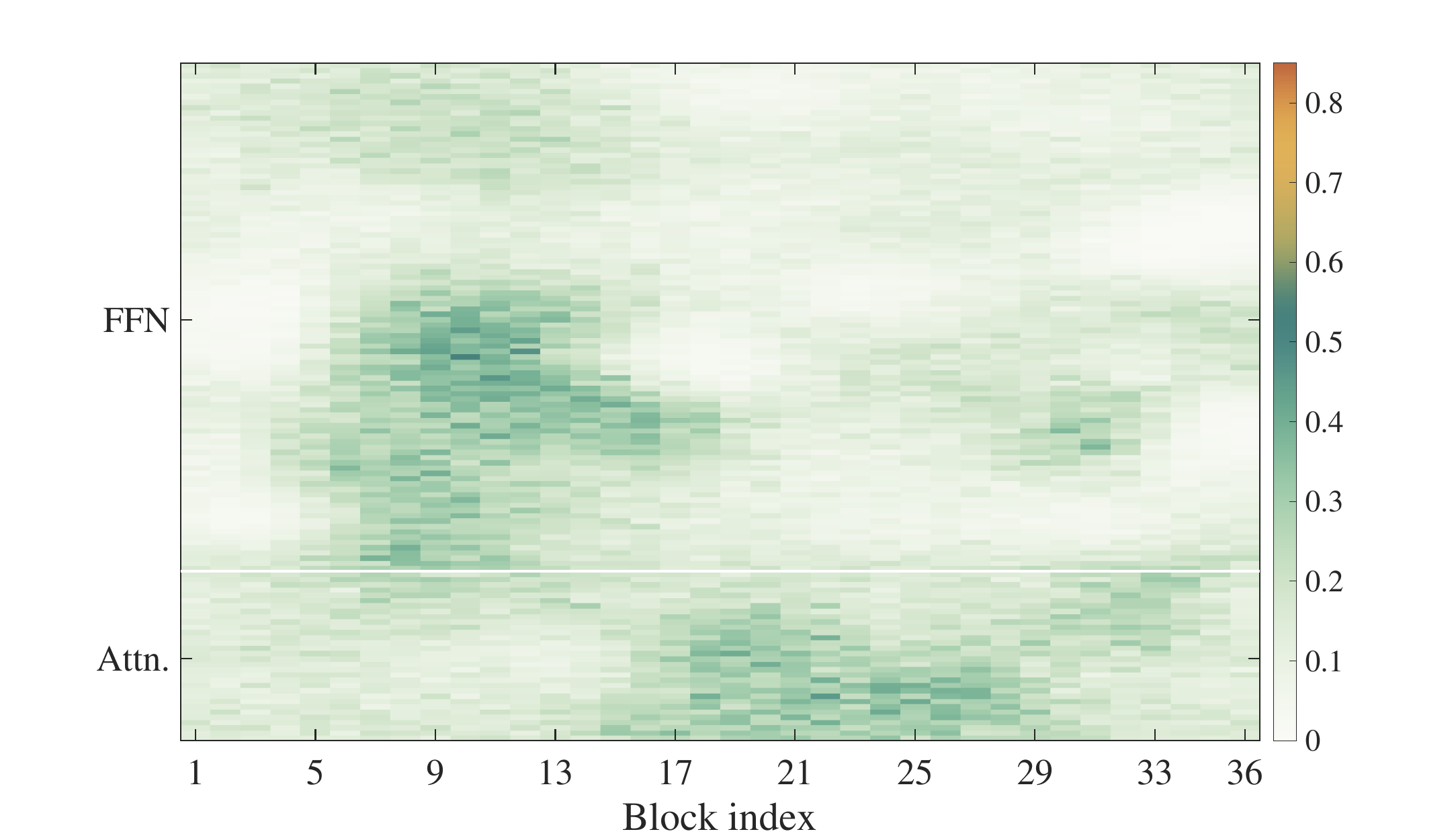}
        {\tiny (a) }
    \end{minipage}
    \begin{minipage}[t]{0.24\textwidth}
        \centering
        \includegraphics[width=\linewidth]{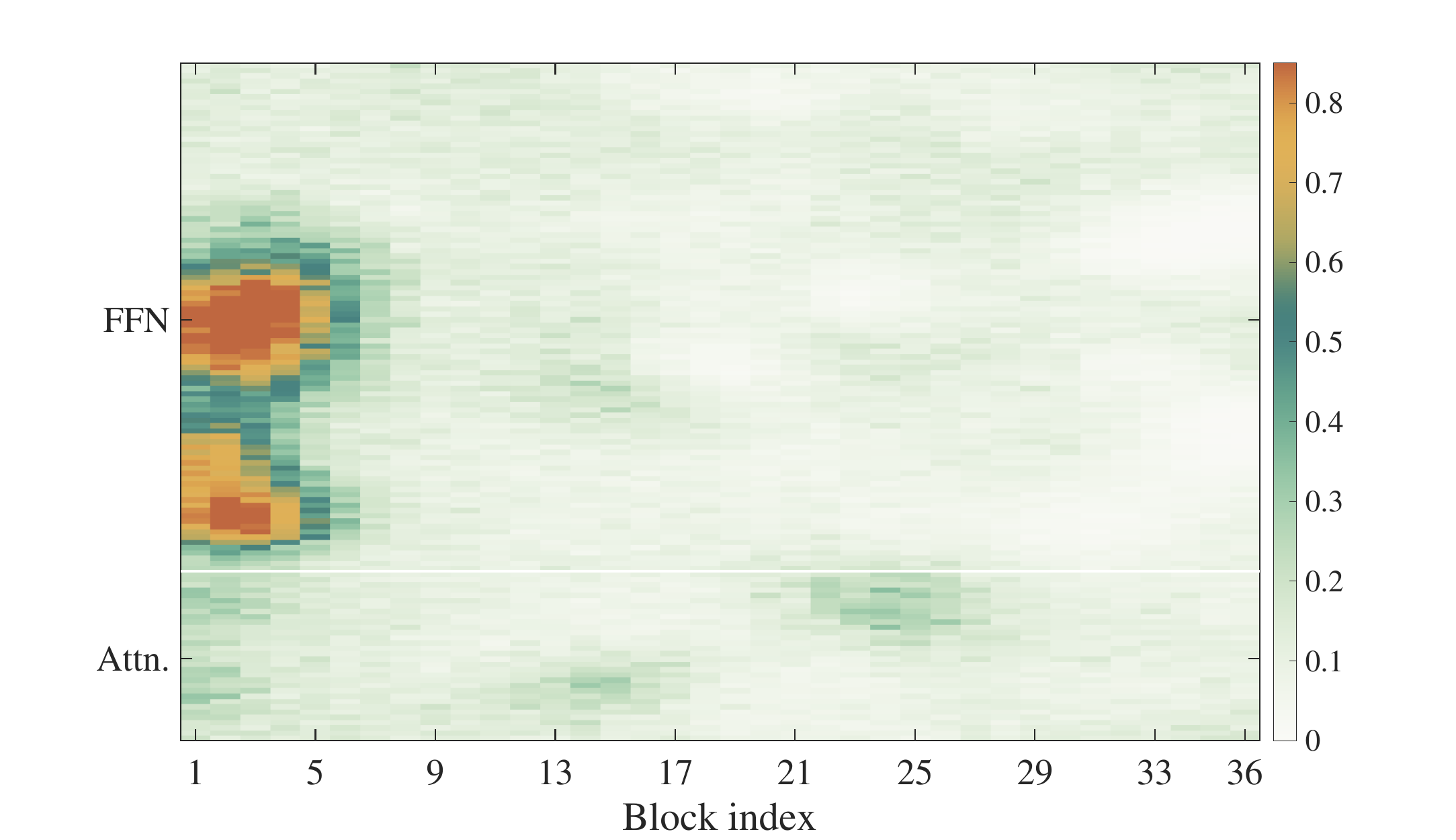}
        {\tiny (b) }
    \end{minipage}
    \begin{minipage}[t]{0.24\textwidth}
        \centering
        \includegraphics[width=\linewidth]{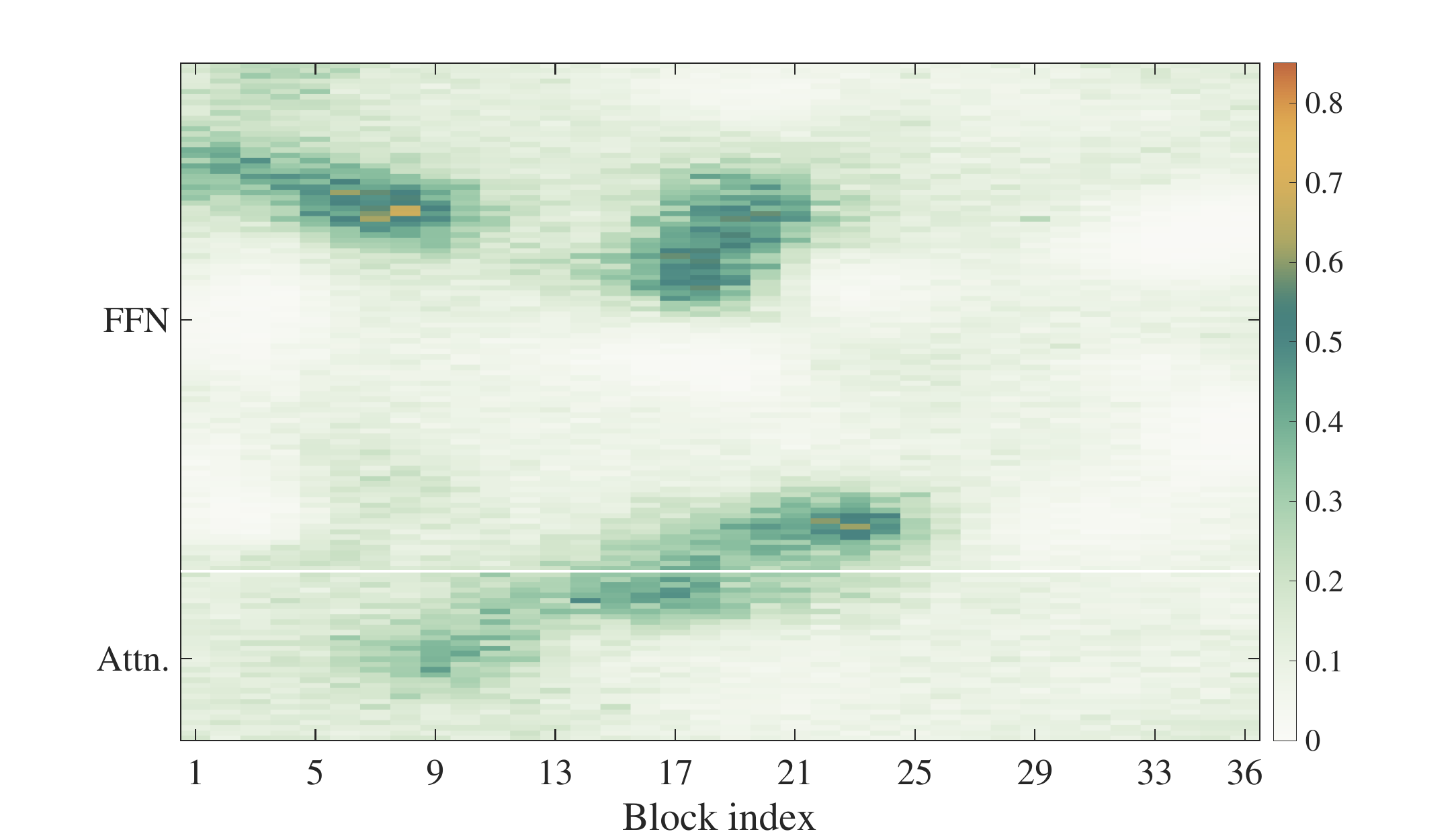}
        {\tiny (c) }
    \end{minipage}
    \begin{minipage}[t]{0.24\textwidth}
        \centering
        \includegraphics[width=\linewidth]{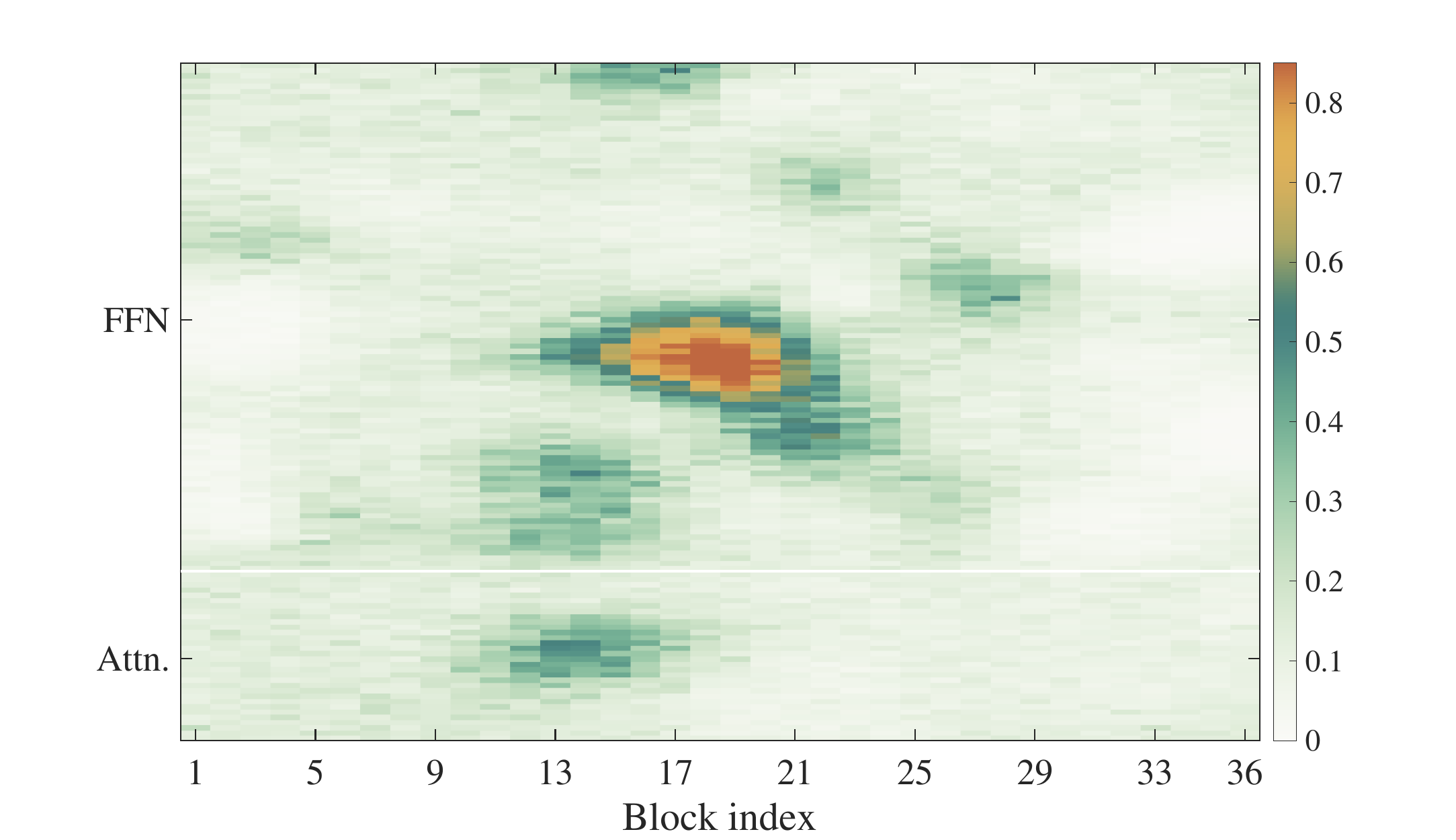}
        {\tiny (d) }
    \end{minipage}


    \begin{minipage}[t]{0.24\textwidth}
        \centering
        \includegraphics[width=\linewidth]{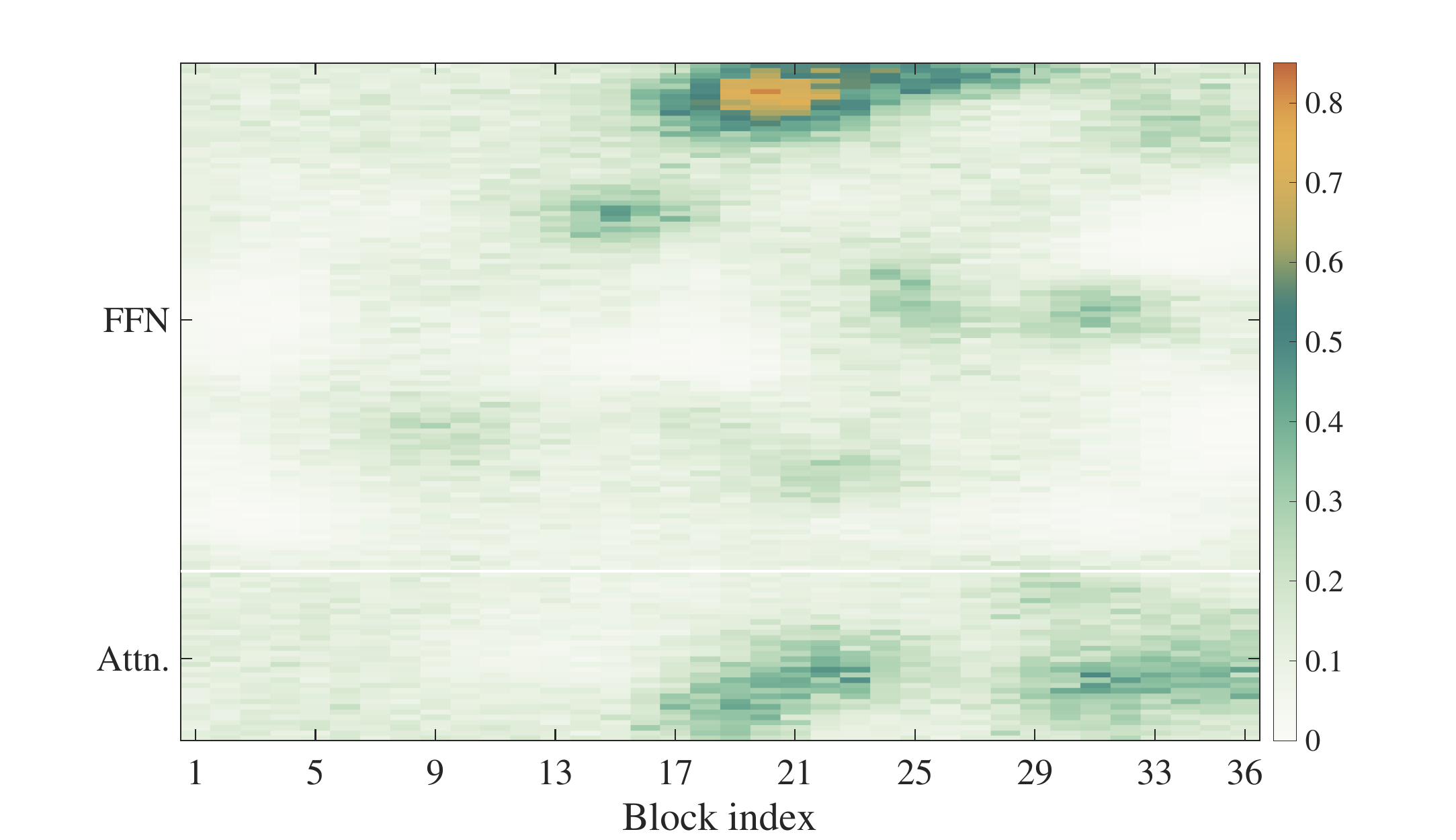}
        {\tiny (e) }
    \end{minipage}
    \begin{minipage}[t]{0.24\textwidth}
        \centering
        \includegraphics[width=\linewidth]{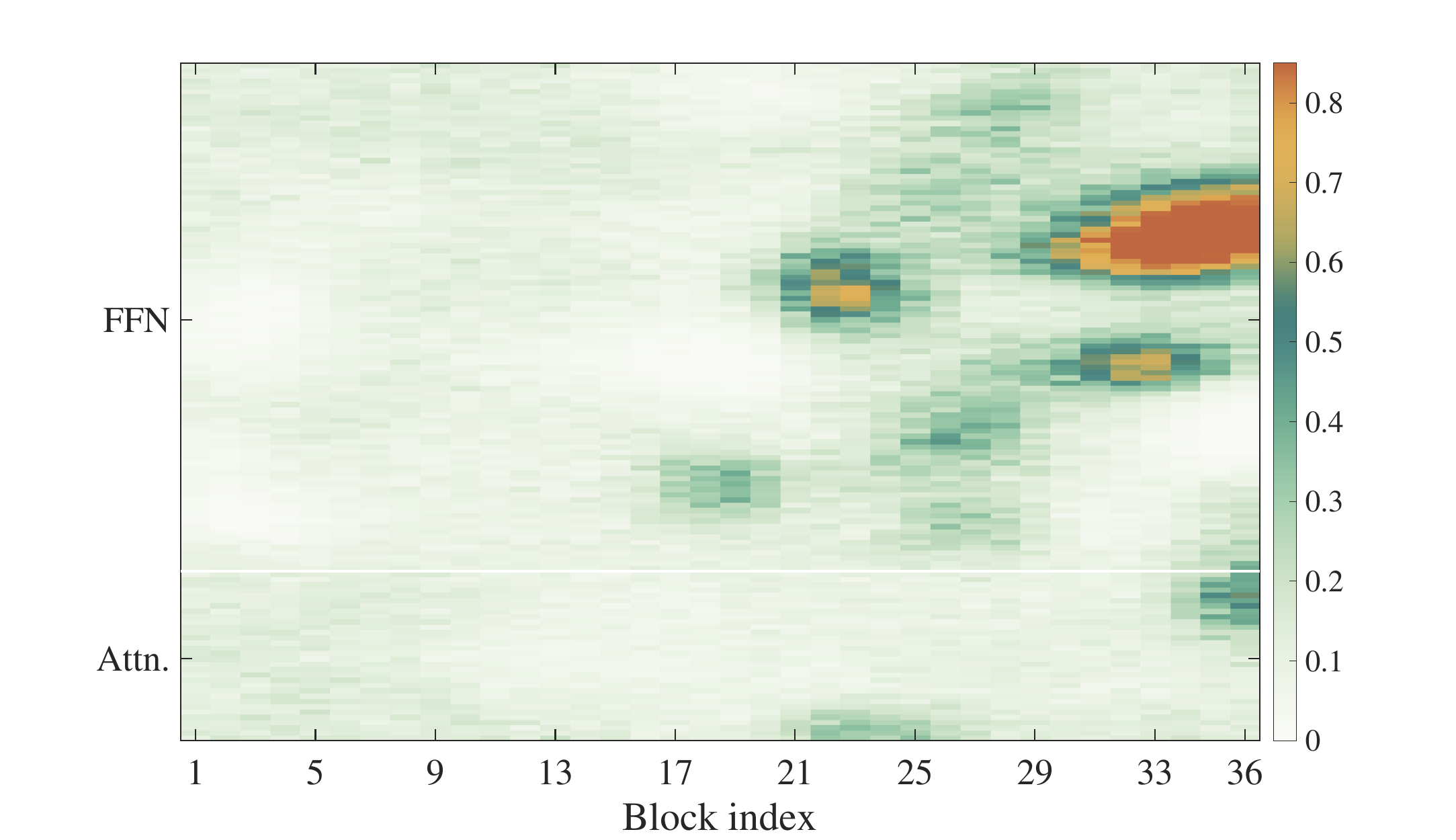}
        {\tiny (f) }
    \end{minipage}
    \begin{minipage}[t]{0.24\textwidth}
        \centering
        \includegraphics[width=\linewidth]{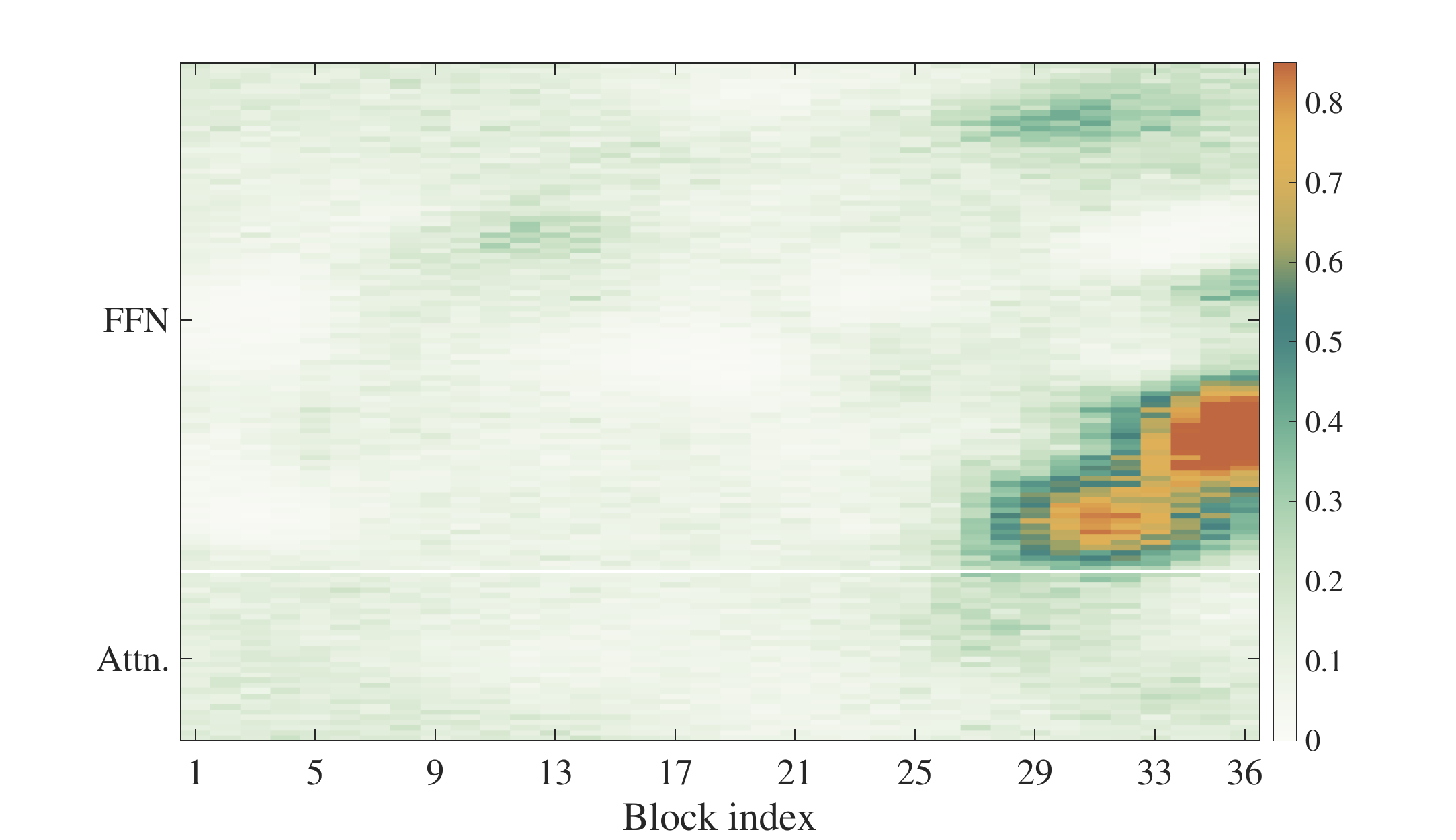}
        {\tiny (g) }
    \end{minipage}

    \caption{
    Visualization of modularization of CORTEX on the Qwen3-8B backbone model trained on the real-domain data mixture. 
    The seven heatmaps correspond to the shared knowledge and six knowledge domains, including (a) Shared, (b) General, (c) Code, (d) Math, (e) Biomedical, (f) Legal, and (g) Reasoning, respectively.
    The columns denote Transformer blocks and the rows denote parameter groups within attention heads (Attn.) and the feed-forward network (FFN).}
    \label{fig:demo_2}
\end{figure}

\begin{figure}[!t]
    \captionsetup{skip=3pt}
    \centering
    \begin{minipage}[t]{0.24\textwidth}
        \centering
        \includegraphics[width=\linewidth]{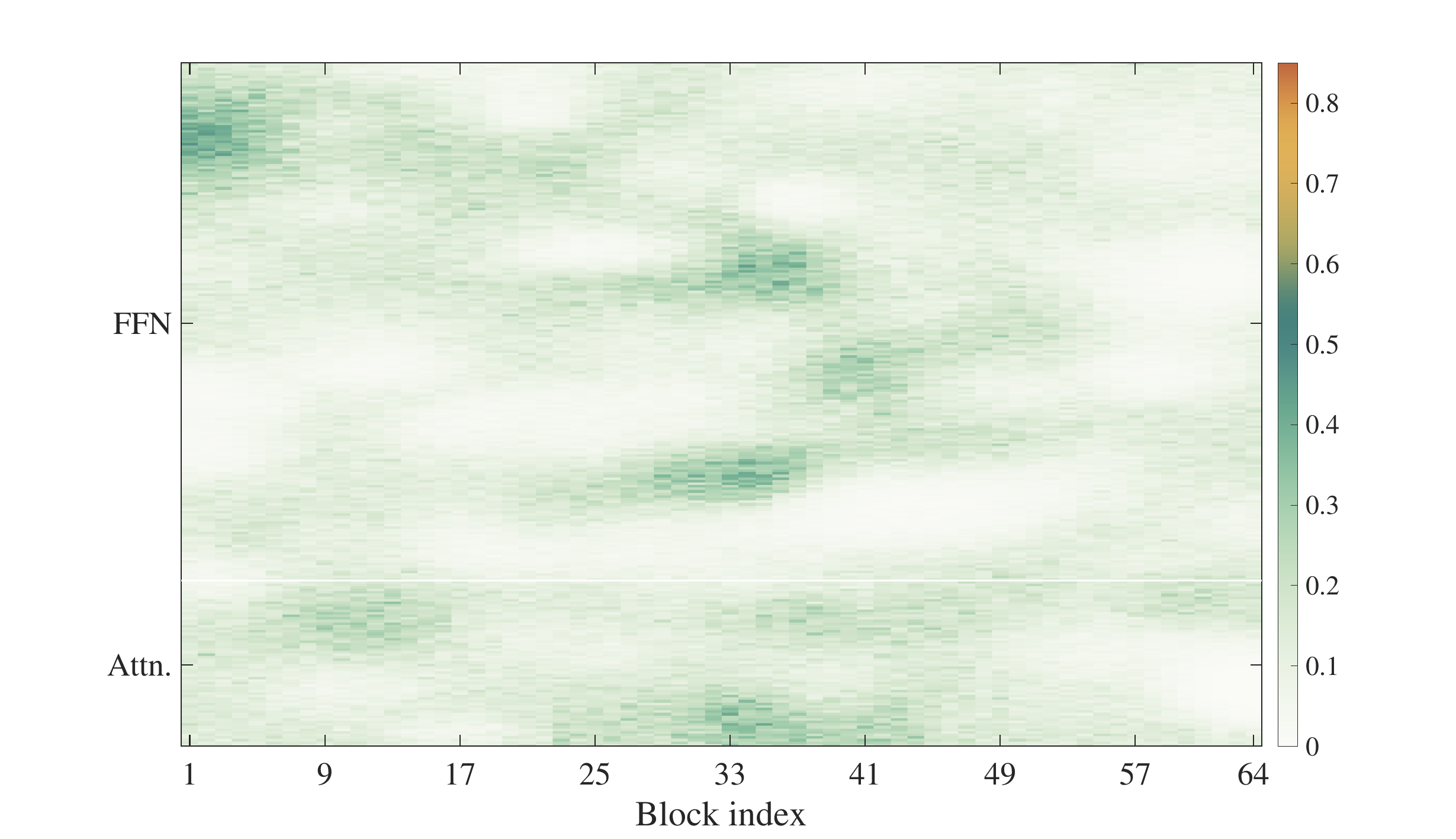}
        {\tiny (a) }
    \end{minipage}
    \begin{minipage}[t]{0.24\textwidth}
        \centering
        \includegraphics[width=\linewidth]{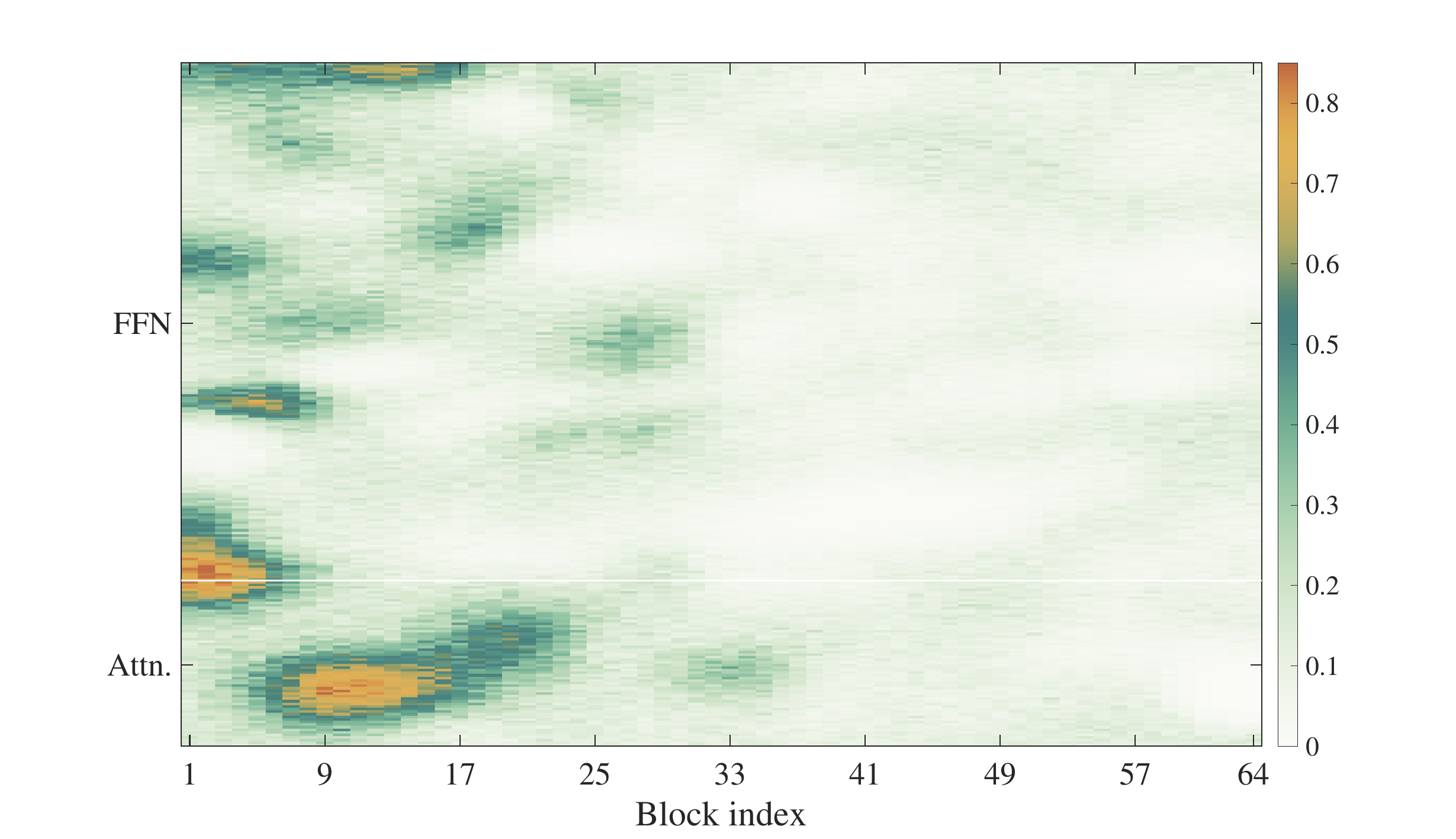}
        {\tiny (b) }
    \end{minipage}
    \begin{minipage}[t]{0.24\textwidth}
        \centering
        \includegraphics[width=\linewidth]{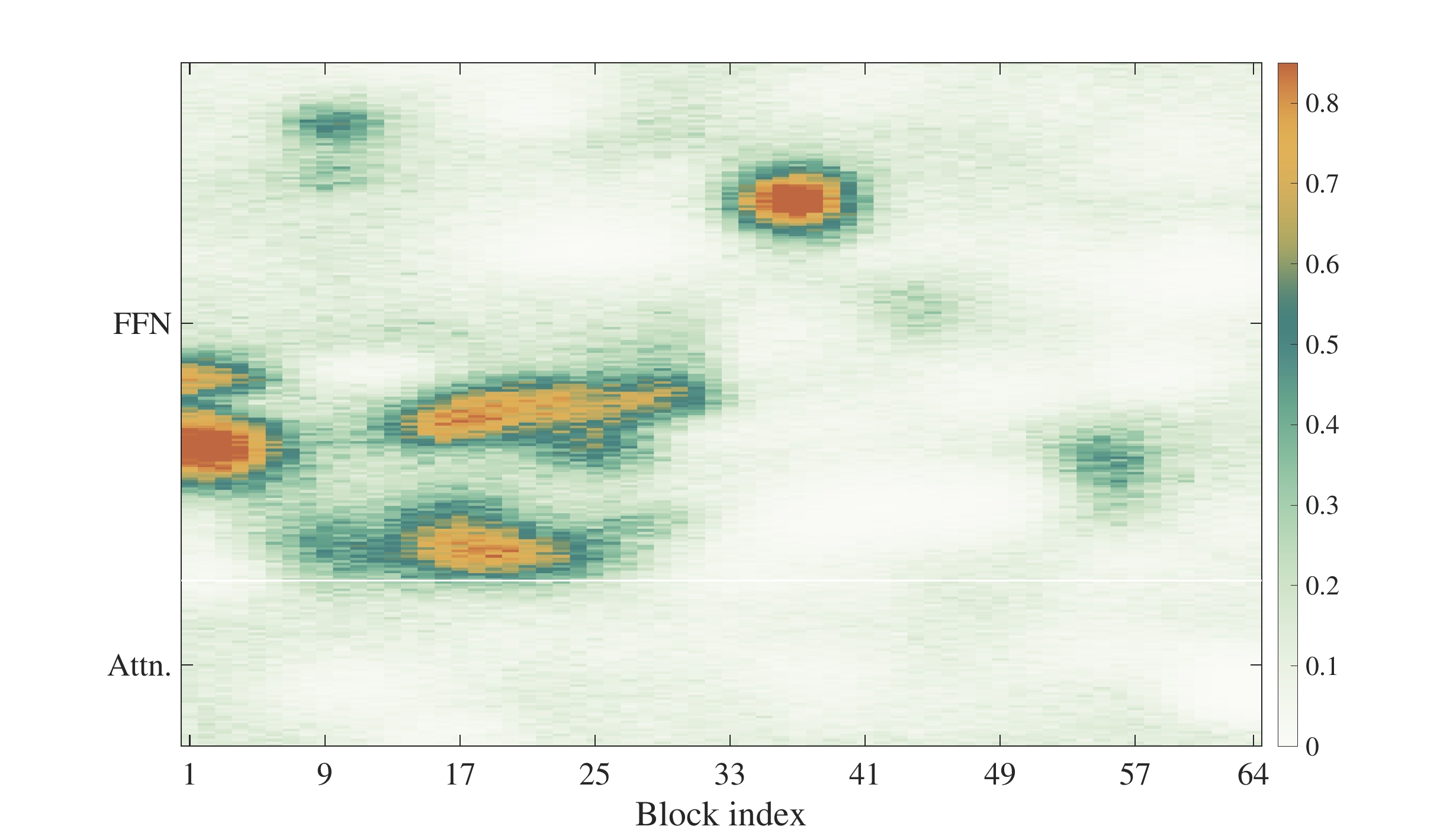}
        {\tiny (c) }
    \end{minipage}
    \begin{minipage}[t]{0.24\textwidth}
        \centering
        \includegraphics[width=\linewidth]{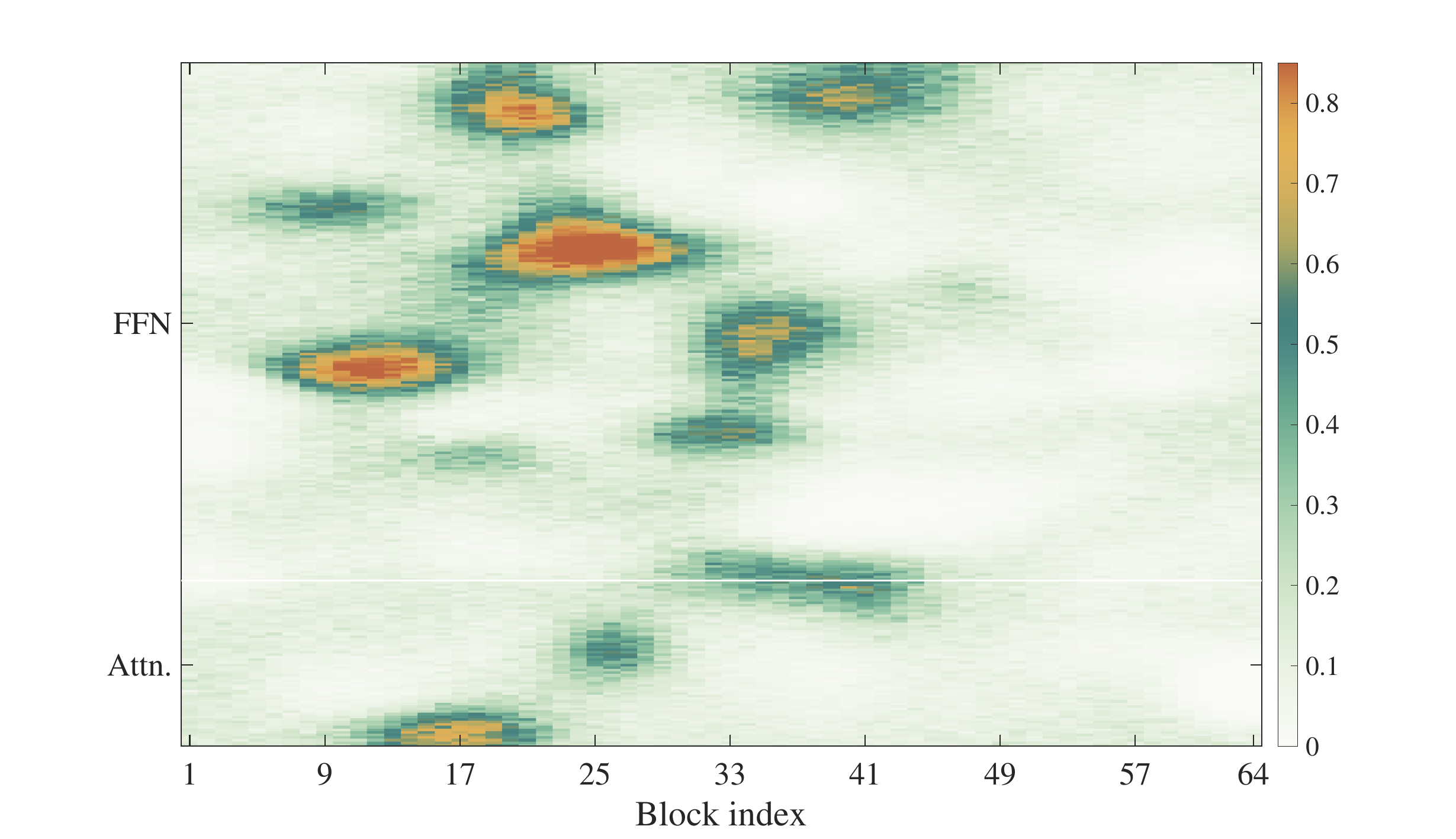}
        {\tiny (d) }
    \end{minipage}


    \begin{minipage}[t]{0.24\textwidth}
        \centering
        \includegraphics[width=\linewidth]{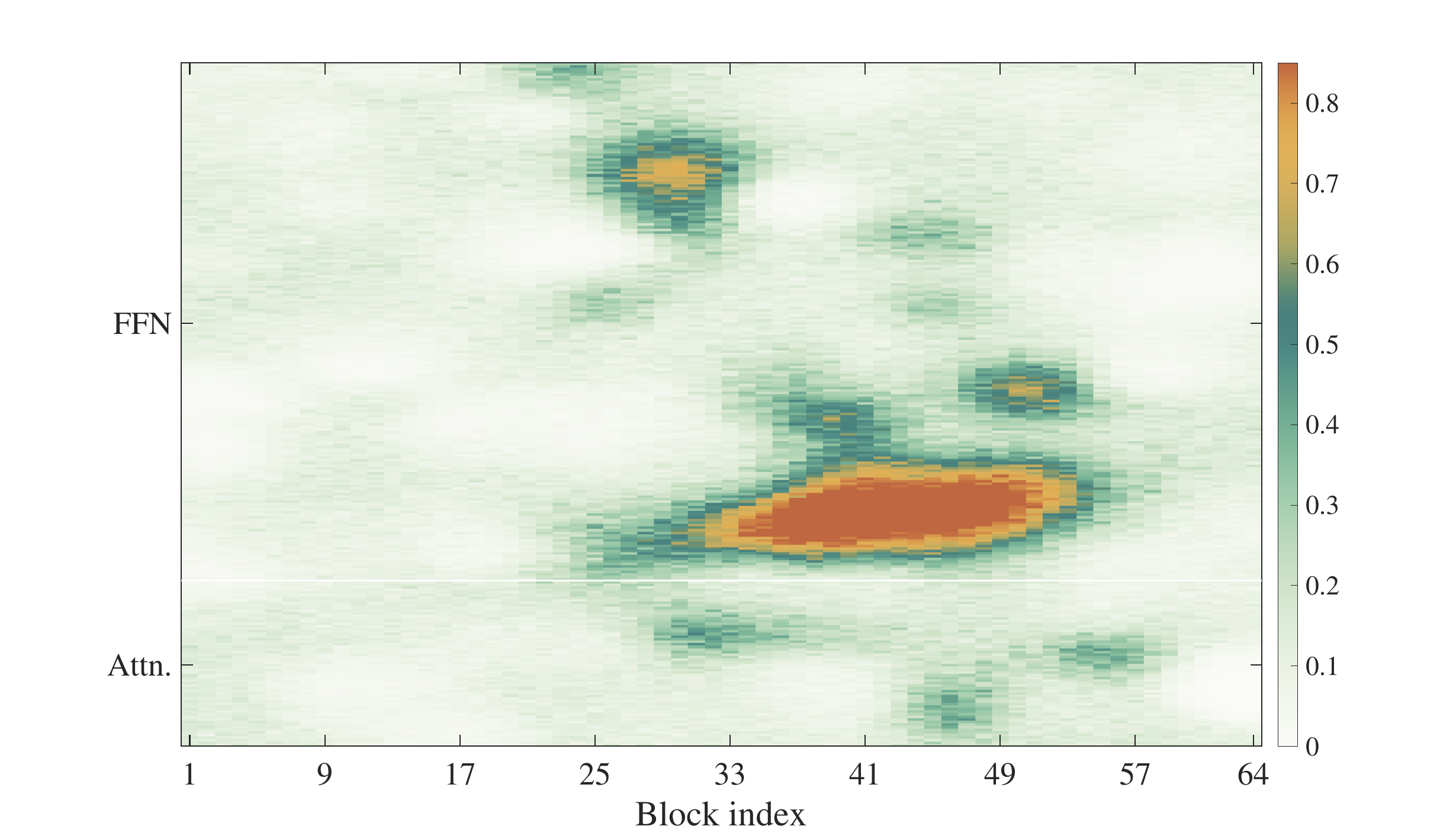}
        {\tiny (e) }
    \end{minipage}
    \begin{minipage}[t]{0.24\textwidth}
        \centering
        \includegraphics[width=\linewidth]{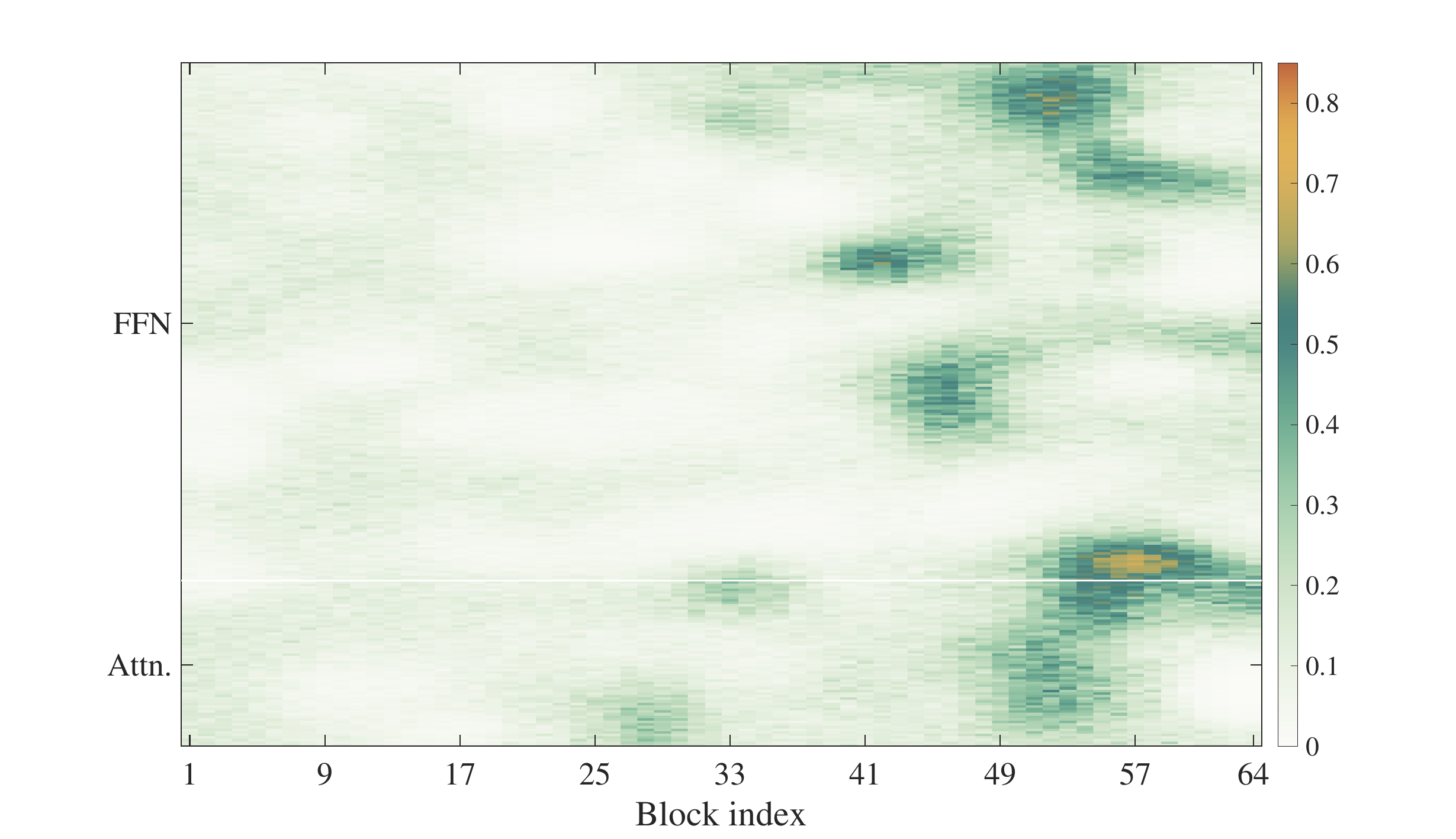}
        {\tiny (f) }
    \end{minipage}
    \begin{minipage}[t]{0.24\textwidth}
        \centering
        \includegraphics[width=\linewidth]{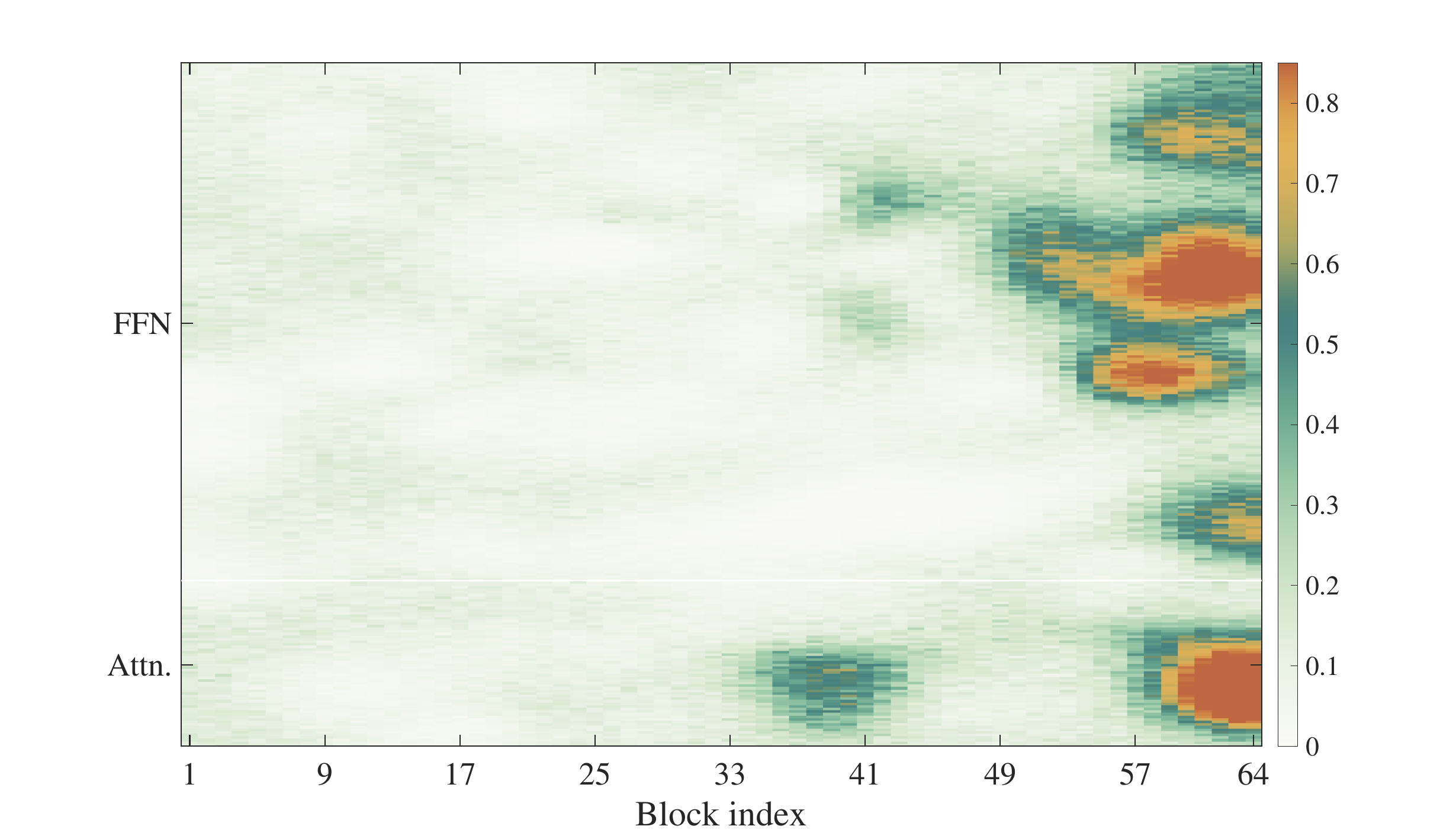}
        {\tiny (g) }
    \end{minipage}

    \caption{
    Visualization of modularization of CORTEX on the Qwen3-32B backbone model trained on the real-domain data mixture. 
    The seven heatmaps correspond to the shared knowledge and six knowledge domains, including (a) Shared, (b) General, (c) Code, (d) Math, (e) Biomedical, (f) Legal, and (g) Reasoning, respectively.
    The columns denote Transformer blocks and the rows denote parameter groups within attention heads (Attn.) and the feed-forward network (FFN).}
    \label{fig:demo_3}
\end{figure}

Figs. \ref{fig:demo_2} and \ref{fig:demo_3} show the modularization of CORTEX on Qwen3-8B and Qwen3-32B.
The module assignment weight patterns show that CORTEX forms both shared and domain-specialized modules across Transformer blocks, indicating that the modular structure persists when the backbone model scales from the 160M model to larger dense LLMs.

\FloatBarrier\captionsetup{skip=3pt}

\begin{figure}[!t]
    \centering
    \includegraphics[width=0.32\linewidth]{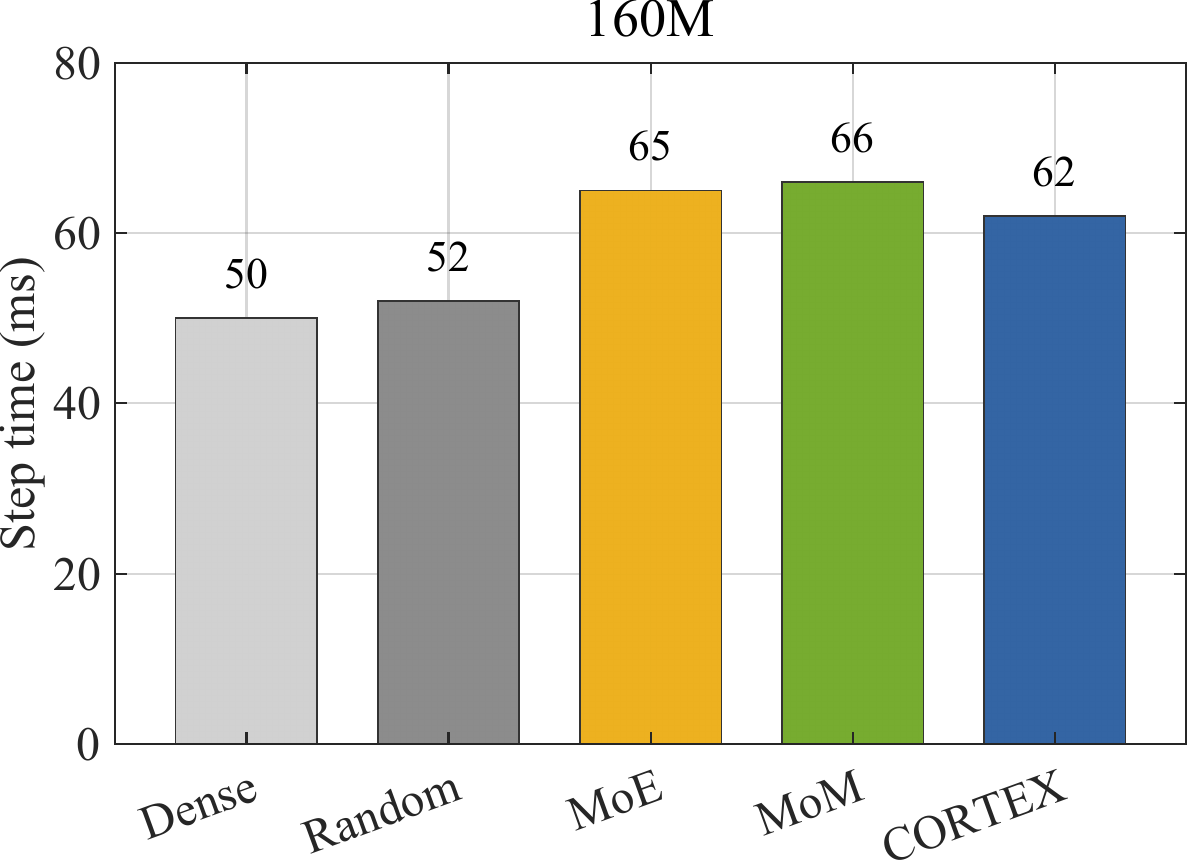}\hfill
    \includegraphics[width=0.32\linewidth]{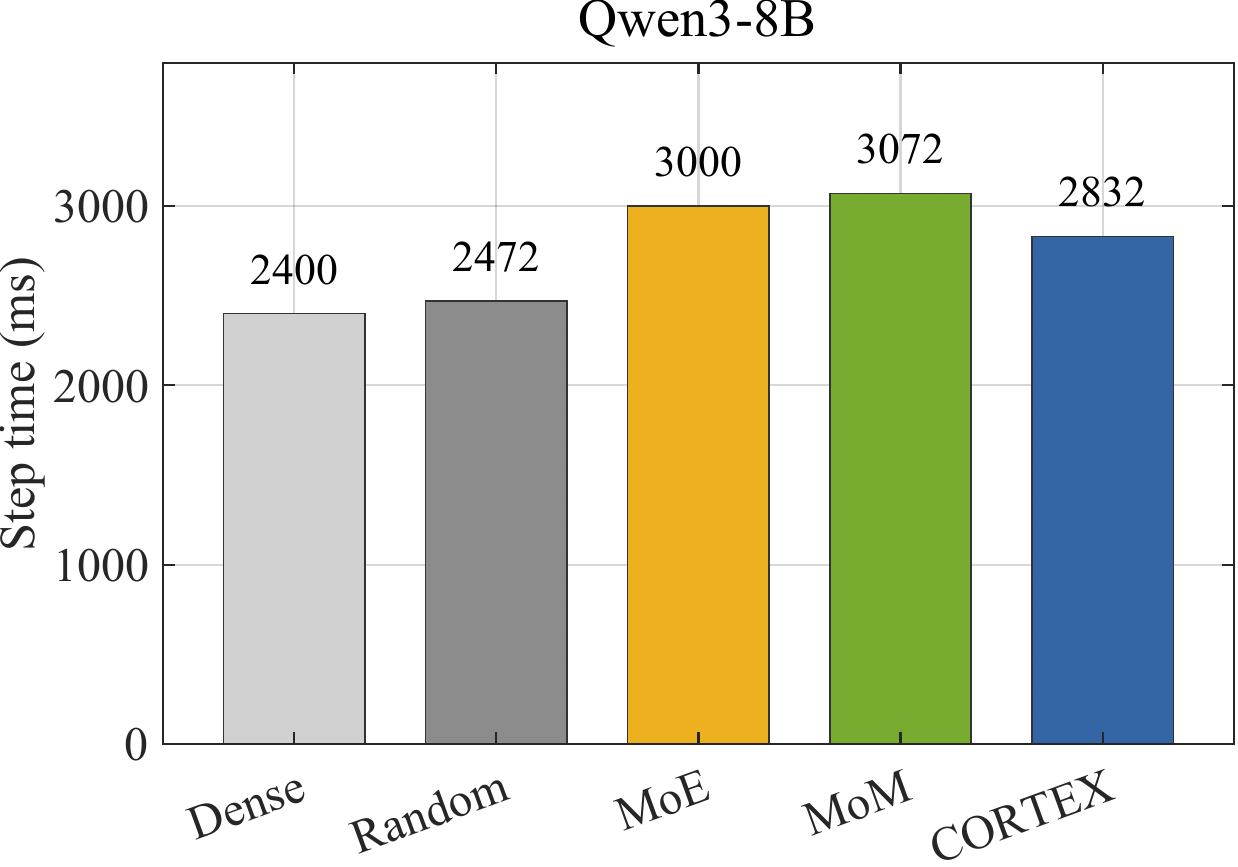}\hfill
    \includegraphics[width=0.32\linewidth]{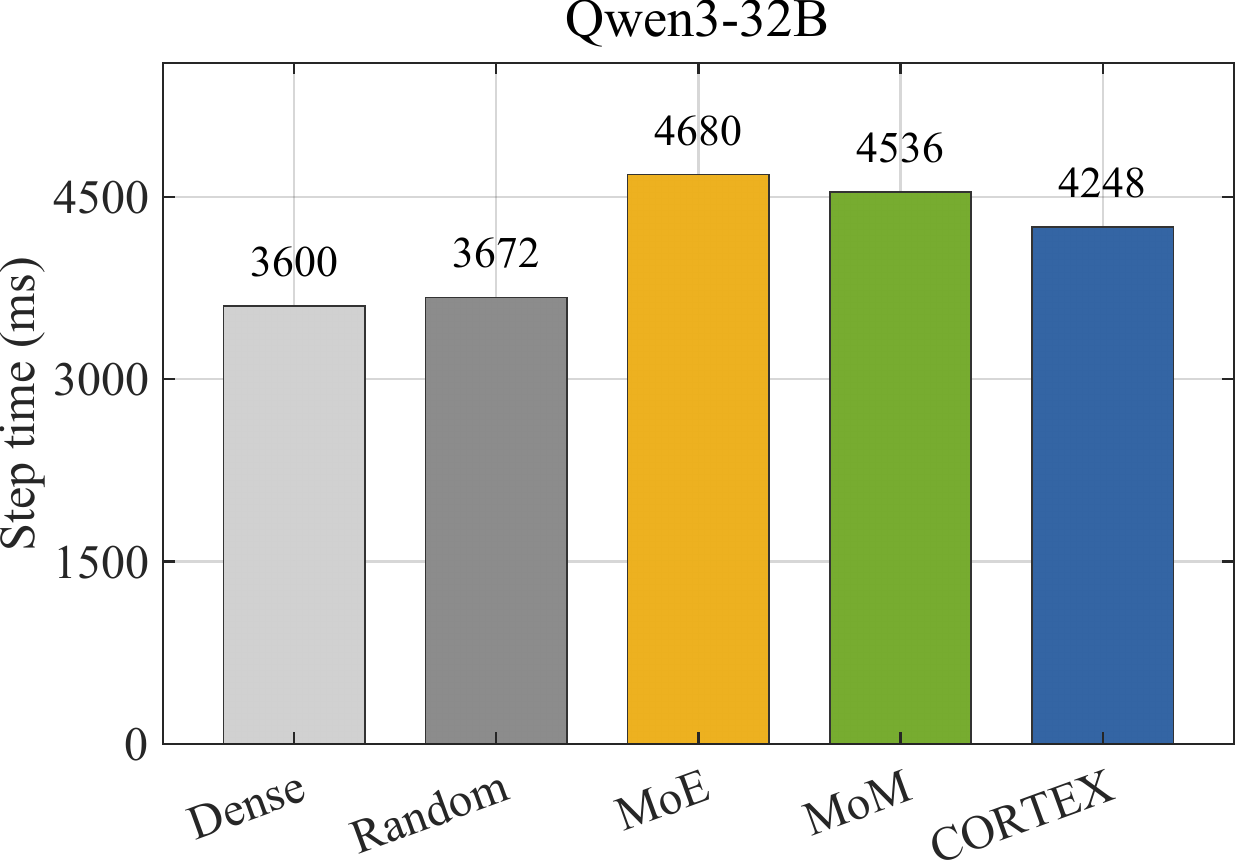}
    \caption{Comparison of training step time on the 160M, Qwen3-8B, and Qwen3-32B backbones.}
    \label{fig:training_time}
\end{figure}

\begin{figure}[!t]
    \centering
    \includegraphics[width=0.32\linewidth]{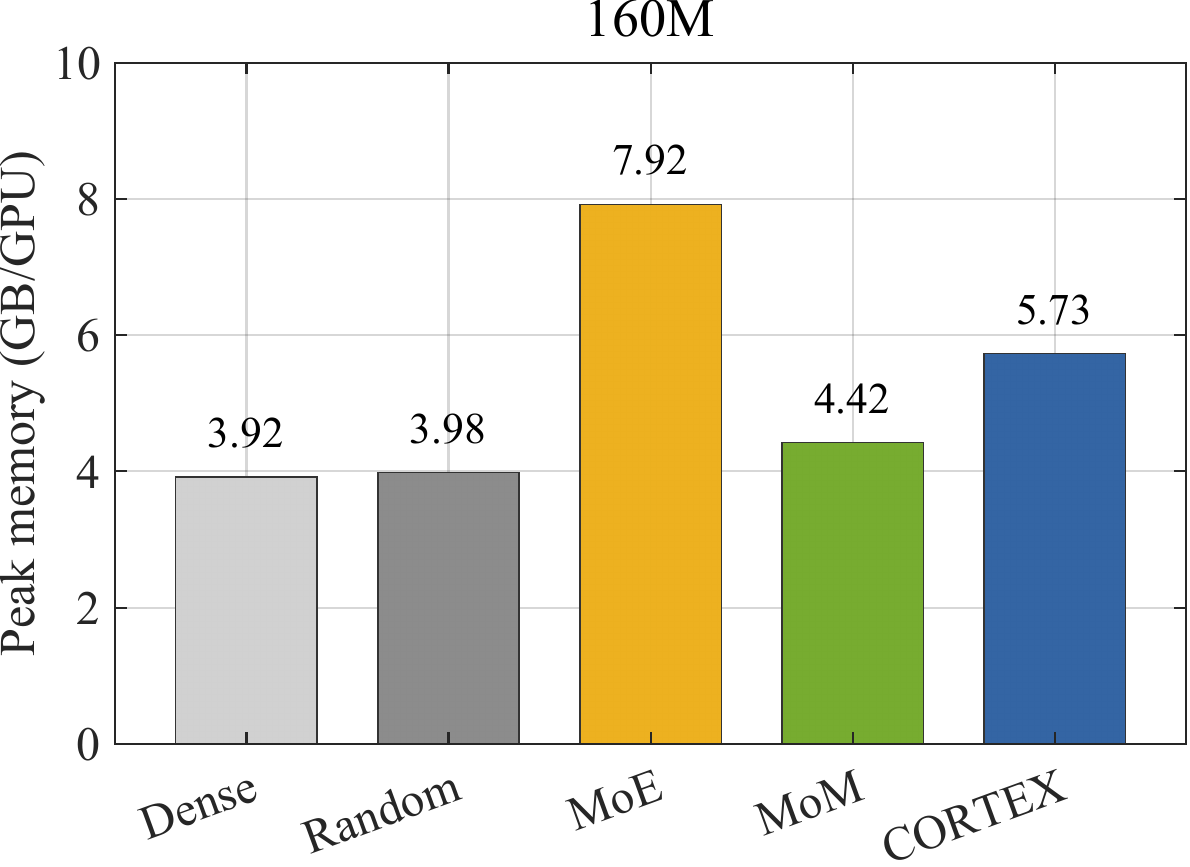}\hfill
    \includegraphics[width=0.32\linewidth]{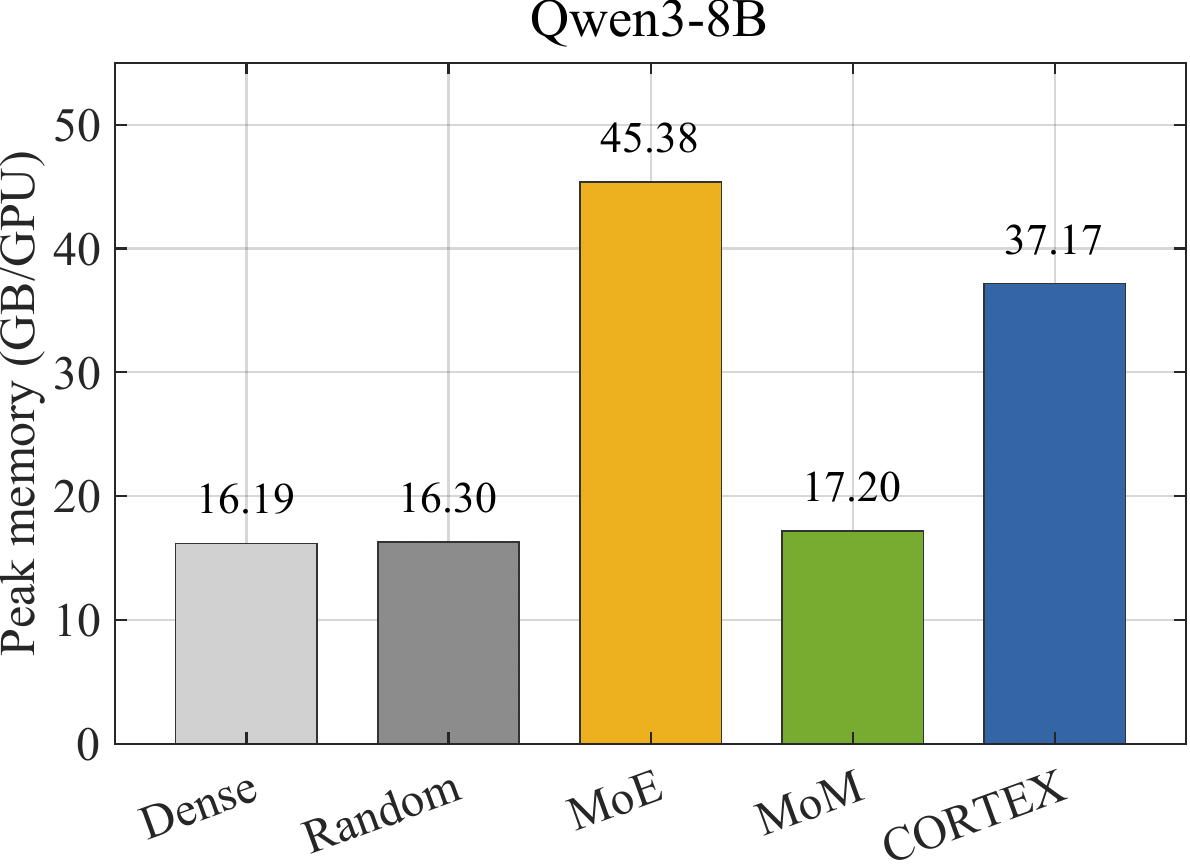}\hfill
    \includegraphics[width=0.32\linewidth]{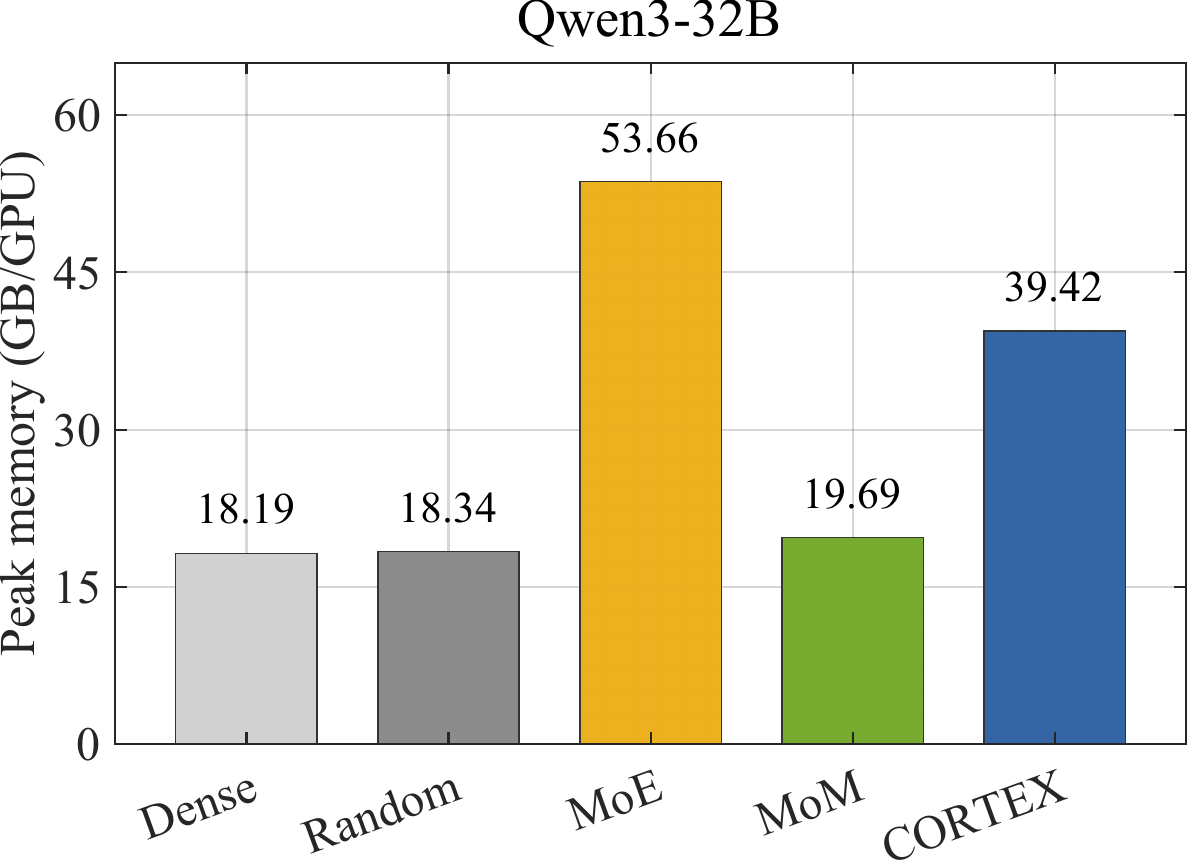}
    \caption{Comparison of peak memory per GPU on the 160M, Qwen3-8B, and Qwen3-32B backbones.}
    \label{fig:training_memory}
\end{figure}

Figs.~\ref{fig:training_time} and \ref{fig:training_memory} compare the training step time and peak memory per GPU across different schemes. Step time is measured over the same token count across methods and includes the complete domain-gradient collection window and both assignment and backbone updates for CORTEX. The 160M, Qwen3-8B, and Qwen3-32B backbones use 1, 4, and 16 H100 80GB GPUs, respectively. When compared with \textbf{Dense}, CORTEX increases the step time by 18--24\% and uses 1.46--2.30 times the peak memory per GPU. CORTEX has a lower step time than MoE and MoM and lower peak memory per GPU than MoE.

\FloatBarrier\subsection{Compute Resources}
\label{sec:compute_resources}
All experiments were conducted on an anonymized institutional H100 GPU cluster using SLURM scheduling. Our GPU experiments were run on NVIDIA H100 SXM5 80GB GPU nodes. For the synthetic benchmark, training and evaluation of the 160M backbone required approximately 523.311 GPU hours, including all baselines and repeated runs. For the real-domain experiments, the Qwen3-8B experiments required approximately 933.486 GPU-hours, and the Qwen3-32B experiments required approximately 8298.656 GPU hours. In total, all experiments in this project used approximately 9755.453 H100 GPU hours. All approaches were evaluated under the same budget within each model scale. These totals include the presented experiments, preliminary experiments, and unsuccessful approach variants that are not reported.

\FloatBarrier\section{Limitations}
\label{sec:limitation}
CORTEX relies on meaningful knowledge-domain labels to estimate domain-conditioned group gradients. 
This assumption is natural for controlled or curated data mixtures, but may be less accurate when knowledge domains become overlapped or noisy. 
The learned modularization also depends on the chosen parameter-group granularity and the number of specialized modules: fine-grained parameter groups may result in noisy module assignment, while overly coarse parameter groups may limit the specialization to target knowledge domains. 
{\color{black}CORTEX introduces overhead from storing and processing domain-conditioned gradients and updating \(\Psi\).} 
No inference-time routing overhead is introduced. 
Finally, CORTEX is not designed as an inference acceleration approach since it aims to improve dense training and provide an interpretable modular decomposition.


\clearpage\endgroup \end{document}